\documentclass[12pt]{article}
\usepackage{arxiv}
\usepackage{natbib}
\usepackage{hyperref}
\usepackage{url}

\usepackage{amsthm}
\theoremstyle{plain}

\newtheorem*{theorem*}{Theorem}

\newtheorem*{example*}{Example}
\newtheorem*{examples*}{Examples}
\newtheorem*{lemma*}{Lemma}

\newtheorem*{corollary*}{Corollary}
\newtheorem*{proposition*}{Proposition}
\theoremstyle{remark}
\newtheorem*{note*}{Note}

\usepackage[utf8]{inputenc}
\usepackage[T1]{fontenc}
\usepackage{xcolor}
\usepackage{amsfonts}
\usepackage{amssymb}
\usepackage{amsmath}
\usepackage{mathtools}
\usepackage{bbm}
\usepackage{dsfont}
\usepackage{subcaption}
\usepackage{booktabs}
\usepackage{multirow}
\usepackage{array}
\usepackage{makecell}
\usepackage{mdframed}

\DeclareMathOperator*{\argmax}{argmax}
\newcommand{\reader}[1]{\mathrm{M}_{#1}}
\newcommand{\areader}[1]{\bar{\mathrm{M}}_{#1}}
\newcommand{\hreader}[1]{\mathrm{M}^{\mathrm{h}}_{#1}}
\newcommand{\readert}[2]{\mathrm{M}_{#1}^{#2}}
\newcommand{\gulliblereader}{\reader{0}}
\newcommand{\brilliantreader}{\reader{1}}
\newcommand{\knowitallreader}{\reader{*}}  %changed this from \infty

\newcommand{\unifpredictor}{\hat{P}_{\mathrm{unif}}}
\newcommand{\expreader}{{\reader{}}_{+}}
\newcommand{\detective}[1]{\mathrm{D}_{#1}}

\newcommand{\culpritrv}{Y}
\newcommand{\culprit}{y}
\newcommand{\trueculprit}{\culprit^{*}}
\newcommand{\culprits}{\mathcal{Y}}
\newcommand{\distractor}{\Bar{\culprit}}
\newcommand{\para}{x}
\newcommand{\storydomain}{\mathcal{X}}
\newcommand{\story}{X}
\newcommand{\storyi}[2]{\story_{#1 \dots #2}}

\newcommand{\clue}{c}

\newcommand{\cluesdomain}{\mathcal{C}}

\newcommand{\clues}{C}
\newcommand{\cluesi}[2]{\clues_{#1 \dots #2}}

\newcommand{\numparas}{L}
\newcommand{\cluesrv}{C}
\newcommand{\storyrv}{X}

\newcommand{\one}{\mathbbm{1}}
\newcommand{\E}{\mathbbm{E}}
\newcommand{\entropy}{\mathrm{H}}
\newcommand{\centropy}{\mathrm{H}}

\newcommand{\surprisalscore}{S}
\newcommand{\coherencescore}{C_{\text{UB}}}
\newcommand{\fairplayubscore}{\text{FP}_{\text{UB}}}

\newcommand{\fairplayskepscore}{\text{FP}_{\text{S}}}
\newcommand{\arfpscore}{\text{FP}_{\text{AR}}}
\newcommand{\expfpscore}{\text{FP}_{\text{ER}}}

\newcommand{\eec}{\text{ERC}}
\newcommand{\eecc}[1]{\eec_{\geq #1}}
\newcommand{\storymodel}{\text{SM}}

\newcommand{\clueeffectiveness}{\textnormal{\sc S-Eff}}
\newcommand{\internalcoherence}[1]{\Delta_{\text{coh}}(#1)}
\newcommand{\intelligencegap}[2]{\Delta_{\text{intel}}(#1, #2)}
\newcommand{\generationsuccess}{{\sc G-val}}
\newcommand{\defn}[1]{\textbf{#1}}

\renewcommand{\shorttitle}{The Challenge and Reward of Fair Play in Narrative}

\begin{document}

\title{The Challenge and Reward of Fair Play in Narrative:\\
A Computational Approach}

\author{
  Eitan Wagner\thanks{Corresponding author. E-mail: eitan.wagner@mail.huji.ac.il} \\
  Department of Computer Science\\
  Hebrew University of Jerusalem\\
  \And
  Renana Keydar \\
  Department of Law and Digital Humanities\\
  Hebrew University of Jerusalem\\
  \And
  Omri Abend \\
  Department of Computer Science\\
  Hebrew University of Jerusalem
}

\maketitle

\begin{abstract}
Good storytelling involves surprise---unpredictability in how the story unfolds---and sense-making, the requirement that the story forms a coherent sequence. However, to date, these two qualities have largely been addressed in isolation.
We formalize these qualities and their relationship in an information-theoretic framework, using detective
fiction as a paradigm case of narratives in which a hidden truth is discovered through reasoning.
Our central theoretical result shows that surprise and coherence must trade off for
any \emph{single} reader model, but can coexist when two reader modes are distinguished:
a pre-revelation mode that forms expectations while the ending is unknown, and a
post-resolution hindsight mode that re-evaluates the story after the culprit is revealed.
The balance of these two dimensions is realized in the common requirement of \emph{fair play}, giving the reader a chance to solve the mystery while maintaining a challenge.
We operationalize the framework using large language models as simulated readers, and define reference-less evaluation metrics for surprise, coherence, and fair play.
Experiments on LLM-generated stories validate our theoretical predictions: while models generally succeed in creating surprise or coherence, achieving fair play poses a challenge even for strong models. Moreover, surprise and coherence do not positively correlate across stories, resisting reduction to a single latent quality.
A human study validates the metrics, confirming they capture aspects of narrative quality that matter to readers.
Our metrics also reproduce established literary intuitions, finding Christie's stories more surprising and more fair-playing than Conan Doyle's.

\end{abstract}


\section{Introduction}

Surprise is a key driver of narrative engagement, with unpredictable events shown to be
highly appreciated by readers, and to enhance their processing of story content
\citep{hoeken2000suspense}.
However, unpredictability was found to be only one component in the subjective experience
of surprise, with a sense-making process---through contrast and consolidation with
existing knowledge---also playing a crucial role \citep{teigen2003surprises,maguire2011making}.
Local and global coherence---in addition to unexpectedness---are necessary for a narrative
to evoke interest \citep{silvia2005interesting,silvia2008interest}, and to be considered
successful \citep{van1984between} or creative \citep{simonton2012taking}.

Detective stories are a representative example of narratives that exploit this duality.
The genre centers on a hidden truth---commonly the identity of the culprit (``whodunit'')---gradually revealed
through sense-making: clues scattered across the narrative form a coherent explanatory
structure that the reader must reconstruct.
This quality has made the genre a paradigm case for reasoning under uncertainty across
disciplines, from historical inquiry \citep{collingwood1993idea} and abductive inference
spanning science, medicine, and psychoanalysis \citep{ginzburg1980morelli} to NLP
benchmarks for multi-step reasoning \citep{10.1162/tacl_a_00001, del-fishel-2023-true,
sprague2024musrtestinglimitschainofthought}: in all cases, an observer (the detective)
reasons from observable clues to a nontrivial hidden truth.
Yet the genre adds a second challenge beyond sense-making: a narrative machinery of
misdirection, red herrings, and delayed revelation, explicitly designed so that the reader
does not see the ending coming \citep{brewer1980event, schraw2001increasing}.

This dual requirement is captured in the literary concept of \emph{fair play}
\citep{knox1929best, huhn1987detective}: ``fair'' meaning the story is retrospectively
solvable---the clues must coherently support the revelation and give the reader a genuine
chance---and ``play'' meaning there is a genuine challenge---the ending must not be
trivially predictable.
A notable failure mode of the ``fair'' dimension is a \emph{Deus ex Machina}, where the
resolution emerges without connection to the prior investigation
\citep{sallas2024logic,Effron_2017}.
Achieving fair play is notoriously difficult: it requires precise authorial control over
what readers infer and when---keeping the revelation both surprising and coherent in
retrospect \citep{sayers1936introduction,tobin2009cognitive}.

In this paper, we formalize and study this
challenge through detective fiction. We use Agatha Christie's \textit{The Murder of Roger Ackroyd}
\citep{christie1926ackroyd} as a running example (Box~1).
We show that high surprise and high retrospective coherence can coexist,
but only by engineering a \emph{shift in the reader's interpretive state}: what surprises the reader mid-story must become coherent---even inevitable---in hindsight.
We formalize these ideas in an information-theoretic framework.
As we show in \hyperref[sec:coherent-surprise]{The Coherence-surprise tradeoffs}, surprise and retrospective coherence necessarily trade off for any \emph{single} reader model. A reader who updates beliefs coherently from the clues cannot simultaneously be surprised by the revelation.
Yet, both qualities can be jointly achieved when two distinct reader modes are distinguished---one registering surprise before the revelation, one finding the resolution
retrospectively coherent.
The tradeoff thus reframes fair play not as a compromise on a single quality axis, but as a two-dimensional target.

This framing has a direct consequence for measurement.
Prior computational work separately proposed metrics for narrative surprise \citep{ely2015suspense, wilmot-keller-2020-modelling}, and for narrative coherence \citep{maimon-tsarfaty-2023-cohesentia, mostafazadeh-etal-2016-corpus}.
Of particular note, Ely et al. \citep{ely2015suspense} propose a probabilistic framework for
suspense and surprise in entertainment, including mystery novels; however, their framework
does not address the tension between surprise and coherence that is at the heart of fair play.
Measuring only one dimension yields an incomplete picture that fails to capture the distinct elements of the detective genre: a surprising story may lack retrospective coherence, and a coherent story may lack genuine surprise.
% We accordingly evaluate both dimensions showing a dissociation between them.

We operationalize the framework using large language models (LLMs), which give us control over the generation process to simulate multiple reader types---from one reading na\"ively,
to one with vast experience, to one with full knowledge of the writing process. We validate our approach through human reader experiments.
Our results show that: (i) many models achieve adequate surprise or retrospective coherence individually, but fail to combine them into fair play, validating the importance
of evaluating both dimensions jointly; (ii) automated metrics on real stories reproduce
established literary intuitions---Christie's Poirot stories are substantially more surprising than Conan Doyle's Sherlock Holmes stories, consistent with Christie's
reputation for narrative twists; and (iii) human subjective ratings for fairness, coherence, and enjoyability are substantially higher for real than generated stories, and significantly correlate with the automated metrics, validating the framework.

\begin{figure}[h]
\begin{mdframed}[backgroundcolor=gray!10,linewidth=1pt,skipabove=4pt,skipbelow=4pt]
\small
\textbf{Box~1. \textit{The Murder of Roger Ackroyd} (1926) --- Synopsis.}\\[4pt]
The story is narrated by Dr.\ James Sheppard, Ackroyd's physician and neighbor, who assists Poirot in the investigation. Suspicion builds around Ralph Paton---Ackroyd's
stepson, who disappears after the murder and is revealed to have had both motive (financial need, a secret marriage) and opportunity. Poirot ultimately uncovers that Sheppard himself is the killer, having used a dictaphone to fabricate an alibi. The story is a canonical example of fair play: the clues pointing to Sheppard are all present in his own narration,
but his trusted-narrator status misleads a gullible reader who does not rigorously analyze the evidence.
\end{mdframed}
\label{box:ackroyd}
\end{figure}

%%----------------------------------------------------------------------
\section{The Probabilistic Framework} \label{sec:probabilistic-framework}

We now formalize stories, readers, and detectives as probabilistic objects. Our framework applies to any narrative with a gradually revealed hidden truth. However, for simplicity, we use
detective story terminology throughout. A more detailed literary discussion is in the SI Appendix.

\subsection{Detective Stories and their Components} \label{sec:detective-stories-components}

The conventional structure of a whodunit story includes three phases \citep{huhn1987detective,
Knight_2003}: a crime is committed and suspects are presented in the \defn{introduction} phase; clues unfold across the \defn{suspicion} phase; and a hidden truth (the culprit's identity) is disclosed in the \defn{revelation} phase
\citep{dove1997reader}.

We formalize this as follows. Let a \defn{story} be a string of text $\story$, divided into $\numparas$ paragraphs $\story=\para_1, \para_2, \ldots, \para_\numparas$. We denote the set of all possible stories with $\storydomain$, and refer to the first $i$ paragraphs of a story as a \defn{story prefix}, denoted $\storyi{1}{i}$.
We define the \defn{revelation point} as the paragraph in which the culprit's identity is first revealed.
The suspicion phase unfolds through \defn{clues} embedded in the story's paragraphs---undisputed facts visible to the reader that collectively build toward the resolution.

The framework centers on a \defn{hidden variable} $\culpritrv$, representing an aspect of the \defn{underlying story} that must be uncovered. We denote
the set of possible values of $\culpritrv$ with $\culprits$. In whodunit detective stories, the underlying story is the crime committed, and we identify $\culpritrv$ with the identity of the culprit.
The set $\culprits$ thus consists of the potential culprits, which specifically include the \defn{true culprit} $\trueculprit \in \culprits$, who actually committed the crime, and at least one \defn{distractor} $\distractor \in \culprits$---a character who is not
the true culprit, but is cast as a likely suspect during (part of) the
investigation.
We assume that all potential culprits are introduced in the \defn{introduction} phase,
and are therefore known at the outset of the \defn{suspicion} phase.

\subsection{Reader Models} \label{sec:reader-models}

Literary theory recognizes the reader as an active participant in the investigation, similar to a detective \citep{huhn1987detective, dove1997reader}. 
The reader is presented with a written story or part of it and infers the identity of the culprit. Unlike detectives who work from clues alone,
readers have access to the full written text, including narrative framing and stylistic cues. 
% A reader is similar to a detective, but has access to additional input, since the text includes more than just the clues. 
For example, a reader might notice and make use of an author's typical twists or their tendencies regarding traits and genders of characters.
% For example, a reader sees the order in which characters are introduced, and the terms the narrator uses to describe the characters.

Formally, a \defn{reader model} is
a function $\reader{}: \storydomain \times \mathbb{N} \to \Delta^{|\culprits|-1}$,
which represents a mapping from a story and a point within it to a probability distribution over some property of the underlying story $\culpritrv \in \culprits$.
$\Delta^{|\culprits|-1}$ is the $|\culprits|$-dimensional simplex.
In our case, we assume $\culpritrv$ represents the culprit who commits the crime. The probabilities are over culprit identities $y \in \culprits$, and reflect the reader's belief after being exposed to the first $i$ paragraphs.
We use $\reader{}(i)$ instead of $\reader{}(\story, i)$ when the identity of the story is clear. We use $\reader{}(0)$ for the probabilities with the empty prefix.
When reading a new story, a reader determines the output based on prefixes (up to a current paragraph $i$), that is, $\reader{}(\story, i) \coloneqq p(\cdot\mid\storyi{1}{i})$.
Each reader model $\reader{}$ induces a corresponding \defn{culprit predictor} $\hat{y}_{\reader{}}(i) = \argmax_{y \in \culprits} \reader{}(i)_y$, which outputs the
most likely culprit according to the reader at step $i$.

\subsection{Idealized Types of Reader Models} \label{sec:idealized-reader-types}

\begin{figure*}[t]
\centering
\begin{subfigure}[c]{0.33\linewidth}
\centering
\includegraphics[width=\linewidth]{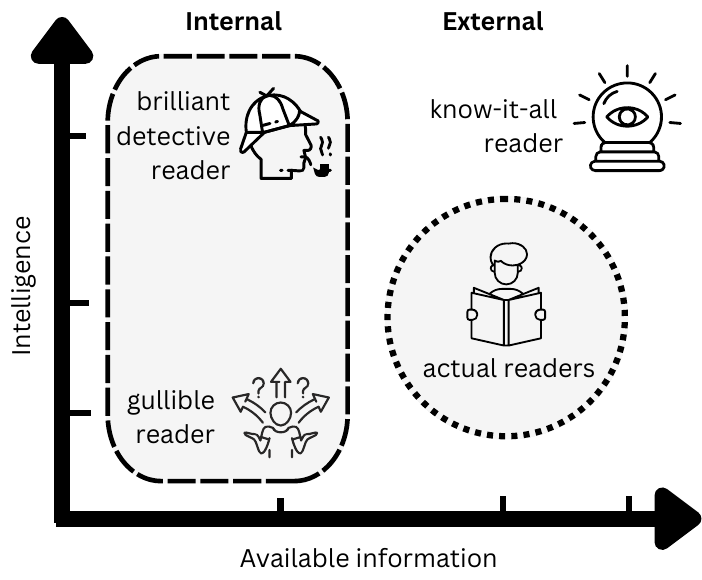}
\phantomcaption\label{fig:readers}
\end{subfigure}
\hfill
\begin{subfigure}[c]{0.66\linewidth}
\centering
\includegraphics[width=\linewidth]{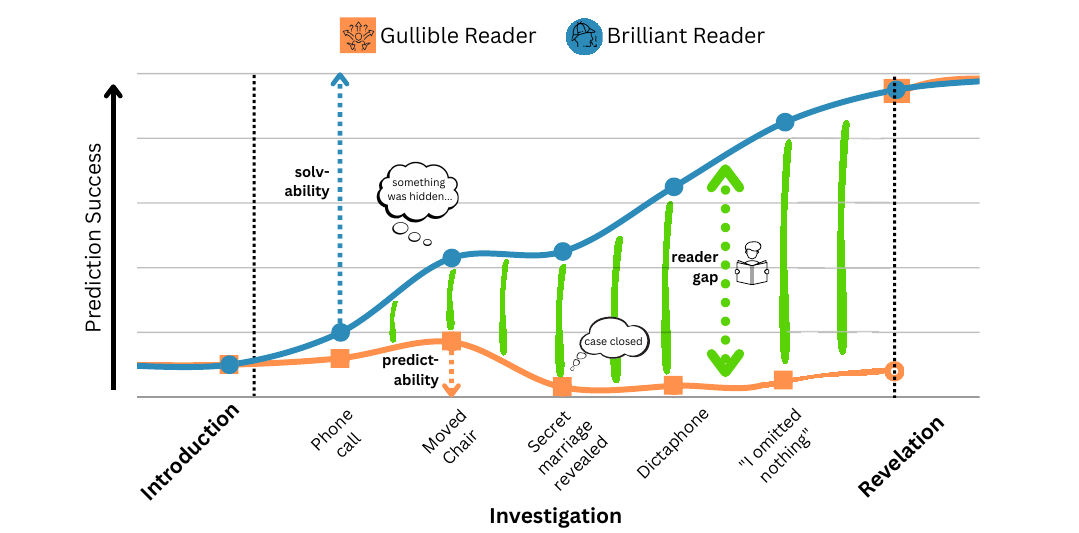}
\vspace{-0.5cm}
\phantomcaption\label{fig:ackroyd-curves}
\end{subfigure}
\caption{\textbf{(A)} Schematic map of the reader models by inference ability and
accessible information. \textbf{(B)} Desired qualities of readers. We expect low
predictability for the gullible reader (orange with squares) and high solvability for the brilliant reader (blue with circles). Combining these qualities leads to a gap between readers, which is challenging to create. The plot illustrates the effects of clues in \textit{The Murder of Roger Ackroyd}: the phone call and the dictaphone construct a
false alibi for the narrator that the gullible reader accepts; the moved chair points to where the recording device was hidden---a detail the brilliant reader connects to the
deception; the revelation of a secret marriage misdirects the gullible reader toward an innocent suspect while the brilliant reader uses it to exculpate him, and the narrator's assurance ``I omitted nothing'' is taken at face value by the gullible
reader but identified as evasion---not denial of guilt---by the brilliant reader.}
\end{figure*}

We define four idealized readers used for quality measurement.
Two are \defn{internal}: they simulate a detective observing only the story's clues.
Two are \defn{external}: they may draw on information beyond the story text.
Within each pair, the two readers differ in inference capability.
Figure~\ref{fig:readers} presents a schematic mapping on the internal/external and intelligence axes.

\paragraph*{Internal readers}
Internal readers evaluate only the story's clues, differing in inference quality.
The \defn{gullible reader} $\gulliblereader$ na\"ively follows surface-level clues and is susceptible to misdirection: in the extreme case it forms its prediction based on a single prominent clue, ignoring the rest. This reader simulates an incompetent detective (typically a sidekick, like Watson or Hastings), as common in detective fiction.
The \defn{brilliant reader} $\brilliantreader$ makes the best possible use of the internal clues, arriving at the true posterior over culprits given the evidence. This reader simulates the brilliant detective (like Holmes or Poirot).
The gullible reader represents the lower bound of inference capability: it selects suspects based only on surface information, falling for misdirections, and a good story
should consistently fool it.

\paragraph*{External readers}
External readers are not limited to internal clues; they may draw on information beyond the story text itself. This includes familiarity with the genre or generating model, as well as meta-information such as their position in the narrative---for example, knowing that the revelation is near because relatively few pages remain, something an internal detective has no access to.

An \defn{actual reader} exploits such external information to predict the culprit. Actual readers form a spectrum: at the lower end, a reader with minimal prior exposure relies primarily on the current story's clues; at the upper end, a highly experienced reader leverages patterns from many previous stories, even when internal clues are insufficient.
We define the \defn{average actual reader} $\areader{}$ as a mixture over this spectrum, weighted by the prevalence of each reader type in the population.

We define the \defn{know-it-all reader} $\knowitallreader$ as an external reader with full knowledge of the true story generation distribution---capturing all that can be known about how stories are written. The know-it-all reader is still constrained by the randomness of the distribution itself.\footnote{\textit{The Mystery of Edwin Drood} \citep{dickens1870mystery} serves as an illustration: Dickens died before completing the novel, prompting many subsequent writers to provide their own endings \citep{dickens1913complete}. The various endings demonstrate the distinction between an actual writer---who may have planned a specific ending---and a story \emph{model}, which represents a distribution over continuations of a given prefix. The know-it-all reader knows this distribution but cannot determine which specific ending would have been chosen.}

\subsection{Expected Reader Behavior} \label{sec:expected-reader-behavior}

The reader models presented above are designed with an expected behavior in mind. 
We turn to describe how we expect them to behave across the
three phases of a detective story: introduction, suspicion, and revelation (see Figure~\ref{fig:ackroyd-curves}).

In the \textbf{introduction}, all suspects have just been presented, and no meaningful clues have been revealed. All reader models thus assign roughly
uniform probabilities over the set of suspects, i.e.,
$\reader{}(i)_y \approx \frac{1}{|\culprits|}$ for all $y \in \culprits$.

In the \textbf{suspicion phase}, the reader models diverge:
\begin{itemize}
    \item The \textbf{gullible reader} $\gulliblereader$ 
    is gradually misled by the red herrings and surface-level clues. The probability for the true culprit decreases, while the probability assigned to distractors increases.
    \item The \textbf{brilliant reader} $\brilliantreader$ optimally interprets the clues and gradually increases the probability assigned to the true culprit.
    \item The \textbf{know-it-all reader} $\knowitallreader$, which has access to the story generation process, similarly increases the probability of the true culprit, potentially with higher confidence than the brilliant detective.
    \item The \textbf{actual readers} fall between the gullible and know-it-all readers, with more experienced readers leveraging external knowledge, such as patterns from previous stories of the same model or genre.
\end{itemize}

In the \textbf{revelation}, the true culprit is revealed, and so $\reader{}(i)_{\trueculprit} \approx 1$ for all readers.

The quality of a detective story can be understood in terms of these expectations. A strong \emph{surprise} occurs when a reader is strongly misled during the suspicion phase---which happens to the gullible reader. \emph{Coherence} reflects the reader's ability to gradually identify the true culprit before the revelation---as with the
brilliant and know-it-all readers. A story achieves \emph{fair play} when both conditions hold simultaneously: the clues support the correct conclusion for a careful
reader, while misleading a less attentive one.

\paragraph*{Optimality} \label{sec:story_generation}
Stories are generated by a probabilistic \defn{story generation model} $\storymodel$,
defining a distribution over story texts, which also induces a distribution over sequences of clues.
Two reader types have natural optimality characterizations: the \defn{know-it-all reader}
$\knowitallreader$ is the Bayes-optimal classifier given the story generation distribution $\storymodel$, and the \defn{brilliant reader} $\brilliantreader$ is the Bayes-optimal classifier given only the internal clue distribution. These optimality conditions formalize the intuition that a coherent story leaves enough information for an ideal reasoner to solve it. See SI Appendix for the formal treatment and proofs.

%%----------------------------------------------------------------------
\section{The Coherence-Surprise Tradeoffs} \label{sec:coherent-surprise}
We formalize two notions of surprise and derive two tradeoffs with coherence: a weak tradeoff showing that intelligent readers face a tension between high coherence and weak surprise, and a strong tradeoff showing that the reader's intelligence limits the simultaneous achievement of high coherence and strong surprise. We then introduce the \emph{hindsight gap}
as the mechanism that allows for the coexistence of high coherence and strong surprise.

\subsection{Surprise}
% \subsubsection*{Informedness of the Reader} \label{sec:cross_entropy}
The know-it-all reader knows all that can be known about how the story might evolve. We
therefore define a reader's degree of \defn{uninformedness} based on how it diverges from the know-it-all reader.
Formally, given a reader $\reader{}$, we define its uninformedness
as its cross entropy relative to  $\knowitallreader$:
\begin{align}
\centropy(\knowitallreader(i);\, \reader{}(i)) = -\sum_{\culprit \in \culprits}
p_{\knowitallreader}(\culprit \mid \storyi{1}{i}) \log p_{\reader{}}(\culprit \mid
\storyi{1}{i})
\end{align}

This divergence is minimal where $\reader{}=\knowitallreader$ and equals the entropy $\entropy(\knowitallreader(i))$.
We assume that any reader $\reader{}$ starts with an uninformed (i.e., uniform) prior
over culprits.
Therefore:
\begin{align}
    \centropy(\knowitallreader(0);\, \reader{}(0)) \hspace{.5em} = \hspace{.5em} \entropy(\knowitallreader(0)) \hspace{.5em} = \hspace{.5em} \log|\culprits|
\end{align}

% \subsubsection*{Surprise.}
In reading comprehension, surprise is commonly measured as the negative log-probability
assigned to a word \citep{hale-2001-probabilistic, smith2008optimal, demey2015dynamics} (See SI Appendix, Related Work, for a discussion of other surprisal measures in narratives). The expectation of
the negative log-probability, over the story generation model, is exactly the
uninformedness of the reader.

A weak notion of surprise is when a reader is uninformed about the outcome until it is revealed.
We say that $\reader{}$ is \defn{weakly surprised} in step $i$ if:
\begin{align}
    \centropy(\knowitallreader(i);\, \reader{}(i)) \geq \log|\culprits| - \epsilon_{\text{surprise}}
\end{align}
for some small $\epsilon_{\text{surprise}} > 0$.

A stronger notion of surprise is when the reader is misled into expecting an outcome other than the actual one.
We say that $\reader{}$ is \defn{strongly surprised} with respect to step $i$ if it is less informed than a uniform prediction:
\begin{align}
    \centropy(\knowitallreader(i);\, \reader{}(i)) \geq \log|\culprits| + \delta_{\text{surprise}}
\end{align}
for some threshold $\delta_{\text{surprise}} > 0$.

Put differently, weak surprise means that the reader is close to having no knowledge.
Strong surprise means that a reader is misled into a belief that is further from the truth than a state of no knowledge.

\subsection{Coherence}\label{sec:formal_coherence}

In narrative terms, coherence means that all parts of a story contribute to a unified whole; in detective fiction, this specifically means that the clues collectively and
meaningfully support the resolution. We formalize this through the cumulative effectiveness of the clues as interpreted by the brilliant reader.

% \subsubsection*{Step effectiveness.}
\phantomsection\label{sec:clue_effectiveness}
A step is effective when it provides information to the reader, allowing better prediction of the culprit identity.
Formally, we define the \defn{step effectiveness} (\clueeffectiveness) for a reader $\reader{}$ at step $i+1$ as the expected reduction in their uninformedness when moving from step $i$ to step $i+1$:
\begin{align}
    &\clueeffectiveness_{\reader{}}(i+1) = \E_{\storyi{1}{i} \sim \storymodel}\centropy(\knowitallreader(\storyi{1}{i});\, \reader{}(\storyi{1}{i})) \nonumber\\& \quad- \E_{\storyi{1}{i+1} \sim \storymodel}[ \centropy(\knowitallreader(\storyi{1}{i+1});\, \reader{}(\storyi{1}{i+1}))]
\end{align}

For brevity, we will simply use $\E(\centropy(\reader{}(i)))$ instead of $\E_{\storyi{1}{i} \sim \storymodel}(\centropy(\reader{}(\storyi{1}{i})))$.
Positive step effectiveness indicates that the reader becomes, in expectation, more
similar to $\knowitallreader$.

% \subsubsection*{Clue effectiveness}
\defn{Clue effectiveness} is the cumulative step effectiveness of the
brilliant reader $\brilliantreader$, indicating the total contribution of the
clues to identifying the culprit, after $i$ steps:
\begin{align}
    \internalcoherence{i} \coloneqq \sum_{j=1}^i\clueeffectiveness_{\brilliantreader}(j)
\end{align}
% Larger values indicate that the clues collectively reveal more about the culprit's identity.

High clue effectiveness means that the clues collectively contribute to identifying the culprit, according to the brilliant reader's judgment, which, assuming clue sufficiency (as defined below), is close to the truth. Low clue effectiveness means that the clues have a low
contribution, generally indicating poor quality.

% \subsubsection*{Clue sufficiency}
The \defn{clue sufficiency property} of a story generation model $\storymodel$ requires
that the step effectiveness of the brilliant reader $\brilliantreader$ is close
to that of the know-it-all reader $\knowitallreader$:
\begin{align}\label{eq:external_coherence}
\big|\clueeffectiveness_{\brilliantreader}(i) - \clueeffectiveness_{\knowitallreader}(i)\big| \leq \epsilon_{\text{ex}}, \quad \forall i \in 1 \dots \numparas
\end{align}
for a small $\epsilon_{\text{ex}}>0$.

This property captures the notion that it is enough to brilliantly interpret the internal clues to (gradually) solve the mystery: the clues are not merely effective, they are \emph{sufficient}.
% Clue sufficiency can be seen as a result of the story generation process not diverging from the genre's conventions (see SI Appendix).

A simple generation model that violates clue sufficiency is one that has very low diversity in its generated stories (e.g., it always generates the same story). In this case, a reader with access to previous stories can predict the culprit more confidently than the brilliant reader who relies only on internal clues.

% \subsubsection*{Intelligence gap}
Given clue sufficiency, the ability of any reader to make inferences based on clues is bounded by that of the brilliant reader.
We define the \defn{intelligence gap}, at step $i$, between a specific reader $\reader{}$
and the brilliant reader as the difference in cumulative step effectiveness:
\begin{align}
    \intelligencegap{i}{\reader{}} \coloneqq \sum_{j=1}^i \left[\clueeffectiveness_{\brilliantreader}(j) - \clueeffectiveness_{\reader{}}(j) \right]
\end{align}

We say that the reader $\reader{}$ is \defn{intelligent} if the intelligence gap is
small:
\begin{align}
    \intelligencegap{i}{\reader{}} \leq \epsilon_{\text{intel}}, \quad \forall i \in 1 \dots \numparas
\end{align}
for a small $\epsilon_{\text{intel}}>0$.

Intelligence means that the reader's interpretation of the clues is similar to that of the optimal detective model.
Consequently, the intelligence gap with the gullible reader $\intelligencegap{i}{\gulliblereader{}}$ serves as a practical upper bound for the intelligence gap with actual readers.

Notably, a reader with expectations similar to the brilliant reader's qualifies as intelligent. Similar culprit probability estimates \emph{imply} intelligence, though the converse does not necessarily hold (see SI Appendix).

% In our empirical metrics, the intelligence gap is implicitly captured by contrasting
% readers at the two extremes: the gullible reader $\gulliblereader$ (maximal gap) and the
% know-it-all reader $\knowitallreader$ (zero gap by definition). The fair play upper bound
% $\fairplayubscore$ measures the performance difference between them, encoding the
% practical consequence of the gap.

\subsection{Tradeoffs} \label{sec:tradeoffs}

We now present the main theoretical result:
\begin{theorem*}[Coherence-surprise tradeoffs] \quad
\begin{enumerate}
    \item
    \textbf{(Weak tradeoff)}
        An intelligent reader $\reader{}$ satisfies a
        tradeoff between clue effectiveness $\internalcoherence{i}$ and weak surprise $-\epsilon_{\text{surprise}}$.
    \item \textbf{(Strong tradeoff)}
        For any reader $\reader{}$, strong surprise $\delta_{\text{surprise}}$ and clue effectiveness $\internalcoherence{i}$ trade off according to the intelligence gap $\intelligencegap{i}$, satisfying
        $\delta_{\text{surprise}} + \internalcoherence{i} \leq \intelligencegap{i}{\reader{}}$.
        In particular, assuming clue sufficiency, an intelligent reader $\reader{}$ cannot be strongly surprised.
\end{enumerate}
\end{theorem*}
\begin{corollary*}[Length difficulty]
As stories become longer ($\numparas$ grows), maintaining fair play becomes increasingly
difficult: either the clues carry very little information---with a plot that seems to
make no progress---or the need for misdirection becomes stronger.
\end{corollary*}
Formal statement and proof are in the SI Appendix.

\subsection{The Hindsight Gap} \label{sec:hindsight}
The tradeoffs above show that surprise and coherence are fundamentally in tension for a
single reader. How can a single actual reader experience both? The answer lies in \emph{hindsight}: a shift in interpretation that occurs after the revelation.

A single actual reader operates in two modes: a \defn{forward-reading mode}, forming
expectations as the story unfolds, and a \defn{hindsight mode}, re-evaluating the clues
after the culprit is revealed.
The \defn{hindsight gap} is the increase in cumulative clue effectiveness when the reader
shifts from forward-reading to hindsight---the ``aha'' moment where earlier clues become
meaningful in retrospect.
Analogously to the intelligence gap---which measures a reader's clue-interpretation shortfall relative to the brilliant reader---the hindsight gap measures a reader's gain relative to its own earlier, forward-reading self.
A story satisfies \defn{hindsight coherence} when the revelation provides a complete explanation of the clues---which in turn leads the hindsight mode to re-evaluate the story as well as the brilliant reader would (formal statement in SI Appendix).

\begin{corollary*}[Hindsight tradeoff]
    Assuming hindsight coherence, if a reader $\reader{}$ is strongly surprised then clue effectiveness and the hindsight gap trade off (proof in the SI Appendix).
\end{corollary*}

\paragraph*{Implications for detective story writing}
Three notable story archetypes occupy different regions of the tradeoff space:

\begin{enumerate}
    \item \textbf{Deus ex machina:} Low step effectiveness throughout the story yields surprise without coherence. The reader has no fair chance to solve the mystery, as the final outcome
    is only weakly constrained by the clues.
    \item \textbf{Non-mystery:} High step effectiveness can make the culprit predictable throughout, yielding coherence without surprise. These stories
    allow the reader to solve the mystery but provide weak narrative tension.
    \item \textbf{Fair play:} A hindsight gap enables both strong surprise and coherence simultaneously. The reader is misled during the story yet can recognize the logical structure in retrospect. Creating such a gap is a central challenge of detective story writing.
\end{enumerate}

Figure~\ref{fig:ackroyd-curves} illustrates these archetypes using \textit{The Murder of
Roger Ackroyd}: the gullible reader's suspicion consolidates around the distractor (Ralph Paton), while
the brilliant reader's confidence in the true culprit (Dr. Sheppard) rises steadily.
% ---a trajectory characteristic of the fair play archetype.

We note that without clue sufficiency, clue effectiveness may be negative---meaning the gullible reader outperforms the brilliant reader---which can allow for strong surprise even for an intelligent reader. While such stories exist, they violate genre conventions.

A successful detective story must maintain diverging predictions by two reader models: a
less-than-brilliant reader, such as the gullible reader, who can be
surprised by misdirection, and an intelligent reader closer to the brilliant reader, who can recognize the coherent logical structure of the clues. A larger gap between the readers allows for simultaneously stronger surprise and coherence.
For these two reader models to be combined in a single actual reader, a shift must occur
during reading---precisely the hindsight gap.\footnote{We note that hindsight
coherence may not always hold: a reader may \emph{feel} the ending was solvable in retrospect due to hindsight bias, even when the clues did not genuinely support it. A reader may judge an outcome as predictable in hindsight, even if it cannot be predicted in practice, thus not being truly fair. This ``faux fair play'' is distinct from genuine fair play, which our framework and metrics are designed to capture.} 

Constructing stories with a hindsight gap is challenging. The reader must be manipulated
to be initially surprised while accepting the true story in retrospect. The challenge is
amplified when writing many stories, as the writer must constantly introduce original
challenging puzzles \citep{sayers1936introduction}.

%%----------------------------------------------------------------------
\section{Experimental Setup} \label{sec:experiments}

Having developed the theoretical framework, we now operationalize it in practice.
We define empirical metrics grounded in the reader models introduced above, and apply them to both LLM-generated and real detective stories.
Our goal is twofold: to test whether the coherence-surprise tradeoff manifests empirically---as a genuine challenge reflected in model performance---and to validate whether surprise and coherence behave as competing qualities that require joint evaluation.

\subsection{Stories}\label{sec:stories}
For generated stories, we use a range of LLMs spanning different model families and
scales---from small open-source models to large frontier models---including Gemini, Llama, and GPT variants.
For real detective stories, we compare works by Arthur Conan Doyle (Sherlock Holmes) and
Agatha Christie (Hercule Poirot). These stories are regarded as milestones in the evolution of detective fiction \citep{moretti2000slaughterhouse}, with Conan Doyle's stories as
``howdunits''---due to their focus on the deductive process \citep{eco1983horns}---and
Christie's as ``whodunits'' \citep{scaggs2005crime}.
See Methods and SI Appendix for additional details on all story sets.

\subsection{Metrics}\label{sec:metrics}
The theoretical framework defines surprise, coherence, and fair play in terms of reader models. Here we translate these into empirical metrics computable from story text, grounding the abstract quantities in measurable terms.
Some metrics require access to the generation model's probabilities and can therefore only be applied when the generating distribution is known. For real stories, these metrics can be approximated using a sufficiently strong predictor as a proxy, although this provides no theoretical guarantees on the quality of the approximation.
Importantly, since we have no guarantees regarding the calibration of model outputs and
human predictions---especially when the goal is maximizing accuracy
\citep[see][]{wagner2025language}---we opt for metrics based on the classifier $\hat{y}_{\reader{}}(i)$ induced by each reader model rather than raw probability estimates.
For related work, including previously used metrics of surprise and coherence, see
SI Appendix, Related Work.

Stories that do not clearly establish both a culprit and a distractor---judged by a held-out model---are excluded from analysis and treated as generation failures; full details are reported in the SI Appendix.

\paragraph*{Surprise and coherence}
Since cross-entropy is frequently unstable in practice---log-probabilities can take infinite values---we use classification accuracy as our base measure.
For any reader model $\reader{}$, we define its \defn{prediction accuracy} as:
\begin{align}
    A(\reader{}) = \frac{1}{\numparas} \sum_{i=1}^{\numparas} \one_{ [ \argmax\reader{}(i) = \trueculprit]}
\end{align}
This is the complement of the 0-1 loss \citep{devroye2013probabilistic}, averaged over story position.
We define the \defn{surprise score} for reader $\reader{}$ as $S(\reader{}) = 1 - A(\reader{})$, and the \defn{coherence score} for reader $\reader{}$ as $C(\reader{}) = A(\reader{})$.

We also define a \defn{uniform predictor} $\unifpredictor$ as a reference baseline that assigns uniform probability $1/|\culprits|$ to each suspect at every pre-revelation step; like all readers, it correctly identifies the culprit at and after the revelation.
It serves as a floor: a gullible reader performing no worse than $\unifpredictor$ (at pre-revelation steps) is not genuinely misled (no misdirection), and an intelligent reader performing no better than $\unifpredictor$ means the story contains no usable clues (not solvable).

A story is surprising when $S(\gulliblereader)$ is high; we write $\surprisalscore \equiv S(\gulliblereader)$ for the \defn{surprise score}, where a score close to $1$ indicates successful misdirection.

A story is coherent when $C(\reader{})$ is high for an intelligent reader: the clues should provide information allowing better identification of the culprit (see \hyperref[sec:clue_effectiveness]{Clue Effectiveness}).
Taking $\reader{} = \knowitallreader$---whose $\argmax$ prediction is Bayes-optimal---defines the \defn{coherence upper bound (CUB) score} $\coherencescore \equiv C(\knowitallreader)$, an upper bound (in expectation)on $C(\reader{})$ for any reader; we also define the \defn{average coherence score} $\overline{C} \equiv C(\areader{})$.
We note that a generation model with very low diversity can yield a misleadingly high $\coherencescore$ without genuine clue content, violating clue sufficiency.

\paragraph*{Fair play}
Surprise and coherence each measure a single reader.
Fair play captures whether clues systematically advantage a more capable reader over a less capable one---a property neither metric alone captures.
We define the \defn{fair play score} between a more capable reader $M$ and a less capable reader $M'$ as:
\begin{align}
    \text{FP}(M, M') = A(M) - A(M')
\end{align}

Specifically, we define the \defn{fair play upper bound} $\fairplayubscore \equiv \text{FP}(\knowitallreader,\gulliblereader) = \coherencescore - (1-\surprisalscore)$, using the know-it-all and gullible readers, and the \defn{actual fair play score} $\arfpscore \equiv \text{FP}(\areader{},\gulliblereader) = \overline{C} - (1-\surprisalscore)$, using the actual reader and the gullible reader.
We also define the \defn{solvability score} $\fairplayskepscore \equiv \text{FP}(\knowitallreader,\unifpredictor) = \coherencescore - A(\unifpredictor)$ using the know-it-all reader and the uniform predictor.

These scores define requirements for a fully fair-playing story:
\begin{enumerate}
    \item \textbf{Intelligence gap}: $\fairplayubscore, \arfpscore \geq \frac{1}{L}$ --- the know-it-all and actual readers outperform the gullible reader; the clues provide a genuine advantage for a capable reader over the gullible one.
    \item \textbf{Solvability}: $\fairplayskepscore \geq \frac{1}{L}$ --- the know-it-all outperforms the uniform predictor; the story contains genuine clues beyond chance. Failure is a sign of \emph{Deus ex Machina} (DeM): no clue signal beyond random guessing.
    \item \textbf{Misdirection}: $\text{FP}(\unifpredictor,\gulliblereader) \geq \frac{1}{L}$ --- the gullible reader performs worse than the uniform predictor; it is actively misled.
\end{enumerate}

For an alternative fair play metric measuring clue content (ERC)---with results broadly consistent with the main metrics---see SI Appendix. Experienced reader fair play analysis is also provided in the SI Appendix.

Together, these metrics allow us to test empirically whether fair play is a genuinely difficult balancing act---requiring simultaneous success on surprise, coherence, and their interaction---and whether failure on one dimension is invisible when evaluating the others in isolation.

%%----------------------------------------------------------------------
\section{Results} \label{sec: results}

We report empirical results on both LLM-generated and real detective stories, evaluating the relationship between surprise and coherence and the challenge of fair play.

\subsection{Generated Stories}

% \paragraph*{The coherence-surprise tradeoffs in practice}
A common assumption in NLP benchmarking is that model quality follows a limited set of latent factors \citep{maimon2025iqtestllmsevaluation}---predicting that larger, newer, and more reasoning-capable models will outperform weaker ones on most tasks. If surprise and coherence were two independent tasks, this would entail a positive correlation between them across stories. However, fair play requires balancing inherently competing demands, not optimizing them independently. The coherence-surprise tradeoff shows that improving one dimension tends to hurt the other when failing to create a sufficient intelligence gap. Moreover, the tradeoff challenges the ``leaderboard approach'' to model quality: LLMs may not follow the expected ranking when fair play depends on balancing, rather than excelling independently. 

\begin{figure*}[t]
\centering
% Left: human vs. LLM fair play scatter
\begin{minipage}[c]{0.3\linewidth}
\begin{subfigure}[t]{\linewidth}
\centering
\includegraphics[width=\linewidth]{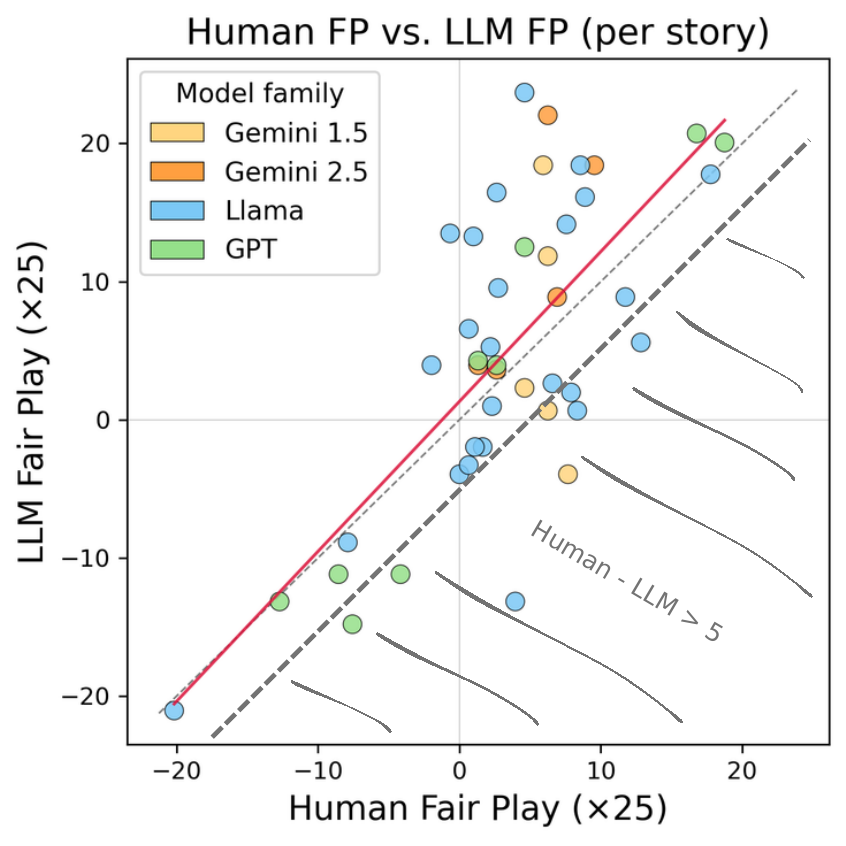}
\caption{LLM vs.\ human fair play}
\label{fig:human-vs-llm-fp}
\end{subfigure}
\end{minipage}
\hfill
% Right column: model ordering (large)
\begin{minipage}[c]{0.67\linewidth}
\begin{subfigure}[t]{\linewidth}
\centering
\includegraphics[width=\linewidth]{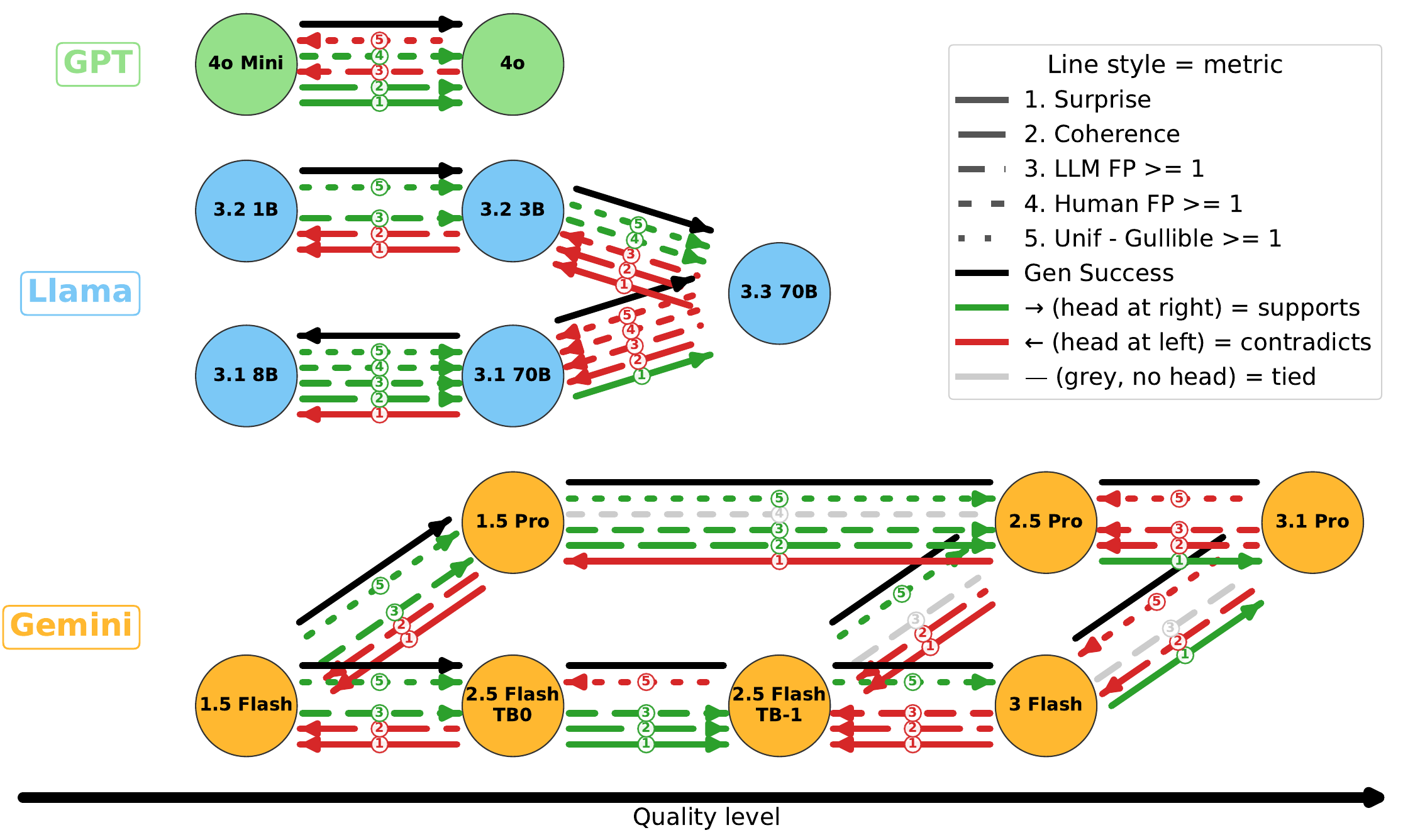}
\caption{Model ordering vs.\ expected quality ranking}
\label{fig:model-ordering}
\end{subfigure}
\end{minipage}
\caption{\textbf{(A)} Scatter plot of LLM-based fair play ($\fairplayubscore$) vs.\ human-based fair play ($\arfpscore$) per story for annotated stories ($n=54$). The marked region represents stories for which $\arfpscore - \fairplayubscore > \frac{5}{\numparas}$, and the sparse population there validates $\fairplayubscore$ as an upper bound.
\textbf{(B)} Hasse diagram of model performance ordering. Newer version, more parameters, more-thinking models are ranked higher in the \emph{expected} quality level (the x-axis); arrows show whether this expected ranking is \emph{confirmed} by actual scores, with one arrow for each measured score.}
\label{fig:scatter-results}
\end{figure*}

We test these predictions and present our results in Figure~\ref{fig:scatter-results}: the model ordering diagram (panel B) assesses the ranking.
Story validity broadly scales with model capability: most models achieve $\geq 80\%$ valid stories, and many frontier models achieve perfect validity ($10/10$).
A minor exception---Llama-3.1-8B ($9/10$) slightly exceeding Llama-3.1-70B ($10/12 \approx 0.83$). Full results reported in the SI Appendix.
These results suggest that story-generation capabilities scale with general model capabilities, as is generally the case with instruction following in LLMs.

Nevertheless, the Spearman correlation between surprise and coherence across generated stories is near zero ($r = -0.095$, $p > 0.05$, $n = 100$) for LLM-based coherence, and significantly negative ($r = -0.383$, $p < 0.05$, $n = 36$) for human-based coherence.
Indeed, Figure~\ref{fig:model-ordering} shows that LLMs do not follow the expected quality ordering: model rankings differ substantially across metrics, with no clear advantage to larger, newer, or more capable models across all dimensions simultaneously. 

\begin{figure*}[t]
\centering
\begin{subfigure}[t]{0.47\linewidth}
\centering
\includegraphics[width=\linewidth]{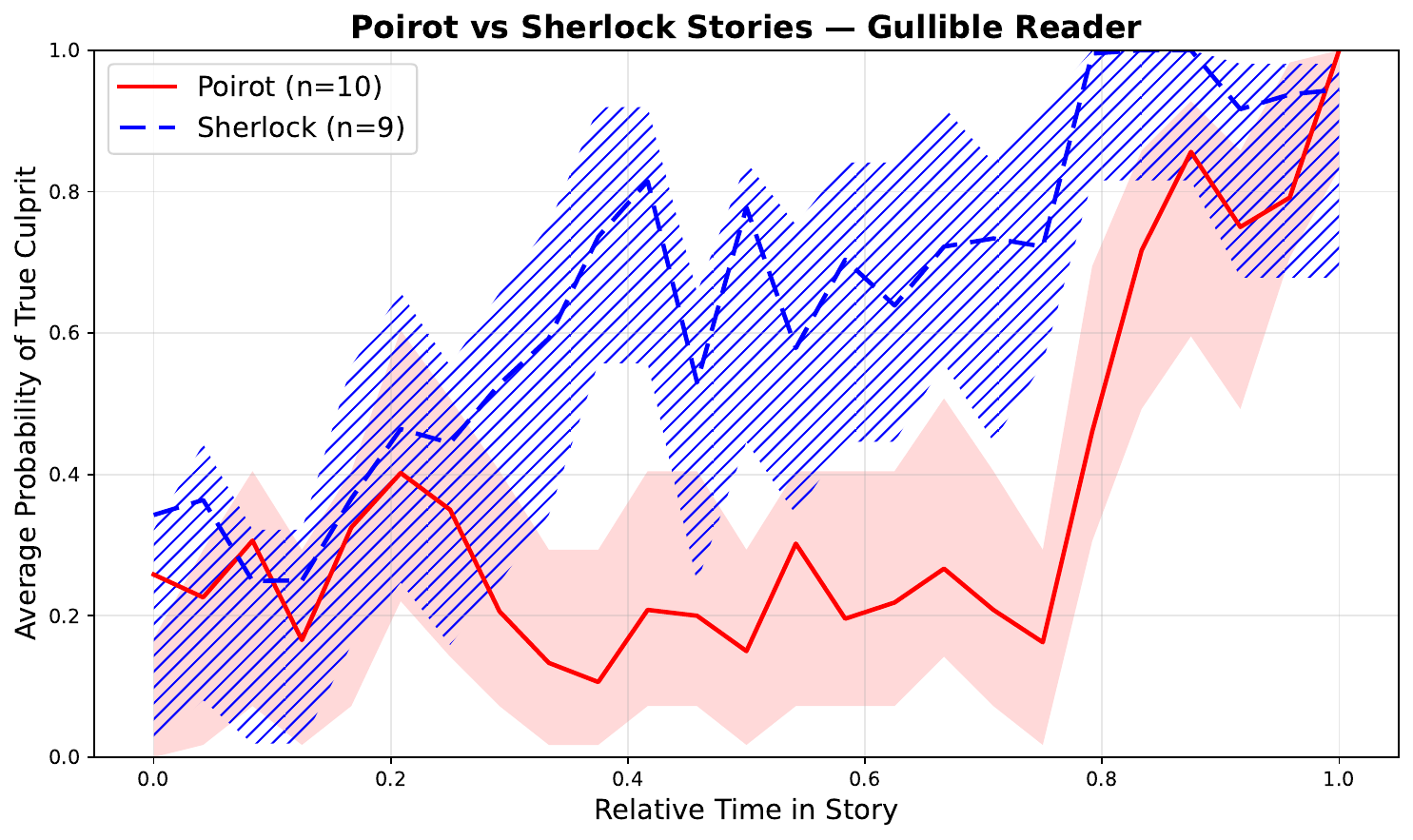}
\caption{$S_{\text{Poirot}} > S_{\text{Sherlock}}$}
\label{fig:surprisal-real-naive}
\end{subfigure}
\hfill
\begin{subfigure}[t]{0.47\linewidth}
\centering
\includegraphics[width=\linewidth]{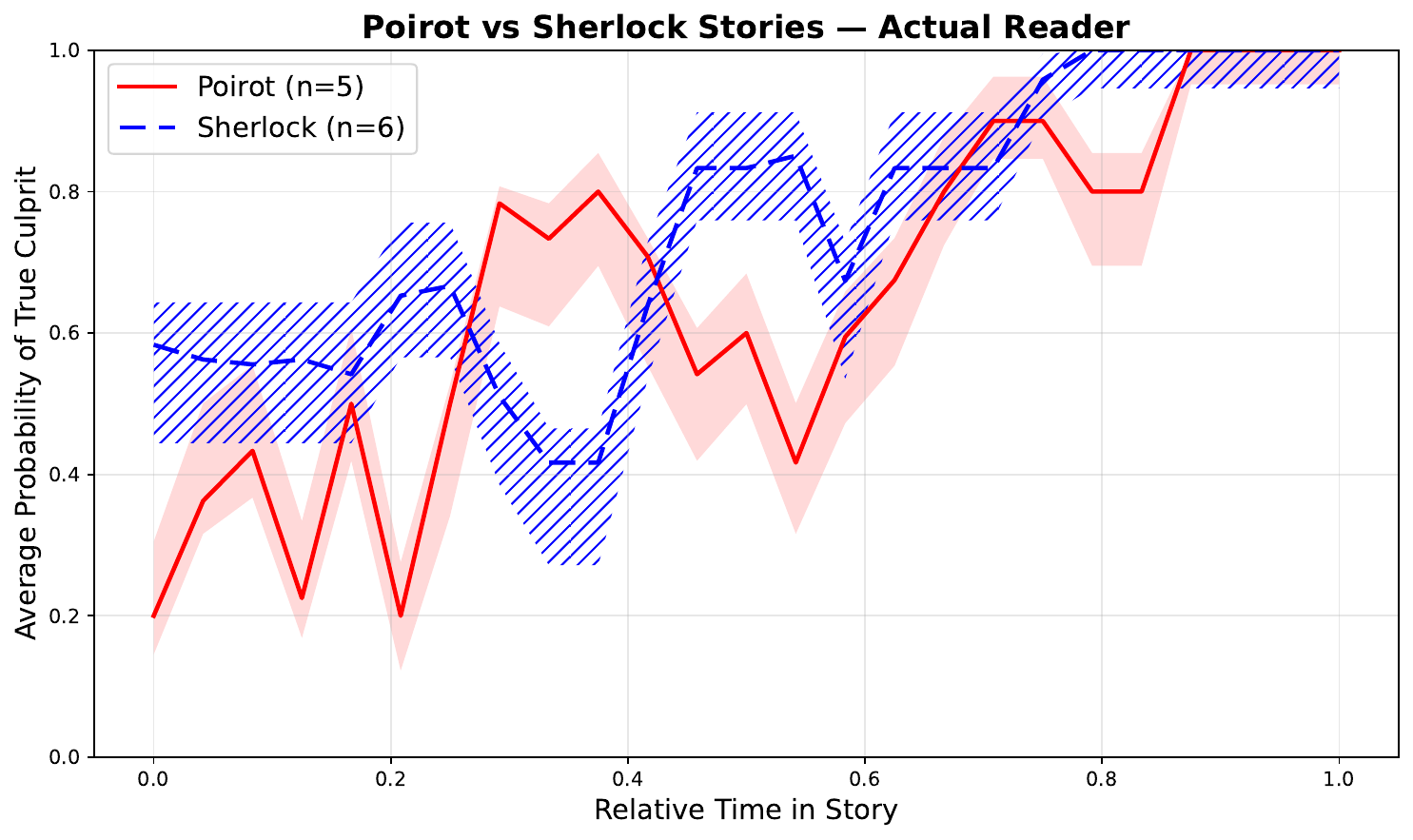}
\caption{(with \textbf{(a)})\: $\text{FP}_{\text{Poirot}} > \text{FP}_{\text{Sherlock}}$}
\label{fig:surprisal-real-actual}
\end{subfigure}
\\[2em]
\begin{subfigure}[t]{0.47\linewidth}
\centering
\includegraphics[width=\linewidth]{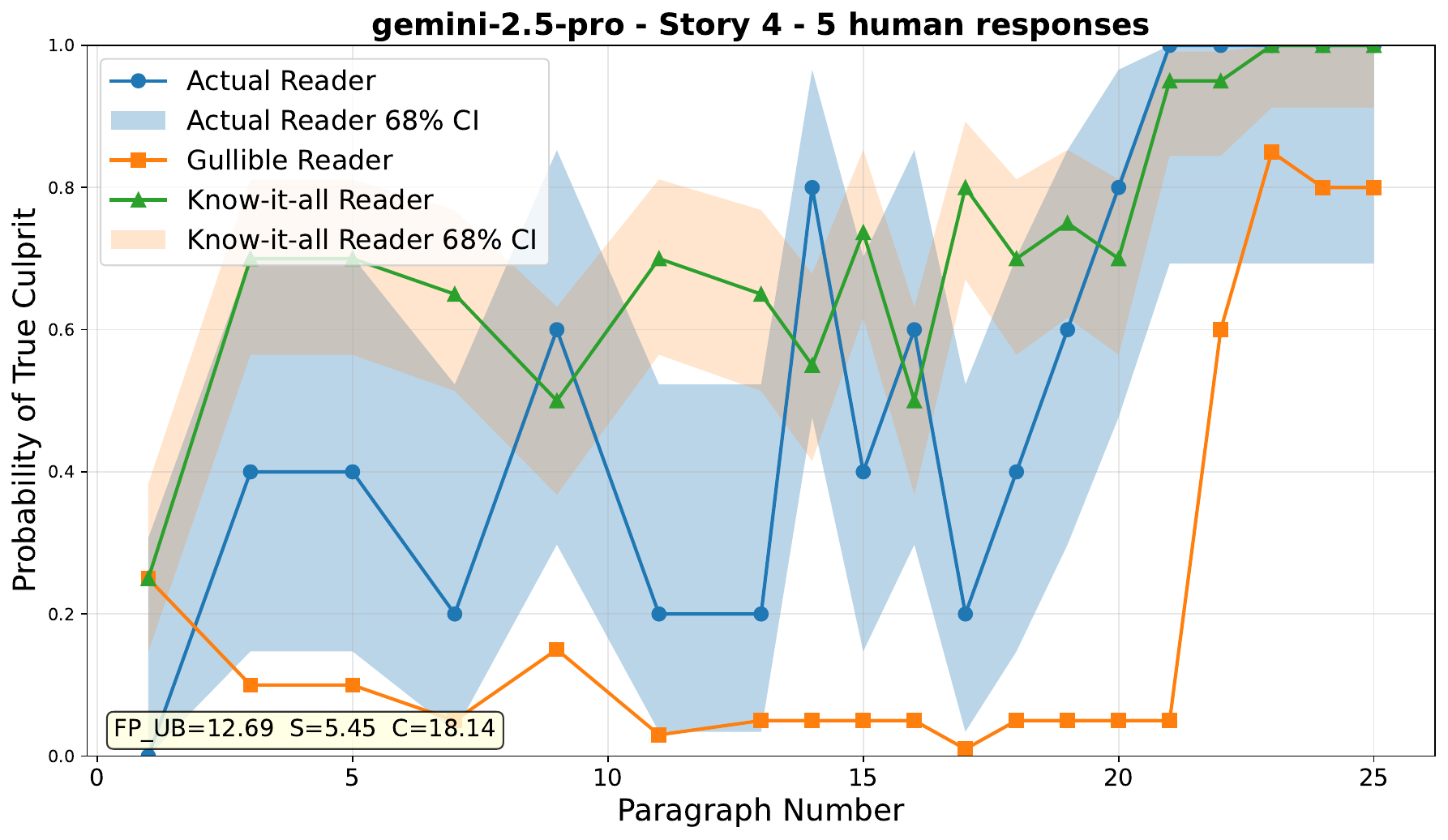}
\caption{Fair play example}
\label{fig:fp-example}
\end{subfigure}
\hfill
\begin{subfigure}[t]{0.47\linewidth}
\centering
\includegraphics[width=\linewidth]{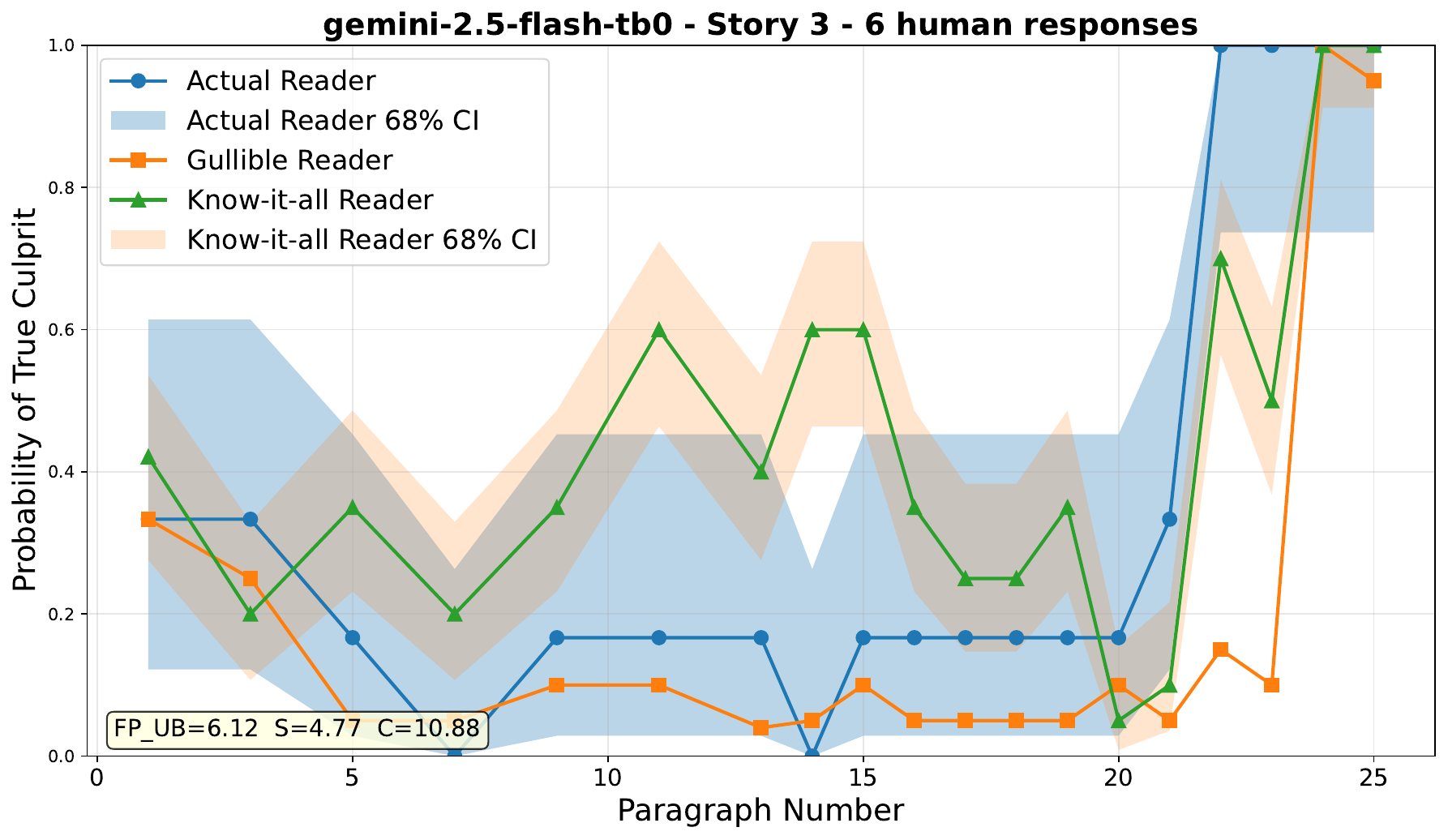}
\caption{Deus ex Machina example}
\label{fig:dem-example}
\end{subfigure}
\caption{\textbf{(A--B)} Reading curves for real detective stories.
\textbf{(A)} Average rate of correct gullible-reader predictions for Sherlock Holmes (blue dashed) and Hercule Poirot stories (red solid). Poirot stories consistently yield lower accuracy, reflecting stronger misdirection.
\textbf{(B)} Same for actual human readers, revealing the dissociation: Sherlock stories are more predictable to humans yet less surprising to the gullible reader.
\textbf{(C--D)} Reading curves for individual generated stories.
\textbf{(C)} A fair-play example (Gemini-2.5-pro, story~4): the know-it-all reader maintains a consistent advantage over the gullible reader, indicating usable clues embedded in the narrative.
\textbf{(D)} A Deus ex Machina example (Gemini-2.5-flash, story~3): the know-it-all reader gains no advantage prior to the revelation, indicating an absence of usable clues.
See SI Appendix for additional reading curve examples.}
\label{fig:reading-curves-main}
\end{figure*}

Generated stories serve as a diagnostic for the difficulty of the task itself: by probing models of varying capability, we can assess whether creating fair play poses a genuine challenge---even for contemporary LLMs---and whether surprise and coherence behave as competing demands.
The results confirm that fair play is challenging---many stories lack a sufficient intelligence gap ($\fairplayubscore < \frac{1}{\numparas}$; criterion~1 in Metrics) (e.g., $6/10$ for Llama-3.3-70B, $4/10$ for GPT-4o), showing that even frontier models often fail to produce stories where the clues meaningfully advantage an intelligent reader; individual failure cases are visible in Figure~\ref{fig:human-vs-llm-fp} as stories falling in the negative fair-play region, and representative reading curves for a fair-play and a Deus ex Machina story are shown in Figure~\ref{fig:reading-curves-main}C--D.
Surprise scores range from $0.352$ (Gemini-1.5-flash) to $0.798$ (Gemini-2.5-pro), and coherence upper bounds vary from $0.4$ to $0.8$, with no consistent relationship to model size or capability. Enabling extended reasoning modestly improves fair play for Gemini-2.5-flash. These results illustrate that fair play does not scale monotonically with model capability: the model ordering by fair play scores does not match the expected ordering of model capabilities (Figure~\ref{fig:model-ordering}).

\paragraph*{Human validation}
Figure~\ref{fig:human-vs-llm-fp} shows the relationship between LLM-based and human-based fair play across stories with human annotations. The $\fairplayubscore$ scores correlate positively with human-based fair play scores $\arfpscore$ across stories, validating the upper bound as a meaningful proxy for human-perceived fair play. Furthermore, the region where $\arfpscore$ substantially exceeds $\fairplayubscore$ is sparsely populated (Figure~\ref{fig:human-vs-llm-fp}), consistent with the theoretical upper-bound relationship.
On the human-annotated subset ($n=54$), the mean $\fairplayubscore = 0.242$ exceeds the mean $\arfpscore = 0.127$, consistent with the design of $\fairplayubscore$ as an upper bound. We note that, for some individual models, however, the human fair play score can exceed $\fairplayubscore$ due to the variance in the estimation or sampling.

Experienced reader results, which further probe clue sufficiency and story diversity, are reported in the SI Appendix. The experienced reader generally approaches the know-it-all upper bound as prior-story context increases; notably, much of this advantage already appears with zero prior stories, suggesting that familiarity with the model's generation process---rather than accumulation of story-specific knowledge---drives culprit prediction.

\subsection{Real Stories}

Our framework can be applied to any story: the gullible reader can be simulated by a language model, and the know-it-all reader can be approximated by a sufficiently strong prediction model.
For real stories, we do not have access to the true data-generating distribution, so we evaluate only the metrics that are reliable in this setting: gullible-reader surprise and human-based coherence and fair play.
Applying our framework to Sherlock Holmes and Hercule Poirot stories, we test three predictions grounded in established literary criticism \citep{Knight_2003}: (1) Poirot stories should achieve higher surprise scores, reflecting Christie's reputation for skillful misdirection in whodunits; (2) Poirot stories should achieve higher fair play scores, since coherence and surprise are more equally balanced in whodunits; and (3) Poirot's fair play should exceed that of generated stories on average, providing a human-authored benchmark.

\paragraph*{Surprise}
Figure~\ref{fig:reading-curves-main}A shows the average gullible-reader correct-prediction rate across stories, as a function of story position.
The mean gullible-reader surprise score for Hercule Poirot stories is $0.642$ and for Sherlock Holmes stories it is $0.342$.\footnote{Sample sizes: $n=10$ Poirot stories and $n=9$ Sherlock Holmes stories for gullible-reader analysis; see SI Appendix for human annotation sample sizes.}
Shaded margins represent $68\%$ Clopper-Pearson confidence intervals \citep{clopper1934use}---corresponding to approximately one standard deviation---computed
by pooling predictions across all stories of each type at each relative time point.
Poirot stories are consistently harder for the gullible reader to predict, reflecting Christie's reputation for skillful misdirection \citep{Knight_2003}. This aligns with the established literary contrast between Christie's whodunits and Conan Doyle's howdunits: in howdunit stories, the identity of the perpetrator may be known from the outset, and the mystery centers on the exact method or procedure of the crime---how was it committed?---rather than who did it. This genre difference predicts that Poirot stories (whodunits with identity misdirection) should yield systematically lower gullible-reader accuracy than Sherlock stories (howdunits where the perpetrator's identity is less central to the mystery), supporting the validity of the surprise metric.

\paragraph*{Human-Based Coherence, Fair Play, and Dissociation}
Figure~\ref{fig:reading-curves-main}B presents the average correct-prediction rate of actual human readers across the stories.
The two series reveal a dissociation: Sherlock stories achieve higher human-estimated coherence ($C = 0.751$) than Poirot stories ($C = 0.661$), yet the simulated gullible reader achieves higher accuracy on Sherlock stories ($S = 0.342$ vs.\ $S = 0.642$). Evaluating coherence alone would favor Sherlock, but this misses that Sherlock's clues are less effective at misdirecting a na\"ive reader.
The human fair play scores capture this dissociation: $0.325$ for Hercule Poirot vs.\ $0.067$ for Sherlock Holmes---Poirot stories are substantially more fair-playing despite Sherlock's higher coherence.\footnote{A notable exception is Sherlock story~5 (``The Man with the Twisted Lip''), where the gullible reader outperforms human participants. A possible reason is that human readers expect a twist and discount the obvious suspect, while the gullible reader takes the clues at face value---illustrating how prior genre experience can work \emph{against} accuracy when a story does not follow genre conventions.}
Using the human-based DeM criterion ($\arfpscore < 1/L$): $3/6$ Sherlock stories qualify as DeM, while none of the 5 Poirot stories with human annotations do ($0/5$), demonstrating that skilled human authors often more reliably achieve the coherence-surprise balance.
% Poirot's human fair play ($0.325$) exceeds the mean for generated stories ($0.127$), confirming the third prediction and demonstrating that skilled human authors more reliably achieve the coherence-surprise balance that the framework demands.

\subsection{Self-Reported Impressions}

Subjects were asked to rate various qualities of the story on a 1--5 scale (see SI Appendix for question details and correlation tables). Real stories were rated substantially higher than generated stories in fairness, coherence, and enjoyability, while
surprise does not clearly distinguish the two groups---possibly because subjective
surprise does not differentiate between stories that are surprising due to genuine
misdirection and those that are simply incoherent or poorly structured.
Subjective surprise is negatively correlated with subjective fairness ($r = -0.13$,
$p < 0.05$), reflecting the tension between being surprised and perceiving the story as
fair; in contrast, coherence and fairness are strongly correlated ($r = 0.65$, $p <
0.001$). Both patterns are consistent with the theoretical tradeoff.

In terms of the correlation of our metrics with subjective ratings, we find that coherence correlates with the measured coherence scores
(human readers: $r = 0.16$, $p < 0.01$; LLM readers: $r = 0.11$, $p < 0.05$).
Subjective fairness ratings correlate significantly with the human subjective fair play score ($r = 0.25$, $p < 0.001$) and with the LLM coherence score ($r = 0.11$, $p < 0.05$).
% Enjoyability correlates with the LLM coherence score ($r = 0.11$, $p < 0.05$). [Moved to SI]
The surprise score correlates significantly with subjective surprise ratings ($r = 0.15$, $p < 0.01$), validating that the gullible reader captures the narrative quality of misdirection as perceived by readers. Most absolute values of the correlations are expected: the $S$ and $C$ metrics capture specific, information-theoretic aspects of these qualities, while subjective ratings also reflect narrative dimensions such as writing quality and character depth.

% Many models display signs of ``mode collapse'' (e.g., Gemini-1.5-pro using the same culprit name in 90\% of stories). However, the true culprit's identity---which is the subject of our experiments---is close to uniformly distributed. Details are discussed in the SI Appendix.

%%----------------------------------------------------------------------
\section{Discussion and Conclusion} \label{sec:discussion}

% Our framework shows that achieving both surprise and coherence requires a \hyperref[sec:hindsight]{The Hindsight Gap}---a shift in the reader's interpretation after the revelation. 
% Our empirical results bear this out: Poirot stories---and to a lesser extent Sherlock---receive higher subjective ratings for fairness and coherence than generated stories, suggesting that skilled authors more reliably create this hindsight gap. 
% For generated stories, the gap between the fair play upper bound ($\fairplayubscore$) and actual reader fair play ($\arfpscore$) varies widely across models.
The central finding of this work is that fair play cannot be achieved by any single reader model, and requires creating a \hyperref[sec:hindsight]{Hindsight Gap}. Creating this gap is a two-objective challenge that, like constrained min-max optimization \citep{daskalakis2020complexityconstrainedminmaxoptimization}, is substantially more complex than optimizing either objective alone. This result clarifies why the task is inherently hard, and the empirical results bear this out.
For generated stories, variation across models highlights that fair play does not scale monotonically with model size; models differ in how they balance the competing demands of surprise and coherence, which are themselves empirically dissociated.

Human evaluations also paint a consistent picture: while surprise ratings are similar for real and generated stories,
subjective ratings of fairness, coherence, and enjoyability are substantially higher for
real than generated stories. 
We do note that even human-written stories are not always of uniformly high quality; yet consistently producing challenging but fair stories---for diverse audiences---is a central quality of detective story writing, achieved by many works \citep{sayers1936introduction}.

% The central finding of this work is theoretical: fair play cannot be achieved by any
% single reader model, and requires a hindsight gap---a shift in the reader's interpretive
% state at the revelation. This result clarifies why the task is inherently hard, not
% incidentally so, and the empirical results bear this out: many LLMs achieve surprise or
% coherence individually but fail to combine them.

Our experiments with human participants validate the proposed metrics, showing that the
automatic measures align with human behavior and judgments.
Since our metrics are reference-less---using LLMs to simulate different readers---they can
serve as automatic tools for evaluating and potentially improving fair play in story generation (as a training or inference signal). While story generation is a central
area in AI \citep{fang2023systematic}, its evaluation remains challenging
\citep{wang2025llmsgenerategoodstories}; our framework offers a step toward formal evaluation using information-theoretic measures.

Our theoretical analysis sheds light on the nature of detective fiction itself. The coherence-surprise tradeoff formalizes the challenge every mystery writer faces and shows that fair play necessarily requires distractors---characters or clues that mislead the reader before the revelation. The clue sufficiency property highlights the role of genre conventions in constraining what counts as a valid mystery, and the hindsight gap captures the cognitive shift that a well-crafted revelation demands from the reader. Our framework also demonstrates how such genre conventions can be cast in quantifiable terms, enabling large-scale literary analysis \citep{hatzel2023machine}.

At its core, the challenge of fair play is the challenge of simultaneously satisfying two readers with fundamentally different interpretive stances---a structure that recurs across many domains. In psycholinguistics, surprisal is defined as the negative log-probability assigned to each word \citep{hale-2001-probabilistic,levy2008expectation}. This quantity has been shown to correlate with behavioral and neural phenomena, such as reading time \citep{smith2008optimal} and event-related potentials \citep{frank2015erp}.
However, this single-pass measure may be insufficient when a later revelation reshapes the interpretation of earlier text---as in garden-path sentences, where readers must revise their syntactic commitments upon encountering an unexpected continuation, with surprisal systematically underestimating the resulting processing difficulty \citep{arehalli_syntactic_2022}.
This limitation of surprisal in the context of narrative-level reinterpretation falls out of the proposed framework. In this sense, our 2-dimensional coherence-surprise framework can be seen as an extension of surprisal theory. 

More broadly, generating stories that cater to multiple reader perspectives simultaneously is a Theory-of-Mind challenge \citep{huang2024understandingplanningllmagents,kosinski2024evaluating}: the author must model how different readers will interpret the same text, and engineer a specific shift in that interpretation at the revelation. Our framework provides a concrete testbed for these multi-perspective reasoning capabilities.
% Moreover, since our metrics are
% reference-less, they can potentially be leveraged by future work to enhance the fair play
% capabilities of language models, e.g., by utilizing policy gradient methods.

\subsection{Limitations}
Our study has several limitations. The human experiment involved a somewhat small participant pool
($n=20$), and fair play estimates for real stories rely on human predictions rather than
true authorial distributions; both limit the reliability of absolute score comparisons, so
results are best interpreted as relative rather than absolute measures. Our main fair play
metric is only an upper bound, and the gap to actual reader performance can reflect
factors beyond fair play, such as story diversity and LLM calibration.

The generated stories use a relatively short fixed length of 25 paragraphs, and fair play dynamics may differ in longer stories. This relates to our theoretical result that longer stories make surprise harder to sustain, suggesting the metrics may behave differently for longer stories.

Our framework assumes a single prominent distractor per story, and restricts the culprit to a known set of suspects introduced before the suspicion phase. Real detective fiction typically features multiple distractors with shifting suspicion, and the true culprit may be a character never initially considered a suspect, or the story may involve multiple culprits simultaneously. While these edge cases fall outside our current formalization, the single-culprit whodunit structure we model covers the canonical cases of the genre.
More broadly, the simplified generation setting---a mandatory single distractor in a short, structured story---may make the culprit easier to guess for a reader who knows the generation instructions, even without reading carefully. This appears to be reflected in our experienced reader experiments, where knowledge of the setting is sufficient for nearly perfect performance.
Mode collapse---when a model places much of its probability mass on a single story structure or culprit---can also
artificially inflate coherence scores; we generally do not find this in our generated
stories (see SI Appendix), but it remains a potential confound.

Finally, the framework is evaluated exclusively on whodunit stories. While a paradigmatic example of the surprise/coherence tension \citep[see][]{knox1929best, huhn1987detective}, results for whodunit stories may not generalize to
other mystery sub-genres such as thrillers, or procedural mysteries.

\section{Methods}
\subsection{Story Generation}
Stories were generated paragraph-by-paragraph---with $L=25$ paragraphs---using long-context LLMs prompted to write whodunit detective stories with a distractor, and with instructions to make the story logical after the revelation. Generation proceeded step-by-step to allow
controlled re-sampling from any intermediate checkpoint. For thinking models, the internal reasoning trace was not carried over between generation steps---each paragraph was generated independently, without access to the previous step's chain of thought. All generations used temperature~$= 1$.
Ten stories were generated per model. Story validity was assessed by judge models (o1-mini and o3-mini), requiring a decisive
($p > 0.5$) identification of both the true culprit and the distractor. See SI Appendix for the full prompts.

Generating models: Gemini-1.5-flash, Gemini-1.5-pro, Gemini-2.5-flash with no thinking
budget ($\text{TB}=0$) and with dynamic thinking ($\text{TB}=-1$); Gemini-2.5-pro, Gemini-3-flash, and Gemini-3.1-pro (with
dynamic thinking); Llama-3.1-8B-Instruct, Llama-3.1-70B-Instruct, Llama-3.2-1B-Instruct, Llama-3.2-3B-Instruct, and Llama-3.3-70B-Instruct; GPT-4o and GPT-4o-mini.
Model documentation: Gemini (\url{https://deepmind.google/models/gemini/}),
Llama (\url{https://www.llama.com/models/llama-3/}),
OpenAI (\url{https://platform.openai.com/docs/models}).

\subsection{Reader Estimation}
The gullible reader $\gulliblereader$ was estimated by prompting o3-mini to predict the
culprit at each paragraph as if the events in the story were real, explicitly ignoring the possibility
of misdirections. The know-it-all reader $\knowitallreader$ was estimated by sampling
5--20 continuations from the generating model at predetermined checkpoints (between paragraphs) and using relative culprit frequencies as probability estimates.

\subsection{Statistical Analysis}
Correlations between subjective ratings and objective measures were computed using
Spearman's rank correlation. Confidence intervals for prediction curves were computed
using the Clopper-Pearson method \citep{clopper1934use}, pooling predictions across stories at each relative time point.

\subsection{Human Experiment}
This study was approved by the Institutional Review Board (IRB) of the Hebrew University
of Jerusalem, Faculty of Medicine (protocol no.\ 10112025). All participants provided
informed consent prior to participation.
Twenty participants read stories paragraph by paragraph, guessing the culprit from a list
of four possible suspects after each paragraph. Participants also provided subjective ratings of
fairness, coherence, surprise, and enjoyability on a 1--5 scale after completing each story. Reading order was randomized and the generating model's identity was not revealed, to control for learning effects; no significant learning effect was found.
In total, $n=54$ generated stories and $n=11$ real stories ($n=6$ Sherlock Holmes and $n=5$ Hercule Poirot) received human annotations.

\subsection{Use of AI Tools}
Large language models (including Gemini 3 Flash, Claude Sonnet 4.5, and Claude Opus 4.6) were used to assist in brainstorming, writing, editing, and code development for this manuscript, all under the authors' explicit direction. All final decisions were made by the authors alone.

\paragraph*{Data Availability}
Code and story data will be made available at acceptance. Generated story
data, human annotations, and evaluation scripts are available from the authors upon
reasonable request.

\section*{Acknowledgments}
The authors would like to thank Dan Rockmore, Jeremy Manning, Gabriel Stanovsky,
Nitay Alon, Matanel Oren, and Nicole Gruber for their valuable feedback.
The authors also benefited from helpful feedback from the ILCC seminar at the University
of Edinburgh (including Mirella Lapata and Alex Lascarides) and the C.Psyd seminar at
Cornell (including Marten van Schijndel).
This research was supported by the Israeli Ministry of Science and Technology (Grant No.\ 3-17938), the Council for Higher Education, and the Israel Science Foundation (Grant No.\ 2912/25). Computational resources were provided in part through a Google Academic Research Credits award.

\bibliographystyle{plainnat}
\bibliography{references}

\appendix
\renewcommand{\thesection}{\Alph{section}}
\section{Appendix Organization}
\begin{itemize}
    \item \textbf{Background:}
    \begin{itemize}
        \item \hyperref[app:related_work]{Related Work} --- surveys prior work on surprise, coherence, and story evaluation.
        \item \hyperref[app:literary]{Literary Framework} --- discusses the literary conventions of detective fiction that motivate our framework.
    \end{itemize}
    \item \textbf{Formal Framework:}
    \begin{itemize}
        \item \hyperref[app:story_generation]{Story Generation Model and Optimality of Readers} --- provides the formal probabilistic framework, defines detective models and internal readers, and establishes the optimality characterizations of the know-it-all reader and brilliant detective. Includes formal definitions of the hindsight gap and hindsight coherence property (\hyperref[app:hindsight_formal]{Hindsight Gap Definitions}).
        \item \hyperref[app:proofs]{Proofs} --- contains all proofs of the theoretical results in the main text.
    \end{itemize}
    \item \textbf{Experimental Methods:}
    \begin{itemize}
        \item \hyperref[app:stories]{The Real Story Dataset} --- lists the stories used in experiments.
        \item \hyperref[app:generation_prompt]{Generation Prompts} --- gives the prompts used to generate stories and estimate reader models.
        \item \hyperref[app:gullible_reader_estimation]{Alternative Gullible Reader Estimation Methods} --- describes and evaluates alternative approaches for estimating the gullible reader.
        \item \hyperref[app:sampling_details]{Sampling Procedure for the Know-It-All Reader} --- details how the know-it-all reader is estimated via sampling.
        \item \hyperref[app:erc]{ERC Metric} --- defines the Expected Retrospective Coherence metric and reports results.
        \item \hyperref[app:experiment]{Human Experiment Details} --- describes the human study design, participants, and results.
        \item \hyperref[app:learning_effect]{Learning Effect Analysis} --- analyzes whether participants learned across stories in the human experiment.
    \end{itemize}
    \item \textbf{Results:}
    \begin{itemize}
        \item \hyperref[app:reading-curves]{Reading Curves} --- defines reading curves and shows examples.
        \item \hyperref[app:experienced-reader]{Experienced Reader: Full Results} --- reports full results for the experienced reader.
        \item \hyperref[app:results_tables]{Full Results Tables} --- provides complete numerical results for all models.
    \end{itemize}
    \item \textbf{Variability and Discussion:}
    \begin{itemize}
        \item \hyperref[app:statistics]{Variability Statistics} --- reports statistics on culprit diversity and story structure.
        \item \hyperref[app:discussion]{Additional Discussion} --- discusses mode collapse and related considerations on story diversity.
    \end{itemize}
\end{itemize}

\section{Related Work}
\label{app:related_work}

\subsection{Surprise and suspense in narratives}
Ely et~al.\ \citep{ely2015suspense} propose a formal framework for suspense and surprise in
entertainment, including mystery novels. Their model defines suspense as the variance of
outcomes under the reader's belief distribution, and surprise as the distance between the
prior belief and the realized outcome. A key assumption is a Martingale property: readers
form beliefs that are, on average, correct. Our framework relaxes this assumption and
specifically shows it is incompatible with fair play: a reader that satisfies the
Martingale property is essentially the know-it-all reader, and cannot simultaneously be
misled by a well-crafted mystery. More precisely, perfect probability estimation is a
property of the know-it-all reader or the brilliant-detective reader (when clue
sufficiency is fulfilled), and corresponds to coherence---which is incompatible with
surprise by the coherence-surprise theorem.

Wilmot and Keller \citep{wilmot-keller-2020-modelling} operationalize narrative suspense using language
model perplexity over story continuations, approximating the reader's uncertainty.
This captures the single-dimension view of surprise (as unpredictability) but does not
address retrospective coherence or the tension between the two qualities that our
framework formalizes.

\subsection{Coherence in narratives}
Maimon and Tsarfaty \citep{maimon-tsarfaty-2023-cohesentia} introduce CoheSentia, a dataset and evaluation
framework for narrative coherence, focusing on the internal consistency and causal
connectedness of story events. Mostafazadeh et~al.\ \citep{mostafazadeh-etal-2016-corpus} present the ROCStories
corpus and a commonsense story cloze task, measuring whether story endings are coherent
with the preceding context. These metrics capture important aspects of coherence but are
not designed to evaluate the specific coherence-of-revelation aspect central to detective
fiction: whether the clues scattered across the story retrospectively support the culprit
identity.

\subsection{Story generation and evaluation}
A substantial body of work addresses story generation \citep{fang2023systematic}, using
neural language models and more recently large language
models with planning \citep{ye2022neuralstoryplanning, xie-riedl-2024-creating}.
Evaluation of generated stories remains an open problem
\citep{wang2025llmsgenerategoodstories}: automatic metrics such as BLEU and ROUGE
correlate poorly with human judgments, and reference-based evaluation is limited by the
diversity of valid stories. Our framework contributes reference-less metrics grounded in
information-theoretic principles.

\subsection{Computational narrative analysis}

Computational approaches to narrative have grown substantially with the rise of large language models.
Piper et~al.\ \citep{piper-etal-2021-narrative} survey narratological theory and its relationship to NLP methods, arguing that connecting computational work to humanistic theory opens new empirical questions.
Piper et~al.\ \citep{piper-2023-computational} outline a research agenda for computational narrative understanding, emphasizing multi-modality, temporal structure, and socio-cultural schemas.
Hatzel et~al.\ \citep{hatzel2023machine} provide an overview of machine learning methods applied to computational literary studies.
Story generation is a central application area: Fang et~al.\ \citep{fang2023systematic} and Wang and Kreminski \citep{wang2025llmsgenerategoodstories} systematically evaluate LLMs for this task, highlighting the difficulty of producing narratively coherent and engaging text.
Schulz et~al.\ \citep{schulz2024narrativeinformationtheory} propose an information-theoretic framework for characterizing general narrative properties.
Acciai et~al.\ \citep{acciai2025narrative} apply the cognitive-psychology Narrative Coherence Coding Scheme \citep[NaCCS; ][]{reese2011coherence} to evaluate coherence of personal narratives generated by LLMs, and Nishigori and Takeda \citep{10.1145/3698061.3734393} evaluate coherence of stories generated in collaboration with LLMs.

\subsection{Detective fiction as an NLP benchmark}

Detective fiction has served as a natural testbed for NLP systems that require multi-step reasoning over long documents.
Frermann et~al.\ \citep{10.1162/tacl_a_00001} treat crime drama as an incremental natural language understanding challenge, modeling perpetrator identification as sequence labeling over multimodal evidence.
Ko{\v{c}}isk{\'y} et~al.\ \citep{kocisky-etal-2018-narrativeqa} include books from the mystery genre in NarrativeQA, a reading-comprehension challenge requiring integrative reasoning across full narratives.
Del and Fishel \citep{del-fishel-2023-true} introduce a deep abductive reasoning benchmark based on mystery narratives, showing that GPT-4 solves fewer than 40\% of puzzles---substantially below human performance.
Sprague et~al.\ \citep{sprague2024musrtestinglimitschainofthought} include murder mystery scenarios in MuSR, a multistep soft-reasoning benchmark that tests the limits of chain-of-thought prompting.
Ahuja et~al.\ \citep{ahuja2025findingflawedfictionsevaluating} evaluate LLMs on plot hole detection in fictional narratives, framing it as a complex reasoning task.
Unlike these works, which use detective stories primarily as evaluation instruments for reasoning, our framework focuses on the narrative quality of generated detective stories as a generative AI challenge.

de~Lima et~al.\ \citep{delima2025characterizinginvestigativemethodsfictional} study detective styles in
fictional texts using LLMs, characterizing the reasoning patterns of different fictional
detectives. Our detective model taxonomy (gullible vs.\ brilliant) draws on the same
intuitions but formalizes them within a probabilistic framework.

\subsection{Surprise metrics}
\label{sec:surprisal_metrics}

Surprisal theory \citep{hale-2001-probabilistic, levy2008expectation} models reading
difficulty as the negative log-probability of a word given its preceding context:
$-\log p(w_t \mid w_{1:t-1})$. Higher surprisal corresponds to greater processing
difficulty, a prediction supported by eye-tracking and self-paced reading experiments
\citep{smith2008optimal}.

Many narrative-level surprise metrics have been proposed. Ely et~al.\ \citep{ely2015suspense} formulate surprise as a change in belief. Wilmot and Keller \citep{wilmot-keller-2020-modelling} suggest an implementation using a full language modeling distribution, and Piper et~al.\ \citep{piper2023modeling} implement it (described as ``revelation'') with word-frequency language modeling. Schulz et~al.\ \citep{schulz2024narrativeinformationtheory} define a similar metric (``pivot'') for a general property of the story. In our work, we assess surprisal regarding the identity of the culprit.

A challenge with belief-change formulations is that they capture changes regardless of direction: a story with a distractor will score high both when belief shifts to the true culprit and when it moves further away from the truth.

Our framework uses a related but distinct quantity from word-level surprisal. We measure \emph{belief-level surprisal}: the cross-entropy of a reader's culprit distribution relative to the know-it-all reader. This operates at the discourse level (over paragraphs) rather than the word level, and captures the reader's predictive state about the underlying story rather than the surface text.

Formally, define the marginal surprisal of culprit
$\culprit$ at step $i$ for reader $\reader{}$ as
$-\log \reader{}(i)_{\culprit}$.
The expectation of this surprisal under the true story distribution is the cross-entropy
$\centropy(\knowitallreader(i);\, \reader{}(i))$, which is our measure of uninformedness.
When $\reader{} = \knowitallreader$, this reduces to the entropy of the true distribution,
which equals the minimum achievable surprisal.

The \emph{Uniform Information Density} (UID) hypothesis \citep{jaeger2010redundancy}
predicts that communicators regulate information density to remain within comprehension
limits. In the context of detective fiction, UID would predict that clues should be
spread evenly across the story, with each paragraph contributing roughly equal information
about the culprit. Our step effectiveness measure (\clueeffectiveness) quantifies exactly
this per-paragraph information contribution, and our framework shows that the requirement
to simultaneously mislead a gullible reader and inform a brilliant one creates systematic
deviations from UID.

\subsection{Coherence metrics}

Wilmot and Keller \citep{wilmot-keller-2020-modelling} define surprisal as entropy reduction.
Although proposed as a surprisal metric, entropy reduction is problematic when a distractor is involved, since a false belief also has low entropy.
However, from the keen reader's view, entropy reduction can be seen as a coherence metric. For example, one can measure the Spearman correlation between the location in the story and the entropy (so monotonic reduction will give $-1$).
In practice, like other information metrics, this method is challenging to compute, especially when using sampling, since a small number of samples will limit the possible values and thus limit the possible decreasing steps.

A numerical metric that fully captures coherence in general is hard to design.
In some synthetic cases, strict logical consistency can be tested \citep{chieppe2022bayesian}.

%%----------------------------------------------------------------------
\section{Literary Framework}
\label{app:literary}
This section provides a detailed literary background for the framework introduced in the main text, with extended references to the narratological and literary scholarship underlying our definitions.

Literary scholars have offered useful analyses of the conventions of detective stories
\citep{todorov_1988}, with many attempting to identify and model what makes detective
stories ``good'' or successful \citep[e.g.,][]{bloch1980philosophical}.

\subsection{Structure of detective stories} \label{sec:conventions}
Classical detective novels follow a remarkably consistent pattern in their narrative
structure \citep{huhn1987detective, Knight_2003}, as evinced by various lists of
commandments and golden rules for detective fiction \citep[e.g., ][]{knox1929best}.
Consider the typical closed-room mystery: a crime occurs, often murder, and the central
puzzle concerns identifying who committed it. This type of narrative, known as a
``whodunit'' \defn{story}, revolves around discovering the identity of the \defn{culprit}
\citep{Knight_2003}.

The author typically introduces complications through carefully selected characters
designed to build the tension, while possibly misleading the reader.
These \defn{distractors} serve as red herrings, drawing suspicion away from the true
perpetrator until the final revelation. The narrative unfolds in three distinct phases
that correspond to how a typical reader experiences the mystery. During the
\defn{introduction}, both the crime and potential suspects are presented, but readers are
not yet expected to reach conclusions. The \defn{suspicion} phase follows, where the
author deliberately guides readers toward believing a distractor is the
culprit.\footnote{A story may contain many distractors. We focus on a common case where
there is a prominent distractor that suspicion is cast on, and whose innocence is proven
only together with the final revelation.} Finally, the \defn{revelation} phase unveils
the true culprit to the unsuspecting characters and reader \citep{huhn1987detective}.

\subsection{Clues and detectives}
Central to the functioning of crime mysteries are the \defn{clues} that appear throughout
the investigation \citep{huhn1987detective, moretti2000slaughterhouse}. These details,
scattered across the narrative, should remain incompletely interpreted until the story's
resolution. The revelation typically provides a full interpretation of the clues,
revealing for each clue its causal relation to the underlying crime.

A recurring motif in crime fiction is the contrast between the \defn{brilliant detective}
and the \defn{gullible detective}---whether this latter figure manifests itself as a
sidekick or represents an entire police department \citep{ketovic2019detective}. Gullible
detectives tend to na\"ively form conclusions based on a set of simply interpreted clues,
whereas the brilliant detective possesses the capacity to solve the mystery through an
explanation that accounts for all available evidence. This contrast emphasizes the
brilliance of the detective \citep{roth1995foul}.

Examples of brilliant detectives are Conan-Doyle's Sherlock Holmes and Christie's Hercule
Poirot. In these cases, both the sidekicks (i.e., Watson and Hastings) and the police
act as gullible detectives.

\subsection{Dual narratives: crime and investigation}\label{sec:crime_investigation}
Literary studies view detective fiction as operating on two interconnected levels.
The \defn{crime} level represents the true sequence of events that the narrative seeks to
uncover---this is the actual story of what happened, the reason for the mystery, and
includes the identity of the real culprit \citep{bloch1980philosophical}.
Running parallel to this hidden truth is the \defn{investigation} level, which constitutes
the told story wherein an investigative process attempts to solve the mystery
\citep{todorov_1988, huhn1987detective,bloch1980philosophical}.

These two levels are conveyed via two forms of discourse---the crime level unfolds
through the detective's investigation, and the investigation level is told to the reader
by the narrator \citep{huhn1987detective}. The first form is \defn{internal}, revealing
the crime to the characters, and the second form is \defn{external}, aimed at the reader
outside the story.

For example, in Sherlock Holmes stories, the story of the crime unfolds by Holmes'
deductions and discoveries, and becomes apparent to the characters in the story. The
story of the investigation is told by Watson, who is the narrator, and is directed to
the reader.

\subsection{Reader's role} \label{sec:readers_role}
The reader's role is central in narratological theory, recognizing the implied reader as
the image of the recipient that the author envisioned while crafting the work
\citep{booth1983rhetoric, schmid2014implied}.
In detective stories, the reader's role takes on special significance because readers are
often assumed to ``participate'' actively in the investigation process
\citep{huhn1987detective,dove1997reader, saunders2019we, link2023dossier}. Moreover, in
many stories, the detective's sidekick clearly states that he is writing the story, so
the story itself addresses the reader \citep{todorov_1988}. This instigates an
``internal'' form of participation, where the reader is invited to participate in the
challenge, which the gullible detectives fail at.
Some stories involve multiple implied readers, as commonly seen in children's literature
designed to be understood differently by young readers and adults
\citep{richardson2007singular}. Specifically, children's detective stories (e.g., Nancy
Drew, Encyclopedia Brown) provide clearly solvable puzzles (e.g., by grown-ups) that are
challenging for the intended age-group \citep{veldhuizen2023children}.
Similarly, detective stories can be read through the eyes of multiple investigators with
different competence levels, with the brilliant detective as the grown-up who solves the
mystery for the child.
In \textit{The Murder of Roger Ackroyd}, this distinction is especially sharp: the gullible
reader trusts the narrator and progressively builds confidence in the wrong suspect, while
the brilliant reader notices subtle evasions in the narrator's account that point toward
his guilt.

\subsection{Coherence and surprise}
Successful detective fiction maintains both coherence and surprise in the eye of the
reader. Narrative \defn{coherence}, in general, describes the quality of a narrative as
a unified structure \citep{baerger1999life, cevasco2023construction}.
In detective fiction, this type of coherence has special importance whereby the sequence
of clues in the story, together with the conclusion, ``make sense'' when viewed in
retrospect \citep{10.1111/j.1540-6245.2011.01470.x}.
``Whodunit'' stories are also arranged to evoke a \defn{surprise}. Memorable detective
stories achieve both qualities simultaneously---the ending shocks the reader while also
making perfect sense upon reflection \citep{singer1884thewhodunit, Knight_2003}.

\subsection{Fair play}
Many writers attempted to define the implicit rules of detective fiction
\citep{knox1929best}.
A core component in these rules is the concept of \defn{fair play}
\citep{9d3be9cd-23e5-3c70-96dc-9acc8c696b2c,Knight_2003,Effron_2017,link2023defining}.
A story achieves fair play when a reader possesses the theoretical ability to solve the
mystery without requiring an explicit solution. Fair play also implicitly assumes that a
challenge is posed and the solution is nontrivial, thus worthy of playing.

In contrast stands a \defn{Deus ex Machina} (DeM) story, where clues contribute nothing
meaningful to the eventual resolution \citep{sallas2024logic}. In detective stories, this
manifests when the investigation process throughout the narrative has minimal impact on
uncovering the mystery's solution.
Such stories generally receive criticism as being of poor quality, as they violate an
implicit convention of the genre \citep{Effron_2017}.
Rules such as fair play are ``external'' to the story as they form an implicit contract
between the actual writer and reader, regardless of the detective's reasoning process.

Simple examples of fair play are in children's detective stories, such as Nancy Drew
and Encyclopedia Brown.
DeM detective stories can be cases where conclusive evidence, like fingerprints or CCTV
footage, is suddenly discovered, with no connection to previous clues. A famous example,
not from detective fiction, is in \textit{The War of the Worlds} \citep{wells1898war}.

\subsection{The challenge of fair play}
Achieving fair play is arguably the most prominent challenge in creating a good detective
story. While fair play requires the appropriate clues to be present, it does not require
them to be obvious \citep{dove1997reader}. On the contrary, a fair play story must
mislead the reader through manipulation and misdirection
\citep{sayers1936aristotle,emmott2018reliability}, and the revelation forces the reader
to retrospectively question assumptions they made \citep{tobin2009cognitive}.

Moreover, the precise nature of fair play remains debated. Roth \citep{roth1995foul} argues
that since the writer has predetermined the outcome, fair play is ultimately an illusion
rather than a true intellectual challenge. Others point to hindsight bias, which can make
any resolution seem retrospectively inevitable \citep{tobin2018elements}, or to the role
of tacit, non-deductive reasoning that resists formalization \citep{keller1990detective}.
These critiques highlight a fundamental ambiguity: the boundary between genuine
solvability and its mere appearance is difficult to draw in informal terms. The
probabilistic framework of the main paper addresses this challenge and formalizes these
notions.

%%----------------------------------------------------------------------
\section{Story Generation Model and Optimality of Readers}
\label{app:story_generation}

We define the story generation model and use it to formally characterize the optimality
of certain reader and detective models.

\subsection{Story and Clue Notation}

The investigation level is modeled as a sequence of propositions, or \defn{clues},
$\clues$, that are gradually revealed across the story. For example, if a bloodstain was found on the kitchen
wall, the clue is a statement to this effect---an undisputed fact not explicitly included
in the underlying crime story. We denote the set of all possible
clue sequences with $\cluesdomain$.

The clues are described in the paragraphs of the story $\story$. We use $\cluesi{1}{i}$
to denote all the clues revealed in the story prefix $\storyi{1}{i}$. Without loss of
generality, we assume that each paragraph contains at most one clue. We assume that clues
appear in the story in the order in which they are revealed to the detective. Information
available to the detective but not explicitly described in the text is not considered a
clue: for example, if Poirot goes out and finds some evidence that is not shared by the
narrator at the time, it will not be included in the set of clues.

\subsection{Two Interacting Generative Levels}

We model two generative levels in a story, corresponding to the dual narratives of crime
and investigation (see the Literary Framework). Since the
story text contains the clues, the external level (=the investigation story) induces the internal level (=the crime story).

For RVs $A,B$, we use the notation $A \vdash B$ to mean that $B$ is inferrable from
$A$---i.e., $B = f(A)$ for some deterministic function $f$---where $A$ and $B$ may be properties of either the external or internal level.

At the \defn{external} level, similar to traditional language modeling, we view a story
as a sample from some underlying distribution over finite strings
\citep{shannon1948mathematical, bahl1983maximum}. The story $\storyrv$ is generated
auto-regressively by a story generation model $\storymodel$; the culprit identity
$\culpritrv$ is deterministically inferred from the completed story:
\begin{align}
    p_{\storymodel}(\story, \culprit) = \prod_{i=1}^{\numparas}
    p_{\storymodel}(\para_{i} \mid \storyi{1}{i-1}) \cdot
    \one_{\story \vdash \hspace{0.25em}\culpritrv=\culprit}
\end{align}
A specific story generation model incorporates assumptions as to what writing style is
favored or disfavored, but can also include preferences that may affect the generation at
any step.\footnote{We emphasize that a story generation model is not necessarily a
simulation of a single human writer but rather any process that generates written stories,
possibly through a mixture or even by fixed rules. In practice the model may be
conditioned on additional information (e.g., the intended culprit identity), which can be
captured as additional context in $p_{\storymodel}(\para_i \mid \storyi{1}{i-1})$.}

At the \defn{internal} level, the story induces a distribution over clue sequences. We
denote this distribution $p_{\text{CLUES}}$; the culprit is deterministically inferred
from the complete clue sequence:
\begin{align}\label{eq:p_clues}
    p_{\text{CLUES}}(\clues, \culprit) = \prod_{i=1}^{\numparas}
    p_{\text{CLUES}}(\clue_i \mid \cluesi{1}{i-1}) \cdot
    \one_{\clues \vdash \hspace{0.25em}\culpritrv=\culprit}
\end{align}
The two levels are connected: the full story $\story$ contains the full clue sequence
$\cluesrv$, so $\story \vdash \cluesrv$, and together with the \defn{fully-contained}
assumption ($\cluesrv \vdash \culpritrv$, i.e., the clues include sufficient evidence to
identify the culprit---for example, an explicit confession or an unassailable indictment)
we have $\story \vdash \culpritrv$ for a complete story.

\subsection{Detective Models and Internal Readers}

Detectives observe clues rather than the full story text.
Formally, a \defn{detective model} is a function $\detective{}: \cluesdomain \times \mathbb{N} \to \Delta^{|\culprits|-1}$, mapping a clue sequence and a step to a probability distribution over suspects.

The \defn{gullible detective} $\detective{0}$ forms predictions based on a na\"ive interpretation of the clues, in the extreme case basing its prediction on a single prominent clue and ignoring the rest.
The \defn{brilliant detective} $\detective{1}$ is the Bayes-optimal detective (defined in the next subsection).

These detective models induce corresponding reader models:
$\gulliblereader(\storyi{1}{i}) \coloneqq \detective{0}(\cluesi{1}{i})$ and $\brilliantreader(\storyi{1}{i}) \coloneqq \detective{1}(\cluesi{1}{i})$.

\subsection{Optimality of Readers and Detectives}

Every reader and detective model estimates the conditional distribution of the culprit
given their observations: external readers condition on the story prefix $\storyi{1}{i}$,
while detectives condition on the clue sequence $\cluesi{1}{i}$. In both cases, the
optimal estimate is the true conditional, obtained by marginalizing the joint distribution
over the unobserved continuation.

For an external reader with access to the true story generation distribution
$p_{\storymodel}$, the optimal estimate marginalizes over all possible story continuations:
\begin{align}
    \knowitallreader(i)_{\culprit} = p_{\storymodel}(\culpritrv=\culprit \mid \storyi{1}{i}) = \sum_{\storyi{i+1}{\numparas}} p_{\storymodel}(\storyi{i+1}{\numparas} \mid \storyi{1}{i}) \cdot \one_{\story \vdash \hspace{0.25em}\culpritrv=\culprit}
\end{align}
This is the know-it-all reader; its classifier $\hat{y}_{\knowitallreader}(i)$ is the
Bayes-optimal classifier for the story generation distribution. Any other external reader
uses an estimated conditional $\hat{p}(\culpritrv \mid \storyi{1}{i})$ in place of the
true one.

Symmetrically, for a detective with access to the true clue generation distribution
$p_{\text{CLUES}}$, the optimal estimate marginalizes over unobserved future clues:
$\detective{1}(i)_y = p_{\text{CLUES}}(\culpritrv =y\mid \cluesi{1}{i})$. This is the
brilliant detective. We adopt the \defn{brilliance assumption}: the clue distribution
$p_{\text{CLUES}}$ induced by the story closely matches the distribution a real investigator would encounter (i.e., the detective's inference from the clue sequence is approximately correct). Under this assumption, the brilliant detective's inference is also approximately
optimal for the story.

%%----------------------------------------------------------------------
\section{Proofs}
\label{app:proofs}

\subsection{Similar Culprit Probabilities Imply Intelligence}
\label{app:proof_clueprobs}
\label{app:proof_clueprobs_detailed}

Recall that a reader $\reader{}$ is called \emph{intelligent} if its intelligence gap $\intelligencegap{i}{\reader{}}$ is small---i.e., its clue effectiveness approaches that of the brilliant-detective reader (see main text, Eq.~9).
We show that if a reader assigns culprit log-probabilities close to those of the brilliant-detective reader at every step, then it is intelligent.

\begin{lemma*}
Let $\reader{}$ be an internal reader (i.e., induced by some detective model $\detective{}$). If for all steps $j \leq i+1$ and all culprits $\culprit$:
\begin{align}
    \big| \log p_{\reader{}}(\culprit \mid \cluesi{1}{j}) - \log p_{\brilliantreader}(\culprit \mid \cluesi{1}{j}) \big| \leq \epsilon_{\textnormal{log}}
\end{align}
then the difference in step effectiveness is bounded by:
\begin{align}
    \big|\clueeffectiveness_{\reader{}}(i+1) - \clueeffectiveness_{\brilliantreader}(i+1)\big| \leq 2\epsilon_{\textnormal{log}}
\end{align}
and consequently $\intelligencegap{i}{\reader{}} \leq 2i \cdot \epsilon_{\textnormal{log}}$.
\end{lemma*}

\begin{proof}
Expanding step effectiveness into cross-entropies:
\begin{align}
    &\clueeffectiveness_{\reader{}}(i+1) = \centropy(\knowitallreader(i);\, \reader{}(i)) - \E[\centropy(\knowitallreader(i+1);\, \reader{}(i+1))] \nonumber\\
    &= \sum_{\culprit} p_{\knowitallreader}(\culprit \mid \storyi{1}{i}) \log p_{\reader{}}(\culprit \mid \storyi{1}{i})- \E\left[\sum_{\culprit} p_{\knowitallreader}(\culprit \mid \storyi{1}{i+1}) \log p_{\reader{}}(\culprit \mid \storyi{1}{i+1})\right]
\end{align}

Since $\reader{}$ is induced by $\detective{}$, we can write $p_{\reader{}}(\culprit \mid \storyi{1}{j}) = p_{\detective{}}(\culprit \mid \cluesi{1}{j})$.
Each cross-entropy term is a weighted average of log-probabilities, where the weights (the know-it-all reader's probabilities) sum to one. Since each log-probability is within $\epsilon_{\text{log}}$ of the corresponding brilliant-detective value, the weighted average inherits the same bound:
\begin{align}
    &\left|\sum_{\culprit} p_{\knowitallreader}(\culprit \mid \storyi{1}{i}) \log p_{\detective{}}(\culprit \mid \cluesi{1}{i}) - \sum_{\culprit} p_{\knowitallreader}(\culprit \mid \storyi{1}{i}) \log p_{\brilliantreader}(\culprit \mid \cluesi{1}{i})\right| \nonumber\\
    &\leq \sum_{\culprit} p_{\knowitallreader}(\culprit \mid \storyi{1}{i}) \big|\log p_{\detective{}}(\culprit \mid \cluesi{1}{i}) - \log p_{\brilliantreader}(\culprit \mid \cluesi{1}{i}) \big|
    \leq \epsilon_{\text{log}}
\end{align}
The same bound holds at step $i+1$ and under the expectation over the next paragraph. Since step effectiveness is a difference of two such cross-entropy terms, the triangle inequality gives:
\begin{align}
    \big|\clueeffectiveness_{\reader{}}(i+1) - \clueeffectiveness_{\brilliantreader}(i+1)\big| \leq \epsilon_{\text{log}} + \epsilon_{\text{log}} = 2\epsilon_{\text{log}}
\end{align}
The intelligence gap bound follows by summing over $i$ steps.
\end{proof}

The converse does not necessarily hold: a reader may deviate from $\brilliantreader$ at individual steps yet still accumulate similar total effectiveness, resulting in a small intelligence gap without pointwise closeness.

\subsection{Clue sufficiency via conditional clue generation}

We show a second sufficient condition for clue sufficiency: if a detective's generative model for the clue sequence is close to the clue distribution induced by the story model, then clue sufficiency holds. Recall that the story model $\storymodel$ induces a distribution over clue sequences $p_{\text{CLUES}}$ (Eq.~\ref{eq:p_clues}). We write $p_{\storymodel}(\clue_j \mid \cluesi{1}{j-1})$ as shorthand for the conditional clue distribution induced by $\storymodel$, and similarly we extend the detective model $\detective{}$---originally defined as a predictor of culprit probabilities---to also have a generative model over clue sequences.

\begin{proposition*}
    Assume a detective $\detective{}$ has a generative model for clues that is close to the one induced by the story model. That is, for all steps $j$ and all clue values:
    \begin{align}
        \big| \log p_{\detective{}}(\clue_j \mid \cluesi{1}{j-1}) - \log p_{\storymodel}(\clue_j \mid \cluesi{1}{j-1}) \big| \leq \epsilon_{\textnormal{gen}}
    \end{align}
    and assume that the true culprit is determined by the complete clue sequence (i.e., $\clues \vdash \culpritrv$).

    Then clue sufficiency holds for the internal reader $\reader{}$ induced by $\detective{}$:
    \begin{align}
        |\clueeffectiveness_{\reader{}}(i) - \clueeffectiveness_{\knowitallreader}(i)| \leq 2(\numparas - i)\epsilon_{\textnormal{gen}}
    \end{align}
\end{proposition*}

\begin{proof}
    The detective $\detective{}$ predicts culprit probabilities by marginalizing over future clue sequences under its generative model:
    \begin{align}
        p_{\detective{}}(\culprit \mid \cluesi{1}{j}) = \sum_{\cluesi{j+1}{\numparas}:\, \clues \vdash \culpritrv = \culprit} p_{\detective{}}(\cluesi{j+1}{\numparas} \mid \cluesi{1}{j})
    \end{align}
    The know-it-all reader computes the same quantity using the true clue distribution $p_{\storymodel}$ instead of $p_{\detective{}}$.

    By the chain rule, the joint probability of a clue continuation $\cluesi{j+1}{\numparas}$ is a product of $\numparas - j$ conditional terms, each of which is within $\epsilon_{\textnormal{gen}}$ in log-space. Summing the per-step bounds:
    \begin{align}
        \big| \log p_{\detective{}}(\cluesi{j+1}{\numparas} \mid \cluesi{1}{j}) - \log p_{\storymodel}(\cluesi{j+1}{\numparas} \mid \cluesi{1}{j}) \big| \leq (\numparas - j) \cdot \epsilon_{\textnormal{gen}}
    \end{align}
    To pass from this bound on continuation probabilities to a bound on posterior culprit probabilities, note that each continuation's probability under $\detective{}$ is within a multiplicative factor of $e^{\pm(\numparas-j)\epsilon_{\textnormal{gen}}}$ of its probability under $\storymodel$. Since the posterior $p_{\detective{}}(\culprit \mid \cluesi{1}{j})$ is a sum of such terms (over all $\culprit$-supporting continuations), and the same multiplicative factor applies to every term, the ratio of the two posteriors inherits the same bound:
    \begin{align}
        \big| \log p_{\detective{}}(\culprit \mid \cluesi{1}{j}) - \log p_{\knowitallreader}(\culprit \mid \cluesi{1}{j}) \big| \leq (\numparas - j) \cdot \epsilon_{\textnormal{gen}}
    \end{align}
    Since the induced reader satisfies $\reader{}(j) = \detective{}(j)$ by definition, the cross-entropy term at step $i$ satisfies the same weighted-average argument as in the previous lemma, with the log-probability bound $(\numparas - i) \cdot \epsilon_{\textnormal{gen}}$ at that step. Similarly, the cross-entropy term at step $i+1$ is bounded using $(\numparas - i - 1) \cdot \epsilon_{\textnormal{gen}}$. Since step effectiveness is a difference of two such cross-entropy terms, the triangle inequality gives:
    \begin{align}
        \big|\clueeffectiveness_{\reader{}}(i+1) - \clueeffectiveness_{\knowitallreader}(i+1)\big| \leq (\numparas - i)\epsilon_{\textnormal{gen}} + (\numparas - i - 1)\epsilon_{\textnormal{gen}} = (2\numparas - 2i - 1)\epsilon_{\textnormal{gen}} \leq 2(\numparas - i)\epsilon_{\textnormal{gen}}
    \end{align}
\end{proof}

\subsection{Proof of Coherence-Surprise Theorem and Corollary}

\begin{theorem*}[Coherence-surprise tradeoffs] \quad
\begin{enumerate}
    \item \textbf{(Weak tradeoff)} An intelligent reader $\reader{}$ satisfies a tradeoff between clue effectiveness $\internalcoherence{i}$ and weak surprise $-\epsilon_{\text{surprise}}$: larger clue effectiveness requires weaker (larger $\epsilon$) surprise.
    \item \textbf{(Strong tradeoff)} For any reader $\reader{}$, strong surprise $\delta_{\text{surprise}}$ and clue effectiveness $\internalcoherence{i}$ trade off: their sum satisfies $\delta_{\text{surprise}} + \internalcoherence{i} \leq \intelligencegap{i}{\reader{}}$. In particular, assuming clue sufficiency, an intelligent $\reader{}$ cannot be strongly surprised.
\end{enumerate}
\end{theorem*}

\begin{proof}
Both claims build on a common bound. By the definition of clue effectiveness, for every
$1\leq i \leq \numparas$, we have:
\begin{align}\label{eq:uninformedness_brilliant}
    \E[\centropy(\knowitallreader(i);\, \brilliantreader(i))] = \E[\centropy(\knowitallreader(0);\, \brilliantreader(0))] - \internalcoherence{i} = \log|\culprits| - \internalcoherence{i}
\end{align}
where we use the assumption that for an empty story, the reader outputs a uniform
probability. Combined with the intelligence gap, we get a bound for any reader $\reader{}$:
\begin{align}\label{eq:uninformedness_reader}
     \E[\centropy(\knowitallreader(i);\, \reader{}(i))] \leq \log|\culprits| - \internalcoherence{i} + \intelligencegap{i}{\reader{}}
\end{align}

\paragraph*{Proof of (1) --- Weak tradeoff}
This part does not require clue sufficiency. If $\reader{}$ is
intelligent and weakly surprised, combining Eq.~\ref{eq:uninformedness_reader} with the
definition of weak surprise gives:
\begin{align}
    \log|\culprits| - \epsilon_{\text{surprise}} \leq \E[\centropy(\knowitallreader(i);\, \reader{}(i))] \leq \log|\culprits| - \internalcoherence{i} + \epsilon_{\text{intel}}
\end{align}
which simplifies to:
\begin{align}\label{eq:tradeoff2}
       \internalcoherence{i} - \epsilon_{\text{surprise}} \leq \epsilon_{\text{intel}}
\end{align}
Assuming small $\epsilon_{\text{intel}}$, the stronger the clue effectiveness (larger
$\internalcoherence{i}$), the weaker the surprise must be (larger $\epsilon_{\text{surprise}}$).

\paragraph*{Proof of (2) --- Strong tradeoff.}
If $\reader{}$ is strongly surprised, combining the definition of strong surprise with
Eq.~\ref{eq:uninformedness_reader}, we get:
\begin{align}\label{eq:tradeoff_base}
      \delta_{\text{surprise}} + \internalcoherence{i} \leq \intelligencegap{i}{\reader{}}
\end{align}
That is, strong surprise and clue effectiveness are bounded by the intelligence gap.
In particular, if $\reader{}$ is intelligent ($\intelligencegap{i}{\reader{}} \leq
\epsilon_{\text{intel}}$), and assuming clue sufficiency, the step effectiveness of the
know-it-all reader is non-negative (it forms a super-martingale), so clue sufficiency
gives $\internalcoherence{i} \geq - i \cdot \epsilon_{\text{ex}}$, and therefore:
\begin{align}\label{eq:tradeoff1}
      \delta_{\text{surprise}} \leq \epsilon_{\text{intel}} + i \cdot \epsilon_{\text{ex}}
\end{align}
For small $\epsilon_{\text{intel}}$ and $\epsilon_{\text{ex}}$ (such that
$\numparas\cdot\epsilon_{\text{ex}}$ is also small), $\delta_{\text{surprise}}$ must be
small, meaning an intelligent reader cannot be strongly surprised.
\end{proof}

\begin{corollary*}
If step effectiveness is non-negative for all steps ($\clueeffectiveness_{\brilliantreader}(j) \geq 0$ for all $j$), then $\internalcoherence{i}$ is non-decreasing in $i$.
Consequently, by Eq.~\ref{eq:tradeoff_base}, the maximum achievable strong surprise
$\delta_{\text{surprise}} \leq \intelligencegap{i}{\reader{}} - \internalcoherence{i}$
decreases as $i$ increases, unless the intelligence gap grows correspondingly.
By Eq.~\ref{eq:tradeoff2}, for an intelligent reader (small $\epsilon_{\text{intel}}$),
the condition $\internalcoherence{i} \leq \epsilon_{\text{intel}} + \epsilon_{\text{surprise}}$
requires $\epsilon_{\text{surprise}}$ to grow with $\internalcoherence{i}$, so even weak
surprise weakens as more clues are revealed.
\end{corollary*}

\subsection{Formal Definitions: Hindsight Gap and Hindsight Coherence}
\label{app:hindsight_formal}

A \defn{hindsight reader model} for reader $\reader{}$, denoted $\hreader{}$, re-evaluates the clues at step $i$ having already read the full story, including the revelation. That is, $\hreader{}(\story, i)$ may depend on the whole story $\story$, not just the prefix $\storyi{1}{i}$; it still outputs a probability distribution indexed by position $i$.

The \defn{hindsight gap} at step $i$ for reader $\reader{}$ is:
\begin{align}
    \delta_{\text{h}}(i, \reader{}) \coloneqq \internalcoherence{i}_{\hreader{}} - \internalcoherence{i}_{\reader{}}
\end{align}
where $\internalcoherence{i}_{\hreader{}}$ is the cumulative clue effectiveness of $\hreader{}$ up to step $i$.

The \defn{hindsight coherence property} of a story generation model $\storymodel$ requires that the step effectiveness of any hindsight reader $\hreader{}$ is close to that of the brilliant reader $\brilliantreader$:
\begin{align}\label{eq:hindsight_coherence}
\big|\clueeffectiveness_{\hreader{}}(i) - \clueeffectiveness_{\brilliantreader}(i)\big| \leq \epsilon_{\text{h}}, \quad \forall i \in 1 \dots \numparas
\end{align}
for some $\epsilon_{\text{h}} > 0$. This property ensures that the revelation provides a complete explanation of the clues---which leads a reader who has seen the resolution to re-evaluate them as well as the brilliant reader would.

\subsection{Proof of Hindsight Tradeoff}

\begin{corollary*}[Hindsight tradeoff]
    Assuming hindsight coherence, if the reader $\reader{}$ is strongly surprised then
    clue effectiveness and the hindsight gap trade off.
\end{corollary*}

\begin{proof}
Starting from Eq.~\ref{eq:tradeoff_base}, we apply the hindsight coherence property,
which gives:
\begin{align}
    \intelligencegap{i}{\reader{}} - \delta_{\text{h}}(i, \reader{}) \leq i \cdot \epsilon_{\text{h}} \Rightarrow \intelligencegap{i}{\reader{}} \leq i \cdot \epsilon_{\text{h}} + \delta_{\text{h}}(i, \reader{})
\end{align}
Substituting into Eq.~\ref{eq:tradeoff_base}:
\begin{align}\label{eq:tradeoff11}
    &\delta_{\text{surprise}} + \internalcoherence{i}  \leq \intelligencegap{i}{\reader{}}  \leq i \cdot \epsilon_{\text{h}} + \delta_{\text{h}}(i, \reader{}) \nonumber \\
    \Rightarrow \quad & \delta_{\text{surprise}} + \internalcoherence{i} - i \cdot \epsilon_{\text{h}} \leq \delta_{\text{h}}(i, \reader{})
\end{align}
This shows that to maintain a large surprise, the hindsight gap must be large, and it
must grow as clue effectiveness increases.
\end{proof}

%%----------------------------------------------------------------------
\section{The Real Story Dataset}
\label{app:stories}

\subsection{Sherlock Holmes stories}
Stories were collected from Project Gutenberg.\footnote{\url{https://www.gutenberg.org/}}
Stories 3--7, 9 include human annotation data (6 stories total with human data).

\subsection{Hercule Poirot stories}
Stories 5--9 include human annotation data.

Each story was segmented into approximately 25 paragraphs, corresponding to the
paragraph-level structure used for generated stories. The segmentation preserves natural
discourse boundaries (scene changes, speaker turns). For consistency with the generated
story setup, we define the \emph{revelation point} as the paragraph in which the
detective explicitly names the true culprit.

Tables~\ref{tab:stories-sherlock} and~\ref{tab:stories-poirot} list the real stories used in the experiments, together with the true culprit and the other suspects, which were presented to the model or human.
The story index corresponds to the identifier used throughout the paper (e.g., \textit{Sherlock} 0, \textit{Poirot} 3).
All Sherlock Holmes stories are drawn from \textit{The Adventures of Sherlock Holmes}; all Hercule Poirot stories are drawn from \textit{Poirot Investigates}.

\begin{table}[p]
\centering
\small
\caption{Sherlock Holmes stories from \textit{The Adventures of Sherlock Holmes}. Stories marked with $\dagger$ were not used in the human experiments.}
\label{tab:stories-sherlock}
\begin{tabular}{cllp{4cm}}
\toprule
\# & Title & Answer & Distractors \\
\midrule
0$^\dagger$ & A Scandal in Bohemia
  & Irene Adler
  & Sherlock Holmes, Dr.\ Watson, Godfrey Norton \\[2pt]
1$^\dagger$ & The Red-Headed League
  & Vincent Spaulding
  & Mr.\ Jabez Wilson, Mr.\ Duncan Ross, Mr.\ Merryweather \\[2pt]
2$^\dagger$ & A Case of Identity
  & Mr.\ Windibank
  & Mr.\ Hosmer Angel, Mrs.\ Sutherland, Mr.\ Hardy \\[2pt]
3 & The Boscombe Valley Mystery
  & John Turner
  & Charles McCarthy, James McCarthy, William Crowder \\[2pt]
4 & The Five Orange Pips
  & James Calhoun
  & John Openshaw, Joseph Openshaw, Elias Openshaw \\[2pt]
5 & The Man with the Twisted Lip
  & Hugh Boone
  & Isa Whitney, Kate Whitney, Neville St.\ Clair \\[2pt]
6 & The Adventure of the Blue Carbuncle
  & James Ryder
  & Henry Baker, John Horner, Catherine Cusack \\[2pt]
7 & The Adventure of the Speckled Band
  & Dr.\ Grimesby Roylott
  & Helen Stoner, Percy Armitage, Miss Honoria Westphail \\[2pt]
8$^\dagger$ & The Adventure of the Beryl Coronet
  & Sir George Burnwell
  & Alexander Holder, Arthur Holder, Lucy Parr \\[2pt]
9 & The Adventure of the Copper Beeches
  & Jephro Rucastle
  & Mrs.\ Rucastle, Toller, Mrs.\ Toller \\
\bottomrule
\end{tabular}
\end{table}

\begin{table}[p]
\centering
\small
\caption{Hercule Poirot stories from \textit{Poirot Investigates}. Stories marked with $\dagger$ were not used in the human experiments.}
\label{tab:stories-poirot}
\begin{tabular}{cllp{4cm}}
\toprule
\# & Title & Answer & Distractors \\
\midrule
0$^\dagger$ & The Adventure of ``The Western Star''
  & Gregory Rolf
  & Mary Marvell, Lady Yardly, Lord Yardly \\[2pt]
1$^\dagger$ & The Tragedy at Marsdon Manor
  & Mrs.\ Maltravers
  & Dr.\ Ralph Bernard, Captain Black, Alfred Wright \\[2pt]
2$^\dagger$ & The Adventure of the Cheap Flat
  & Elsa Hardt
  & Gerald Parker, Mrs.\ Robinson, Elsie Ferguson \\[2pt]
3$^\dagger$ & The Mystery of Hunter's Lodge
  & Zoe Havering
  & Roger Havering, Mrs.\ Middleton, Harrington Pace \\[2pt]
4$^\dagger$ & The Million Dollar Bond Robbery
  & Mr.\ Shaw
  & Philip Ridgeway, Mr.\ Vavasour, Miss Esm\'{e}e Farquhar \\[2pt]
5 & The Adventure of the Egyptian Tomb
  & Dr.\ Ames
  & Dr.\ Tosswill, Mr.\ Harper, Hassan \\[2pt]
6 & Jewel Robbery at the \textit{Grand Metropolitan}
  & The chambermaid
  & C\'{e}lestine, Mrs.\ Opalsen, Mr.\ Opalsen \\[2pt]
7 & The Kidnapped Prime Minister
  & Captain Daniels
  & O'Murphy, Mrs.\ Everard, Frau Bertha Ebenthal \\[2pt]
8 & The Disappearance of Mr.\ Davenheim
  & Mr.\ Davenheim
  & Lowen, Billy Kellett, Mrs.\ Davenheim \\[2pt]
9 & The Adventure of the Italian Nobleman
  & Mr.\ Graves
  & Dr.\ Hawker, Miss Rider, Signor Ascanio \\
\bottomrule
\end{tabular}
\end{table}

%%----------------------------------------------------------------------
\section{Generation Prompts}
\label{app:generation_prompt}

Stories are generated paragraph by paragraph. At each step $i$, the model receives the
story so far and is prompted to generate the next paragraph. The system prompt and initial
user prompt are as follows.

\subsection{System prompt}
\begin{mdframed}[backgroundcolor=gray!10,linewidth=0.5pt]
\small
You are a skilled author of detective fiction. Your task is to write a whodunit mystery
story, paragraph by paragraph. The story must have: (1) a clear culprit who committed the
crime; (2) a prominent distractor---a character who is made to look guilty through
misdirection but is ultimately innocent; (3) clues scattered throughout the story that, in
retrospect, point to the true culprit; (4) a revelation at the end that names the true
culprit. The story should be fair play: all the information needed to solve the mystery
should be present in the story, but arranged so that a careful reader can solve it while a
naive reader is misled. Make the story logical and coherent, so it makes full sense after
the revelation.
\end{mdframed}

\subsection{Per-step user prompt}
\begin{mdframed}[backgroundcolor=gray!10,linewidth=0.5pt]
\small
Continue the story. Write the next paragraph (paragraph \{i\} of \{L\}). The suspects are:
\{suspect list\}. The true culprit is: \{culprit\} (do not reveal this yet). The
distractor is: \{distractor\} (do not reveal their innocence yet). Keep the story
consistent with what has been written. This is paragraph \{i\} of a \{L\}-paragraph
story.

[Previous paragraphs: \{story so far\}]
\end{mdframed}

The final paragraph (paragraph $L$) uses a modified prompt instructing the model to
reveal the culprit and provide an explanation of the clues. The number of paragraphs is
fixed at $L = 25$ across all experiments.

\subsection{Gullible Reader Prompt}
\label{prompt:naive}

The gullible reader is estimated by prompting a model to predict the culprit at each step
of the story, acting as a naive investigator who takes the story at face value.

\begin{mdframed}[backgroundcolor=gray!10,linewidth=0.5pt]
\small
I want you to give me likelihood estimates for the true and distracting culprit identities in the following story.\\
I want lists of four prediction probabilities, a value for each of four suspects (representing the likelihood that the suspect is the true or distracting culprit). Even if a suspect is completely ruled out, he should still be included (and assigned a low probability).\\

Give me a JSON dictionary with the probabilities. Make sure to follow the exact format in the example. Give me only the JSON. Do not put comments in the JSON.\\
The suspects are: \{suspect list\}. Please do not change the identity and order of suspects.\\
        
For example, if it's clear that A is the distracting culprit and B is the most likely true culprit, then return something like:\\
'''json\\
\{\\
"suspects": ["A", "B", "C", "D"],\\
"probabilities": [0.0, 0.9, 0.05, 0.05],\\
"distractor\_probabilities": [1.0, 0.0, 0.0, 0.0]\\
\}\\
'''

I want you to give probability estimations as if the events in the story are real, ignoring the interest of the writer. You should estimate the most likely truth, even if is boring.

The story is:\\
\#\# BEGINNING OF STORY \#\#\\
\\
\{story\}\\
\\
\#\# END OF STORY \#\#
\end{mdframed}

Generation failure rates (stories without a clear culprit and distractor) are reported in Table~\ref{tab:results}.

This prompt is designed to elicit na\"ive, face-value reasoning by explicitly instructing
the model to ignore misdirection. We use o3-mini as the primary gullible reader model,
with o1-mini used in a small number of cases where o3-mini was unavailable.

\subsection{Experienced Reader Prompt}
\label{prompt:experienced}

The experienced reader $\expreader$ is estimated by prompting Gemini-3-flash with the current story prefix and, optionally, a set of previous stories from the same generating model. The model is asked to predict the culprit, taking into account patterns it may have observed across stories.

\begin{mdframed}[backgroundcolor=gray!10,linewidth=0.5pt]
\small
I want you to give me likelihood estimates for the true and distracting culprit identities in the following story.\\
I want lists of four prediction probabilities, a value for each of four suspects (representing the likelihood that the suspect is the true or distracting culprit). Even if a suspect is completely ruled out, he should still be included (and assigned a low probability).\\

Give me a JSON dictionary with the probabilities. Make sure to follow the exact format in the example. Give me only the JSON. Do not put comments in the JSON.\\
The suspects are: \{suspect list\}. Please do not change the identity and order of suspects.\\
        
For example, if it's clear that A is the distracting culprit and B is the most likely true culprit, then return something like:\\
'''json\\
\{\\
"suspects": ["A", "B", "C", "D"],\\
"probabilities": [0.0, 0.9, 0.05, 0.05],\\
"distractor\_probabilities": [1.0, 0.0, 0.0, 0.0]\\
\}\\
'''

I want you to give probability estimations for what will be revealed (or was already revealed) at the end of story. That is, the probability for each culprit is the probability that the story will be continued in way that ends up with this character being the culprit.\\
I Note that the story is generated paragraph by paragraph by a language model, for a total of {num\_paragraphs} paragraphs.\\
The generating model is instructed to introduce at least four characters, one of which is the true culprit and one is a distractor.\\
The clues are instructed to be misleading and point to the distracting character until the end of the story when the truth is discovered.

Try your best to deduce the culprit's identity. Use any knowledge at your disposal, including previous stories you've seen.

I'm also giving you some previous full stories written by this model. Use them to deduce the culprit's identity if relevant.\\
\\
\#\#\#\#\#\#\#\#\#\#\#\#\#\#\#\#\#\#\#\#\#\#\#\\
\#\# PREVIOUS STORIES \#\#\\
\\
\#\# BEGINNING OF PREVIOUS STORY 1\#\#\\
\\
\{previous story \#1\}\\
\\
\#\# END OF PREVIOUS STORY 1\#\#\\
\\
...\\
\\
\#\# END OF PREVIOUS STORIES \#\#\\
\#\#\#\#\#\#\#\#\#\#\#\#\#\#\#\#\#\#\#\#\#\#\#\#\#\#\#\#\#\\

--------------------------------------------------------------------------------
    
Now I'll give you the (possibly partial) story for which you must predict the culprit.

The story is:\\
\#\# BEGINNING OF (PARTIAL) STORY FOR PREDICTION \#\#\\
\\
\{story\}
\\
\#\# END OF (PARTIAL) STORY FOR PREDICTION \#\#    
\end{mdframed}

\subsection{Paragraph Filling Prompt}
\label{app:paragraph}

For paragraph filling (used in the \hyperref[app:erc]{ERC Metric} estimation), the following prompt is used:

\begin{mdframed}[backgroundcolor=gray!10,linewidth=0.5pt]
\small
I will give you a story with a missing paragraph (marked by ``[MISSING]'') and six options for filling it.

Give me a list of probabilities for the paragraph that can fill in the missing spot.
I stress that the goal is not to predict what was in the spot but rather to answer about what makes sense given the actual ending.

Notice that some paragraphs might be hidden (marked by [HIDDEN]). I will NOT ask you about filling those paragraphs, only the [MISSING] one.

For example, if the second option is the best, the first is also possible and the others make very little sense, then your answer should look like:
\begin{verbatim}
{
"options": ["a", "b", "c", "d", "e", "f"],
"probabilities": [0.25, 0.55, 0.05, 0.05, 0.05, 0.05]
}
\end{verbatim}

Your response should be the JSON dictionary only, with no additional text.

The story is:

\#\# BEGINNING OF STORY \#\#

\{story text\}

\#\# END OF STORY \#\#

The optional paragraphs are:

\{list of the optional paragraphs, in the form a. first paragraph, b. second paragraph\}
\end{mdframed}

%%----------------------------------------------------------------------
\section{Alternative Gullible Reader Estimation Methods}
\label{app:gullible_reader_estimation}

We evaluated several alternative prompting strategies for the gullible reader and
verified that they yield broadly similar trends to the main results.

\subsection{The Super-Naive Reader}

One alternative is the \emph{super-naive} reader, which additionally receives an
LLM-generated report of where the revelation occurs in the story. This reader is then modeled by placing zero mass on the
correct prediction throughout the suspicion phase (only predicting correctly after the
revelation). As a result, the super-na\"ive reader yields a larger upper bound on fair play
estimates and a surprise score that essentially reflects \emph{when} the story reveals
its culprit, rather than how well the story misleads the reader.

We evaluated both modes and found an important difference in their relationship to
subjective surprise ratings. The default gullible reader's surprise score correlates
significantly with human-reported surprise ($r=0.152$, $p<0.01$; Spearman, $n=359$),
while the super-naive surprise score does not ($r=-0.005$, $p=0.93$), indicating that revelations do not generally occur early in the story. This shows that
the default mode captures the narrative quality of misdirection---how effectively the
story fools the reader---whereas the super-naive mode captures a structural property
(timing of the revelation). We therefore use the default gullible reader for the main
results.

%%----------------------------------------------------------------------
\section{Sampling Procedure for the Know-It-All Reader}
\label{app:sampling_details}

The know-it-all reader $\knowitallreader(i)$ is defined by marginalization over the story model. Since this is generally intractable, we estimate it with Monte Carlo, sampling multiple story
continuations from the generating model starting from paragraph $i$, and using the
relative frequency of each culprit outcome as a probability estimate.

Concretely, at each checkpoint $i \in \{1, 5, 10, 15, 20, 25\}$, we generate $K = 15$
independent continuations (paragraphs $i+1$ through $L$) using temperature $= 1$, the
same setting used for the original generation. Each continuation is evaluated by the judge
model (o3-mini) to determine the culprit identity of the completed story. The know-it-all
probability for culprit $\culprit$ at step $i$ is then:

\begin{align}
    \hat{\reader{}}_{*}(i)_{\culprit} = \frac{1}{K} \sum_{k=1}^K \one[\text{culprit in continuation } k = \culprit]
\end{align}

For the real stories (Sherlock Holmes and Poirot), we cannot sample from the true
authorial distribution. We instead use Gemini-1.5-flash as a proxy story model,
conditioned on the story prefix, to sample plausible continuations. This provides an
approximation of the know-it-all reader, albeit without theoretical guarantees. The
resulting upper bounds for real stories should therefore be interpreted with caution.

The argmax predictor $\hat{y}_{\knowitallreader}(i) = \argmax_\culprit
\hat{\reader{}}_{\infty}(i)_\culprit$ is used for the coherence and fair play scores.

%%----------------------------------------------------------------------
\subsection{Computational Cost}

Each run of the pipeline for a single model (generating and evaluating 10 valid stories)
involves two language-model clients: a \emph{generator} (client 1, the model under
evaluation) and an \emph{evaluator} (client 2, used for culprit identification and
probability estimation).

Story generation is paragraph-by-paragraph with a growing context, giving quadratically
growing input tokens per story.  Assuming $\approx\!200$ output tokens per paragraph and
25 paragraphs per story, generation requires $\approx\!70\text{K}$ input tokens per story
attempt.  The number of attempts varies by model (fewer for higher-acceptance models);
the estimates below assume $\approx\!15$ attempts.  The generation phase totals roughly
$1.05\text{M}$ input + $75\text{K}$ output tokens for client 1, and $81\text{K}$ input
+ $2.3\text{K}$ output tokens for client 2.

The dominant cost is \emph{know-it-all reader estimation} (sampling): for each story,
$K = 20$ continuations are sampled at each of 19 checkpoints (due to budget and availability considerations, $K = 5$ continuations were sampled for GPT-4o, Gemini-1.5-flash, and Gemini-1.5-pro; see Table~\ref{tab:results}).
Because each sample regenerates up to 22 paragraphs with growing context, the
per-story sampling cost is $\approx\!13.3\text{M}$ input + $672\text{K}$ output tokens
for client 1 and $\approx\!2.3\text{M}$ input for client 2. See Table~\ref{tab:tokens} for a summary of token usage.

\begin{table}[p]
\centering
\caption{Estimated token usage per model run (10 generated + 10 sampled stories).}
\label{tab:tokens}
\small
\begin{tabular}{l|cc}
\hline
Phase & Client 1 (generator) & Client 2 (evaluator) \\
\hline\hline
Generation ($\times 10$ stories) & $\sim$1.05M in + $\sim$75K out & $\sim$81K in + $\sim$2.3K out \\
Sampling ($\times 10$ stories) & $\sim$133M in + $\sim$6.7M out & $\sim$23M in + $\sim$93K out \\
\hline
\textbf{Total} & $\sim$134M in + $\sim$6.8M out & $\sim$23M in + $\sim$95K out \\
\hline
\end{tabular}
\end{table}

The sampling cost scales quadratically with paragraph count and linearly with the number
of samples $K$ and checkpoints. With $K = 5$ instead of $K = 20$, sampling
cost drops to $\approx 25\%$.

%%----------------------------------------------------------------------
\section{ERC Metric}
\label{app:erc}

The \defn{Expected Retrospective Coherence (ERC)} metric measures the degree to which the
explicit clues in the story are causally connected to the final revelation. Unlike the
CUB score (which measures prediction accuracy), ERC measures the informational content
of the clues as parsed propositions.

Formally, for a story with clue sequence $\clues = \clue_1, \ldots, \clue_\numparas$ and
revelation $\trueculprit$, we define:
\begin{align}
    \eec = \frac{1}{\numparas} \sum_{i=1}^{\numparas}
    \one[\clue_i \text{ is causally connected to } \trueculprit]
\end{align}
where causal connection is assessed by a judge LLM given the full story
(including the revelation). We also define the threshold variant
$\eecc{t} = \one[\eec \geq t]$ as a binary quality indicator.

The ERC metric requires explicit clue extraction and causal judgment, making it more
expensive to compute than the CUB score. Results with ERC show broadly consistent trends
with the main metrics: models with high $\fairplayubscore$ also tend to have high \eec,
and vice versa. However, ERC tends to be more noisy due to the difficulty of consistently
parsing clues from long texts.

\subsection{Estimation}

Directly estimating the ERC requires an expectation over all possible clues, which is difficult to measure due to the unbounded  space.
We therefore use multiple relaxations. First, since text segments include the clues, we perform prediction over paragraphs instead of clues.
Second, we estimate the probability of a paragraph as a multiple-choice question with other paragraphs sampled from the same model.

For each narrative position $i$, we use the generated alternative paragraphs (from the same prefix, across different story samples) as the multiple-choice candidates; one of these is the original paragraph $x_i$. Masking $x_i$, we ask a judge model (o3-mini; see \hyperref[app:paragraph]{Paragraph Filling Prompt}) to identify the original paragraph from among the candidates and measure accuracy.
To avoid including superfluous errors from paragraphs similar in content, we only count errors where the selected paragraph leads to a story with a different culprit than the true culprit.

We evaluate three conditions: (1)~\textbf{None}~-- only the prefix up to the masked paragraph, which serves as a baseline (the best prediction achievable from the prefix alone); (2)~\textbf{AR}~-- prefix plus paragraphs \emph{after} the revelation, testing whether the paragraph is retrospectively coherent given the resolution; and (3)~\textbf{BR}~-- prefix plus paragraphs \emph{before} the revelation (up to the revelation point), testing whether the preceding clues make the paragraph predictable. Comparing AR and BR against None isolates the contribution of post-revelation context and pre-revelation foreshadowing, respectively.
Unlike the reader-based metrics, ERC is not an upper bound on clue effectiveness---it directly measures paragraph-level predictability rather than bounding it. However, its estimation is more brittle, as it depends on a judge model consistently parsing and evaluating clues from long texts.

%OA
\subsection{Results}

\renewcommand{\arraystretch}{1.4}
\begin{table}[p]
    \caption{Paragraph-filling (ERC) results for a subset of models for which ERC was computed (excludes Gemini-2.5-pro TB$-$1, Gemini-3 models, Llama-3.3-70B, and Llama-3.2-1B). Three conditions: \textbf{None}~-- only the prefix up to the masked paragraph; \textbf{AR}~-- prefix plus paragraphs \emph{after} the revelation; \textbf{BR}~-- prefix plus paragraphs \emph{before} the revelation (up to the revelation point). We report the ERC value multiplied by~25 and the ratio of stories with $\eec \geq 0.04$ (corresponding to one paragraph in a 25-paragraph story). The last two columns show the gain of AR and BR over the None baseline. All Llama models use the Instruct version.} \label{tab:erc}
    \centering
    \begin{tabular}{l|ccc|ccc|ccc}
         \hline
        \multirow{2}{*}{Model} & \multicolumn{3}{c|}{\eec~(x25)} & \multicolumn{3}{c|}{$\eecc{0.04}$ (ratio)} & \multicolumn{3}{c}{\eec~gain over None (x25)} \\
         & None & AR & BR & None & AR & BR & AR & BR & $\Delta$(BR$-$AR) \\
         \hline \hline
         Llama-3.2-1B & 1.70 & 2.62 & 3.09 & 0.67 & 1.00 & 1.00 & +0.93 & +1.39 & +0.46 \\
         Llama-3.2-3B & 1.88 & 3.62 & 5.35 & 0.63 & 0.88 & 0.88 & +1.74 & +3.47 & +1.73 \\
         Llama-3.1-8B & 2.06 & 1.59 & 6.69 & 0.56 & 0.67 & 1.00 & $-$0.46 & +4.63 & +5.09 \\
         Llama-3.1-70B & 1.69 & 1.55 & 4.47 & 0.50 & 0.60 & 1.00 & $-$0.14 & +2.78 & +2.92 \\
         Llama-3.3-70B & 1.81 & 1.94 & 4.58 & 0.50 & 0.70 & 0.90 & +0.14 & +2.78 & +2.64 \\
         \hline
         Gemini-1.5-flash & $-$0.02 & 1.23 & 1.78 & 0.30 & 0.60 & 0.60 & +1.25 & +1.81 & +0.55 \\
         Gemini-1.5-pro & 0.88 & 1.16 & 3.10 & 0.40 & 0.50 & 0.90 & +0.28 & +2.22 & +1.94 \\
         Gemini-2.5-flash TB 0 & 0.93 & 3.29 & 3.56 & 0.60 & 0.80 & 0.90 & +2.36 & +2.64 & +0.27 \\
         Gemini-2.5-flash TB -1 & 0.51 & 1.76 & 2.45 & 0.50 & 0.60 & 0.70 & +1.25 & +1.94 & +0.69 \\
         Gemini-2.5-pro & 0.54 & 1.95 & 2.65 & 0.40 & 0.60 & 0.70 & +1.41 & +2.10 & +0.69 \\
         \hline
         GPT-4o-mini & 4.66 & 6.57 & 8.83 & 0.75 & 0.75 & 1.00 & +1.91 & +4.16 & +2.26 \\
         GPT-4o & 1.23 & 3.12 & 5.10 & 0.60 & 0.60 & 1.00 & +1.89 & +3.87 & +1.99 \\
         \hline
    \end{tabular}
\end{table}

We see a consistent pattern across models: providing paragraphs before the revelation (BR) yields the largest ERC, followed by paragraphs after the revelation (AR), with the no-context baseline (None) generally lowest. This shows that earlier narrative context is more predictive of a missing paragraph than the revelatory ending, suggesting that the clues form a chain linked more strongly to the investigation than to the final resolution. For Llama-3.1-8B, the AR condition actually slightly underperforms the None baseline, which may reflect noise given the small sample. In many cases, both AR and BR ERC values remain below $0.04$ (one paragraph worth), indicating that even when positive, the contextual gain is small. GPT-4o-mini is a notable outlier with high absolute ERC even in the None condition, likely reflecting low diversity in its generated paragraphs rather than genuine coherence.

%%----------------------------------------------------------------------
\section{Human Experiment Details}
\label{app:experiment}

\subsection{Participants}
Twenty participants were recruited for the study ($n=20$; mean age 28.3 years, SD 4.7;
12 female, 8 male). All were fluent English readers with at least some prior experience
with mystery fiction. Participants were compensated with a gift card.

\subsection{Procedure}
\begin{figure}[p]
\centering
\begin{subfigure}[t]{0.47\linewidth}
    \centering
    \includegraphics[width=\linewidth]{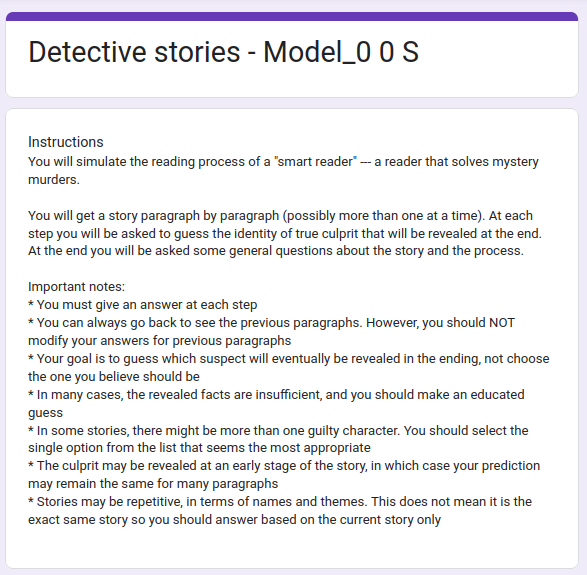}
    \caption{Instruction page for the human experiment.}
    \label{fig:instructions}
\end{subfigure}
\hfill
\begin{subfigure}[t]{0.47\linewidth}
    \centering
    \includegraphics[width=\linewidth]{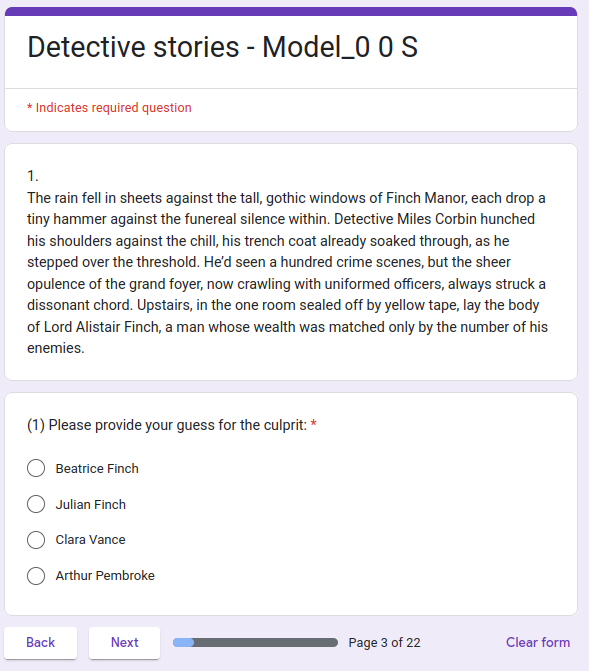}
    \caption{Example paragraph and suspects. The question repeats throughout the experiment as new paragraphs are shown.}
    \label{fig:example}
\end{subfigure}

\vspace{1em}
\begin{subfigure}[t]{\linewidth}
    \centering
    \includegraphics[width=\linewidth]{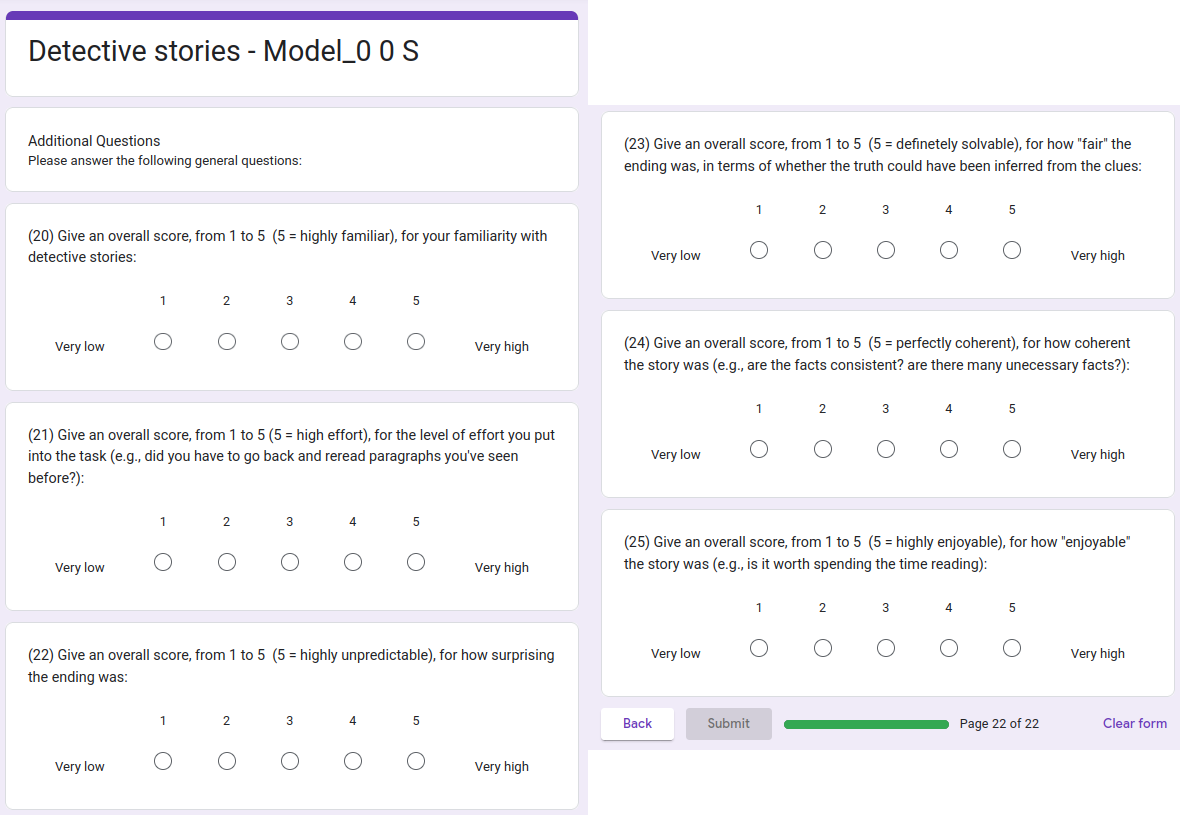}
    \caption{Questions for subjective rankings, given after the story.}
    \label{fig:additional}
\end{subfigure}
\caption{Human experiment interface screenshots.}
\label{fig:experiment-ui}
\end{figure}

The experiment was carried out using Google Forms.
The general instructions are shown in Figure~\ref{fig:instructions}, an example of a
repeating question is shown in Figure~\ref{fig:example}, and the additional questions
given after each story are shown in Figure~\ref{fig:additional}.

Participants read each story paragraph by paragraph on a web interface. After each
paragraph, they were prompted to select which suspect they believed was most likely the
culprit from the list of suspects presented at the beginning of the story. A ``not sure /
too early to tell'' option was provided but discouraged after the introduction phase.

After completing each story, participants rated the story on four dimensions (1--5 Likert
scale): fairness, coherence, surprise, and enjoyability. The exact questions are shown in Figure~\ref{fig:additional}.

Each participant read a subset of stories: all real stories assigned to them (either
Sherlock Holmes or Poirot, counterbalanced across participants) and a stratified random
sample of generated stories (3--5 per participant, sampled across models). Reading order
was randomized within each participant to control for learning and fatigue effects.

\subsection{Subjective ratings}
Table~\ref{tab:subjective} reports participants' subjective ratings of surprise, fairness, coherence, and enjoyability on a 1--5 scale, as mean $\pm$ standard deviation across per-story means.
Real stories (Poirot and Sherlock) are rated substantially higher than generated stories in fairness, coherence, and enjoyability: real stories score roughly $1$ point above the scale midpoint ($3$) on average, while most generated stories score at or below the midpoint on these dimensions.
Among generated stories, Gemini-2.5-pro and GPT-4o stand out as the most positively rated, with above-average coherence and enjoyability.
Llama-3.3-70B and Llama-3.2-3B receive the lowest ratings across most dimensions.
Surprise shows a different pattern: it does not clearly distinguish real from generated stories.

Among subjective measures, fairness, coherence, and enjoyability are strongly intercorrelated (Spearman $r = 0.62$--$0.71$, $p < 0.001$). Surprise is negatively correlated with subjective fairness ($r = -0.13$, $p < 0.05$), showing that readers tend to perceive unexpected outcomes as unfair. Surprise is not significantly correlated with subjective coherence ($r = -0.03$, $p = 0.54$) or enjoyability ($r = 0.06$, $p = 0.29$).
Subjective fairness and coherence are strongly correlated ($r = 0.71$), suggesting participants view these as related qualities.
Enjoyability does not significantly correlate with the measured surprise or coherence scores, though real stories were rated substantially more enjoyable than generated ones (relative scores of $+0.98$ to $+1.34$ vs.\ mostly negative for generated stories).

Participants were also asked to rank their familiarity with detective stories. Scores range from moderate to high (mean $3.71 \pm 1.01$, range $2$--$5$ on a 1--5 scale; $n = 22$).

\begin{table}[p]
    \caption{Human subjective evaluation (mean ratings on a 1--5 scale). Average rows show the mean $\pm$ between-story SD.} \label{tab:subjective}
    \centering
    \small
    \setlength{\tabcolsep}{4pt}
    \begin{tabular}{l|c|cccc}
         \hline
        Model & \# & Surprise & Fairness & Coherence & Enjoyability \\
         \hline \hline
         \makecell[l]{Llama 3.2-3B}  & 28 & 2.79 & 3.18 & 2.68 & 2.39 \\
         \makecell[l]{Llama 3.1-8B} & 24 & 3.96 & 2.75 & 2.67 & 2.50 \\
         \makecell[l]{Llama 3.1-70B} & 26 & 3.39 & 2.58 & 2.69 & 2.58 \\
         \makecell[l]{Llama 3.3-70B} & 55 & 3.38 & 2.40 & 2.47 & 2.29 \\
         \hline
         \makecell[l]{Gemini 1.5-pro} & 19 & 3.11 & 3.16 & 3.42 & 2.42 \\
         \makecell[l]{Gemini 2.5 flash TB 0} & 30 & 3.80 & 2.67 & 3.27 & 3.00 \\
         \makecell[l]{Gemini 2.5 flash TB -1} & 30 & 3.47 & 2.60 & 2.90 & 2.53 \\
         \makecell[l]{Gemini 2.5-pro} & 29 & 3.50 & 3.11 & 3.72 & 3.41 \\
         \hline
         \makecell[l]{GPT-4o mini} & 23 & 3.61 & 2.57 & 2.74 & 2.35 \\
         GPT-4o & 26 & 3.42 & 3.23 & 3.73 & 3.39 \\
         \hline
         \makecell[l]{Avg. generated} & 290 & \makecell{3.44 {\scriptsize$\pm$0.57}} & \makecell{2.77 {\scriptsize$\pm$0.56}} & \makecell{2.99 {\scriptsize$\pm$0.64}} & \makecell{2.65 {\scriptsize$\pm$0.54}} \\
         \hline
         \hline
         Sherlock & 33 & 3.09 & 3.85 & 4.21 & 4.30 \\
         Poirot & 37 & 3.70 & 4.03 & 4.51 & 4.11 \\
         \hline
         \makecell[l]{Avg.\\real} & 70 & \makecell{3.32 {\scriptsize$\pm$0.49}} & \makecell{3.91 {\scriptsize$\pm$0.45}} & \makecell{4.39 {\scriptsize$\pm$0.35}} & \makecell{4.21 {\scriptsize$\pm$0.38}} \\
         \hline
    \end{tabular}
\end{table}

\subsection{Correlations between subjective and objective measures}
\label{app:correlations}

Tables~\ref{tab:corr-subjobj} and~\ref{tab:corr-intersubj} report Spearman correlations
at the individual observation level ($n \approx 360$). The inter-subjective correlations
(Table~\ref{tab:corr-intersubj}) show that coherence and fairness are strongly
correlated, while surprise is negatively correlated with fairness, reflecting the
perceived tension between being surprised and viewing the story as fair. The
correlations with objective measures (Table~\ref{tab:corr-subjobj}) are modest but
significant for several pairs, validating that the computational metrics capture
relevant aspects of narrative quality.

\begin{table}[p]
\centering
\small
\caption{Spearman correlations between subjective ratings and objective measures ($n \approx 359$--$360$ individual observations). Surprise score uses the default gullible reader.}
\label{tab:corr-subjobj}
\begin{tabular}{llcc}
\toprule
Subjective & Objective & $r$ & $p$ \\
\midrule
Surprise & $\surprisalscore$ (default) & $0.152$ & $0.004$ \\
Surprise & Human-guessing surprise & $0.268$ & $<0.001$ \\
Coherence & $\arfpscore$ (human) & $0.155$ & $0.003$ \\
Coherence & $\coherencescore$ (LLM) & $0.106$ & $0.045$ \\
Fairness & $\arfpscore$ (human) & $0.250$ & $<0.001$ \\
Fairness & $\arfpscore$ (default) & $0.104$ & $0.048$ \\
Enjoyability & $\coherencescore$ (LLM) & $0.113$ & $0.032$ \\
\bottomrule
\end{tabular}
\end{table}

\begin{table}[p]
\centering
\small
\caption{Spearman correlations among subjective ratings ($n \approx 358$--$360$ individual observations).}
\label{tab:corr-intersubj}
\begin{tabular}{llcc}
\toprule
Dimension 1 & Dimension 2 & $r$ & $p$ \\
\midrule
Surprise & Fairness & $-0.131$ & $0.013$ \\
Surprise & Coherence & $-0.033$ & $0.537$ \\
Coherence & Fairness & $0.650$ & $<0.001$ \\
Coherence & Enjoyability & $0.713$ & $<0.001$ \\
Fairness & Enjoyability & $0.624$ & $<0.001$ \\
\bottomrule
\end{tabular}
\end{table}

%%----------------------------------------------------------------------
\section{Reading Curves}
\label{app:reading-curves}

\subsection{Definition}

The probability a reader model $\reader{}$ assigns to a given culprit $y$ at each point in the story defines a trajectory over paragraphs, which we call the \defn{reading curve} for $y$:
\[
    \readert{}{y} : \mathbb{N} \to [0,1], \qquad \readert{}{y}(i) \coloneqq \reader{}(\story, i)_y
\]
We are primarily interested in the reading curve for the true culprit, $\readert{}{t}$, and for the main distractor, $\readert{}{d}$.
These curves visualize how a reader's belief evolves across the three narrative phases (introduction, suspicion, revelation) and provide an intuitive complement to the scalar metrics defined in the main text.

Reading curves for different reader types exhibit characteristic shapes.
For the \textbf{gullible reader} $\gulliblereader$, the true-culprit curve starts near $\frac{1}{|\culprits|}$ in the introduction phase, decreases toward $0$ during suspicion as the distractor attracts probability mass, then jumps to $\approx 1$ at revelation. The distractor curve mirrors this, rising toward $1$ during suspicion and collapsing to $\approx 0$ at revelation. For the \textbf{know-it-all reader} $\knowitallreader$ (and, when clue sufficiency holds, the brilliant-detective reader $\brilliantreader$), the true-culprit curve rises monotonically from $\frac{1}{|\culprits|}$, reaching high confidence before or at the revelation, while the distractor curve declines.

\subsection{Experimental reading curves}

Experimental reading curves serve as interpretive case studies: they make it possible to understand \emph{why} a story receives a particular metric score by tracing the evolution of reader beliefs across the narrative. In particular, they reveal the structural difference between fair-play and \textit{Deus ex Machina} stories---in fair-play stories, the know-it-all and brilliant-detective readers gradually converge on the culprit well before the revelation, whereas in \textit{Deus ex Machina} stories these readers remain uncertain until the final reveal.

\begin{figure}[p]
\centering

\begin{subfigure}[t]{0.45\textwidth}
    \includegraphics[width=\linewidth]{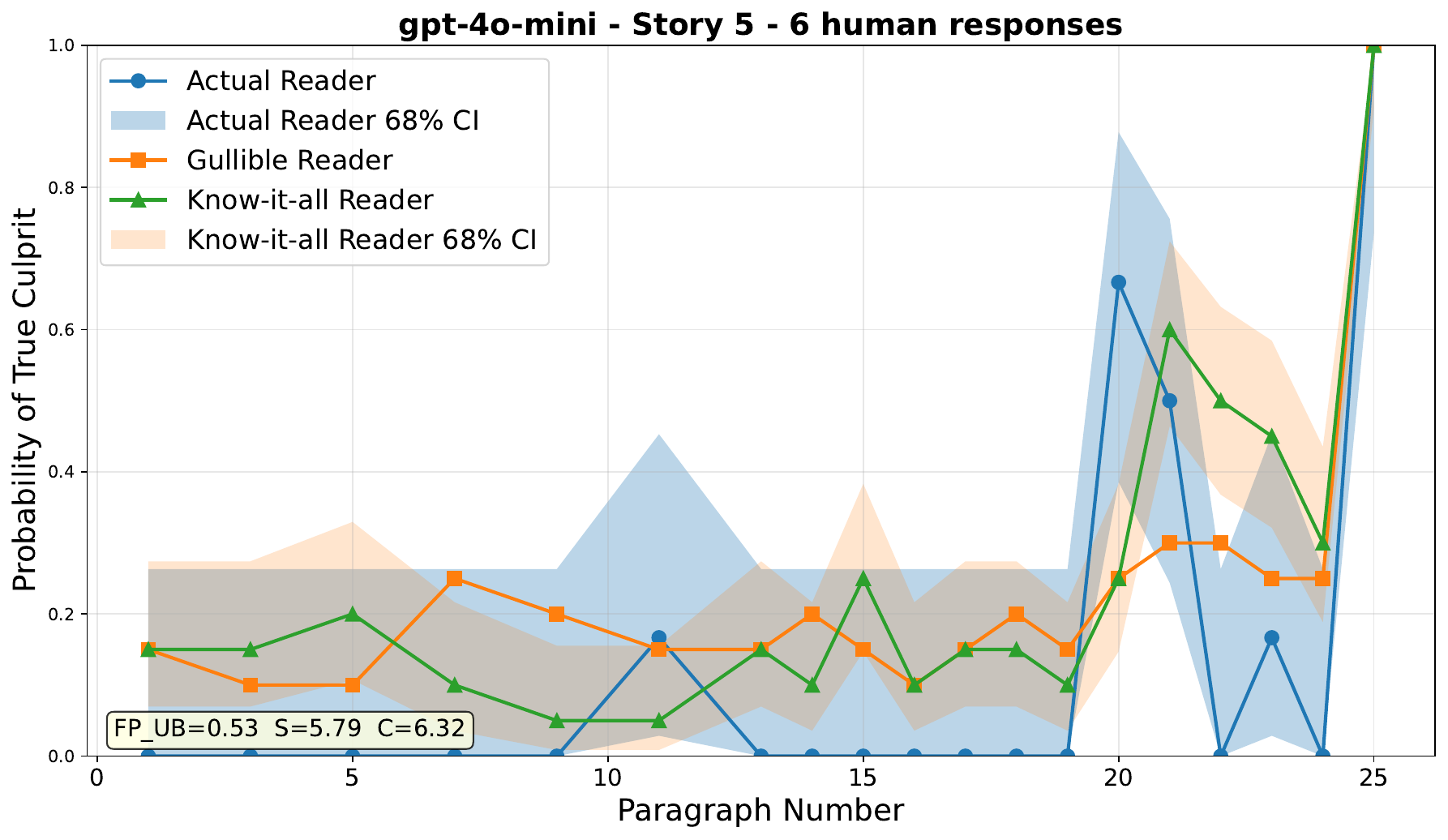}
    \caption*{\textit{Deus ex Machina} (1)}
\end{subfigure}
\hfill
\begin{subfigure}[t]{0.45\textwidth}
    \includegraphics[width=\linewidth]{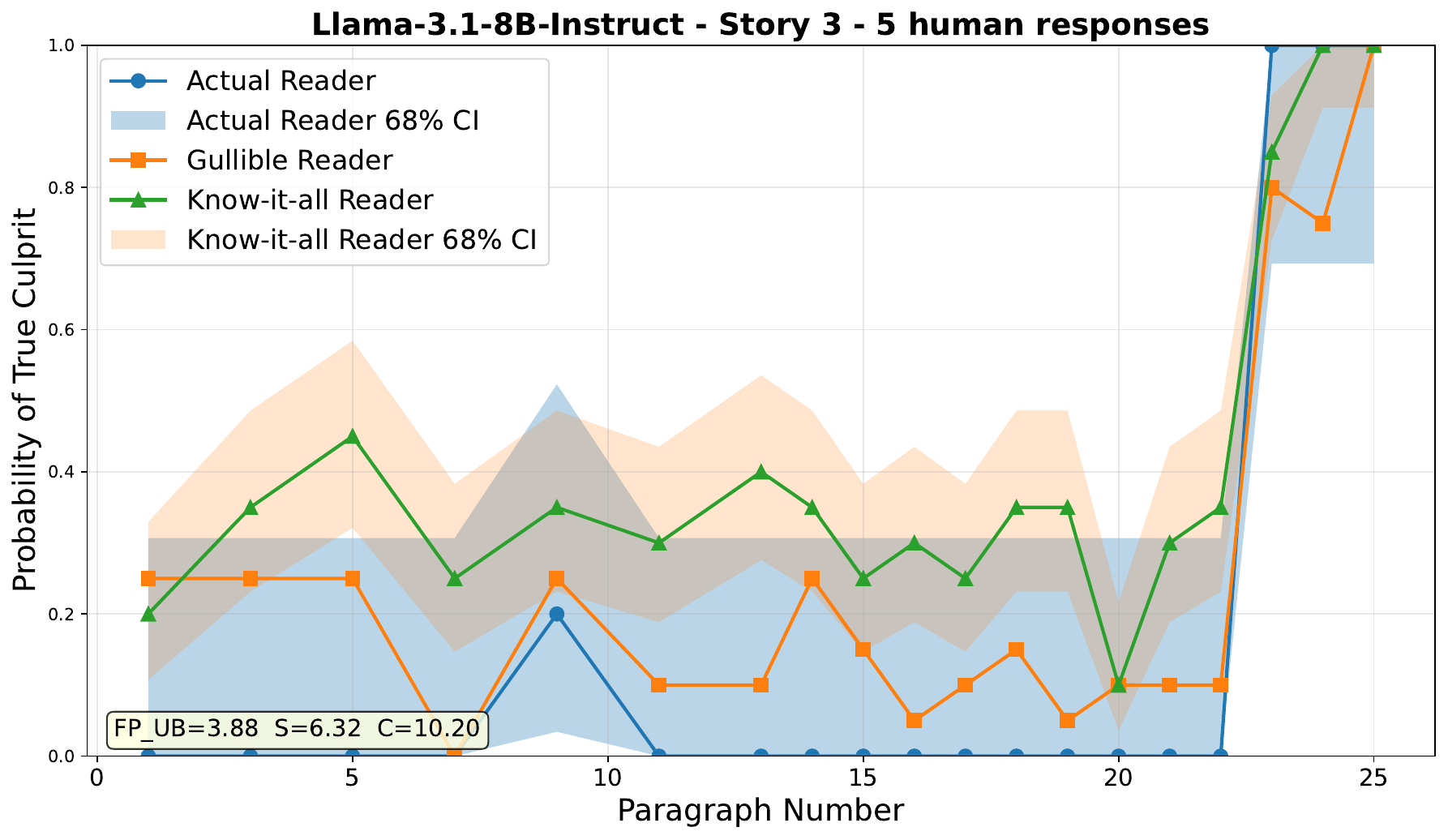}
    \caption*{\textit{Deus ex Machina} (2)}
\end{subfigure}

\vspace{0.4em}

\begin{subfigure}[t]{0.45\textwidth}
    \includegraphics[width=\linewidth]{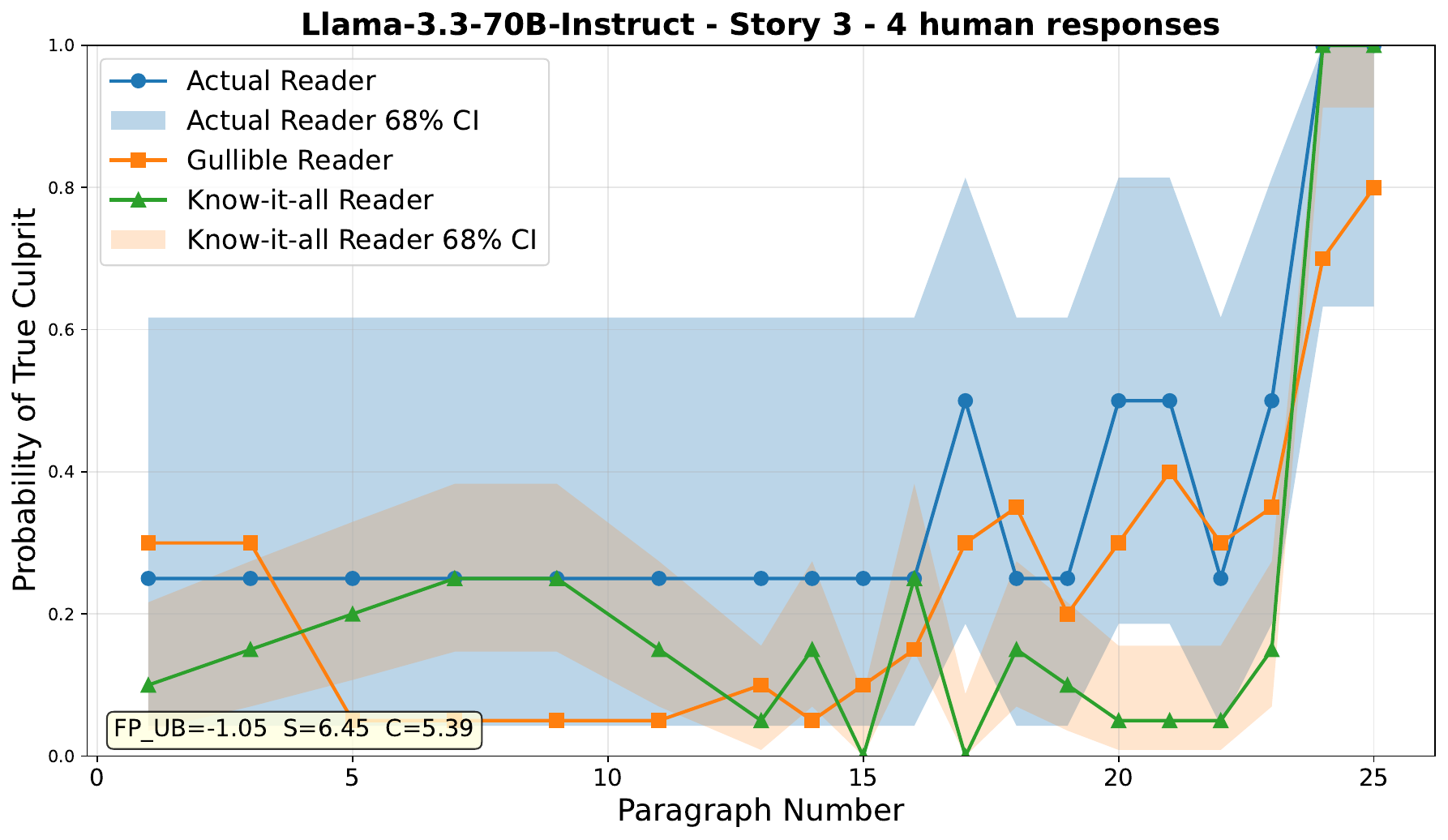}
    \caption*{\textit{Deus ex Machina} (3)}
\end{subfigure}
\hfill
\begin{subfigure}[t]{0.45\textwidth}
    \includegraphics[width=\linewidth]{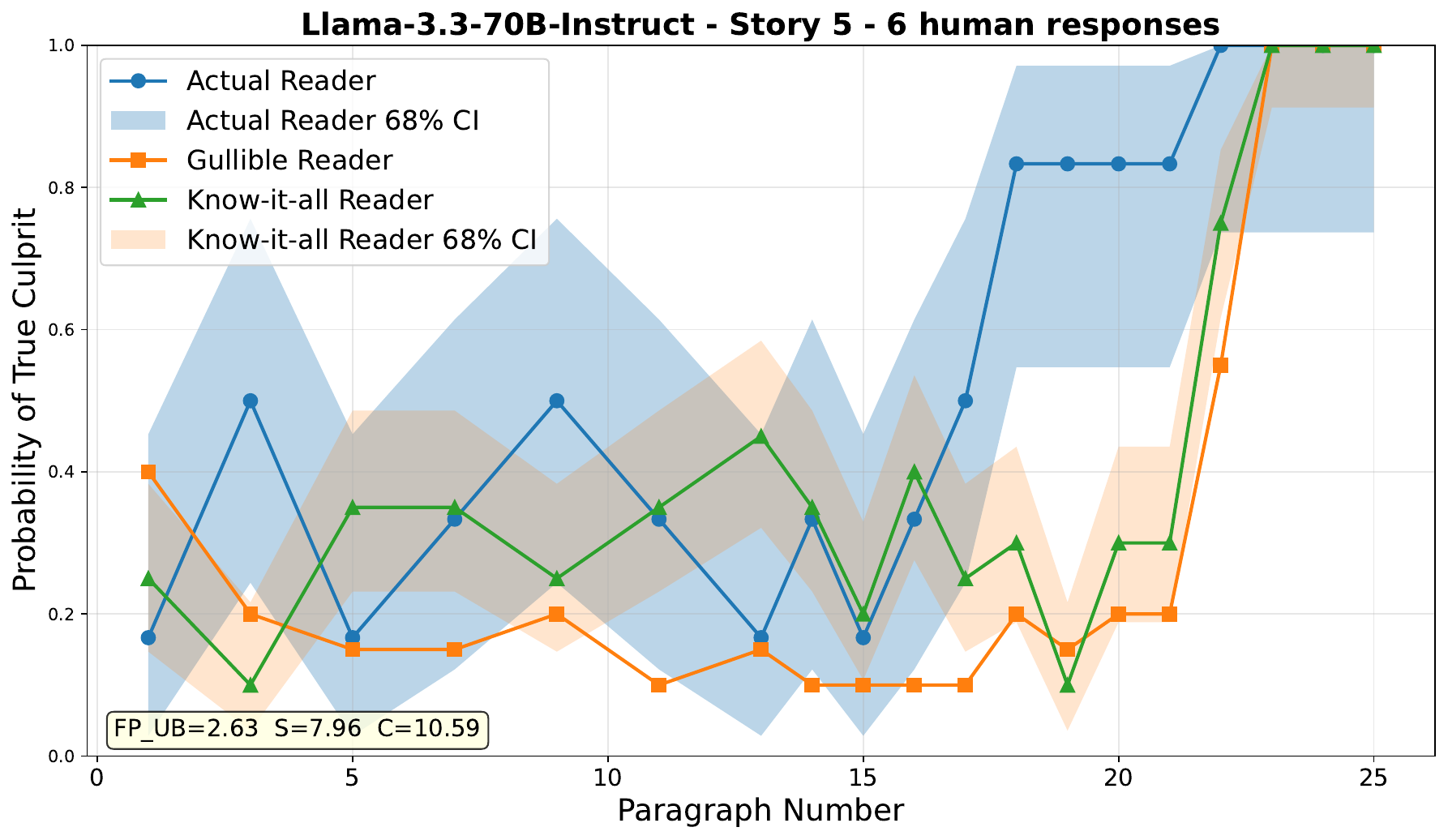}
    \caption*{\textit{Deus ex Machina} (4)}
\end{subfigure}

\vspace{0.4em}

\caption{Reading curves for representative generated stories with low fair play scores (\textit{Deus ex Machina}), shown as raw
probability assigned to the true culprit at each paragraph.
The know-it-all and gullible reader curves track each other closely throughout, indicating that the clues provide
little genuine misdirection.
Shaded bands are 68\% Clopper-Pearson confidence intervals \citep{clopper1934use} on the actual-reader curve.}
\label{fig:app-reading-curves-generated-dem}
\end{figure}

\begin{figure}[p]
\centering

\begin{subfigure}[t]{0.45\textwidth}
    \includegraphics[width=\linewidth]{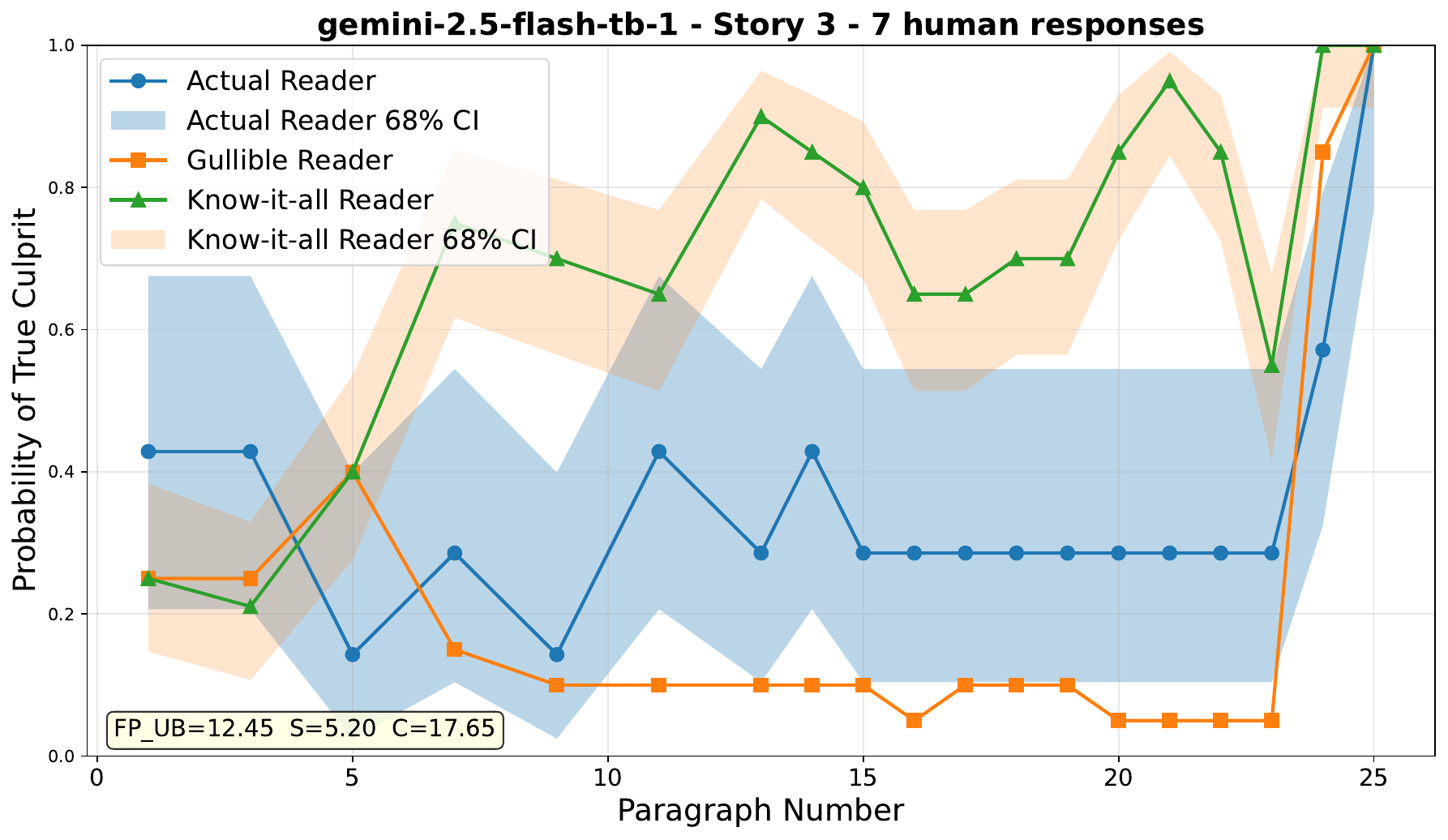}
    \caption*{\textit{Fair Play} (1)}
\end{subfigure}
\hfill
\begin{subfigure}[t]{0.45\textwidth}
    \includegraphics[width=\linewidth]{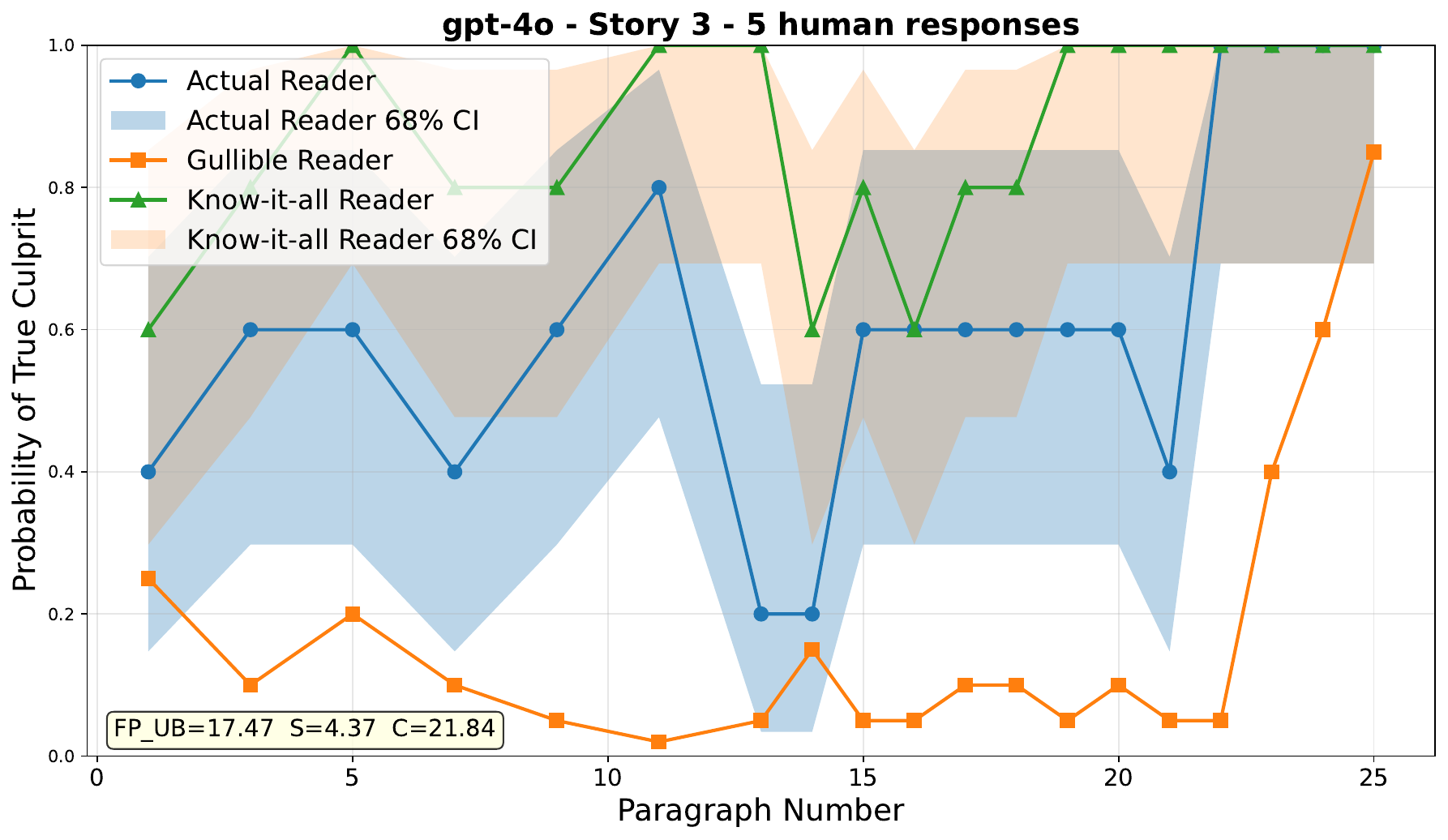}
    \caption*{\textit{Fair Play} (2)}
\end{subfigure}

\vspace{0.4em}

\begin{subfigure}[t]{0.45\textwidth}
    \includegraphics[width=\linewidth]{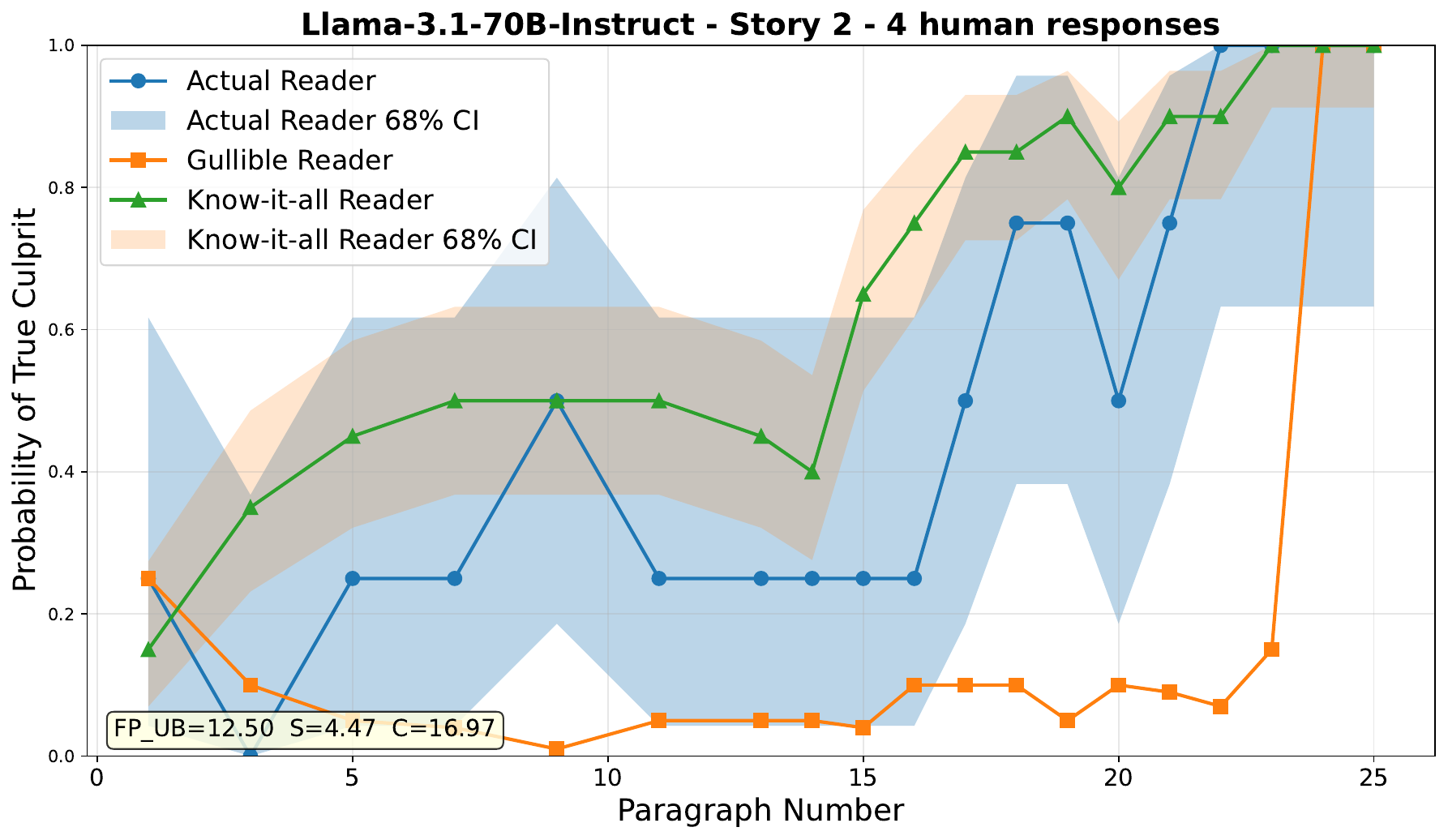}
    \caption*{\textit{Fair Play} (3)}
\end{subfigure}
\hfill
\begin{subfigure}[t]{0.45\textwidth}
    \includegraphics[width=\linewidth]{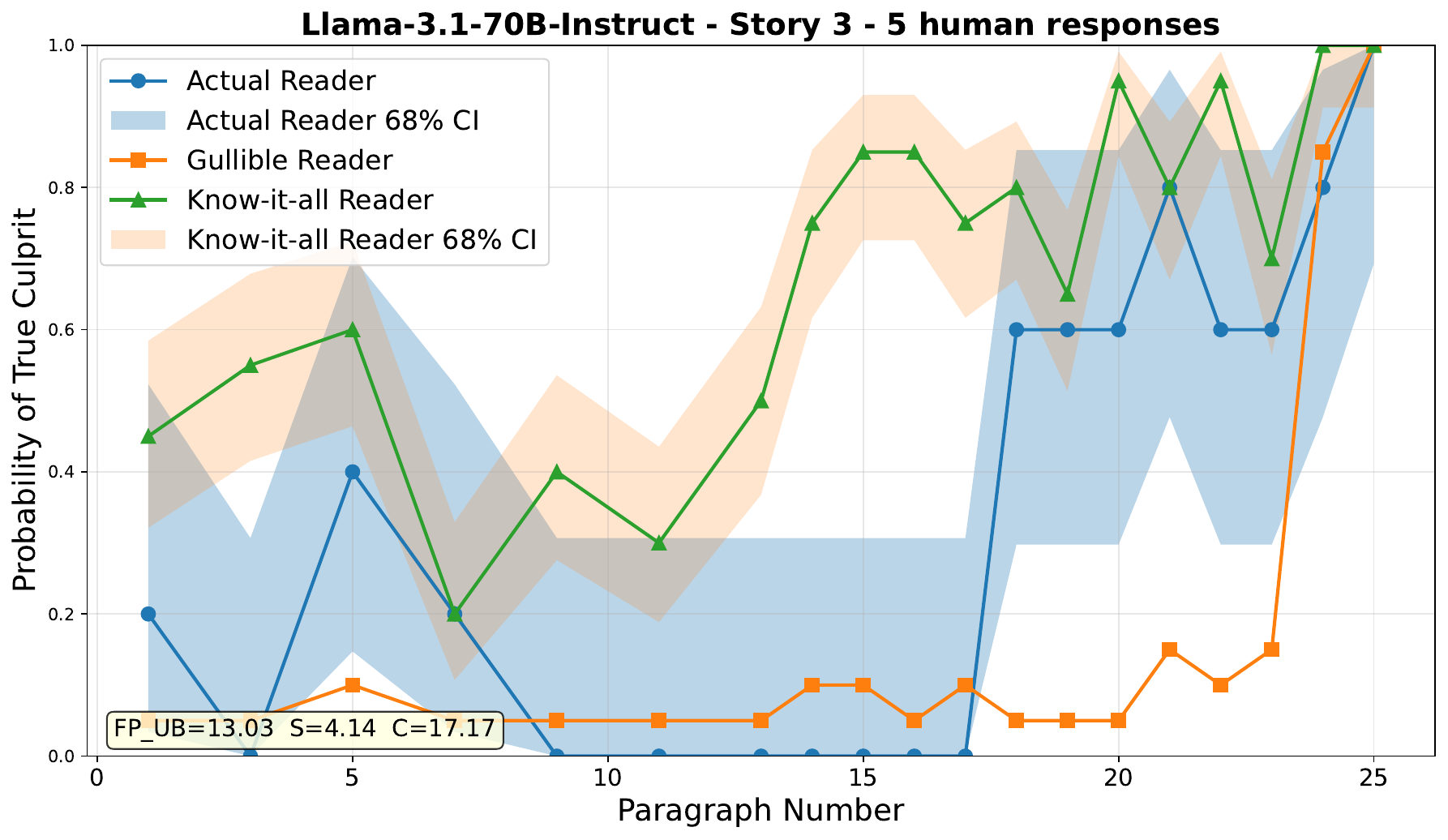}
    \caption*{\textit{Fair Play} (4)}
\end{subfigure}

\vspace{0.4em}

\caption{Reading curves for representative generated stories with high fair play scores (\textit{Fair Play}), shown as raw
probability assigned to the true culprit at each paragraph.
A sustained gap between the know-it-all and gullible curves reflects successful misdirection combined with
coherent clues.
Shaded bands are 68\% Clopper-Pearson confidence intervals \citep{clopper1934use} on the actual-reader curve.}
\label{fig:app-reading-curves-generated-fp}
\end{figure}

Representative examples of fair-play and \textit{Deus ex Machina} reading curves are shown in the main text (Figure~3C--D). Figures~\ref{fig:app-reading-curves-generated-dem} and~\ref{fig:app-reading-curves-generated-fp} show additional examples of generated stories with low and high fair play scores, respectively. Figure~\ref{fig:app-reading-curves-sherlock5} shows a notable real-story example.

\subsection{Reading curves for real stories}

\begin{figure}[p]
\centering
\includegraphics[width=0.8\textwidth]{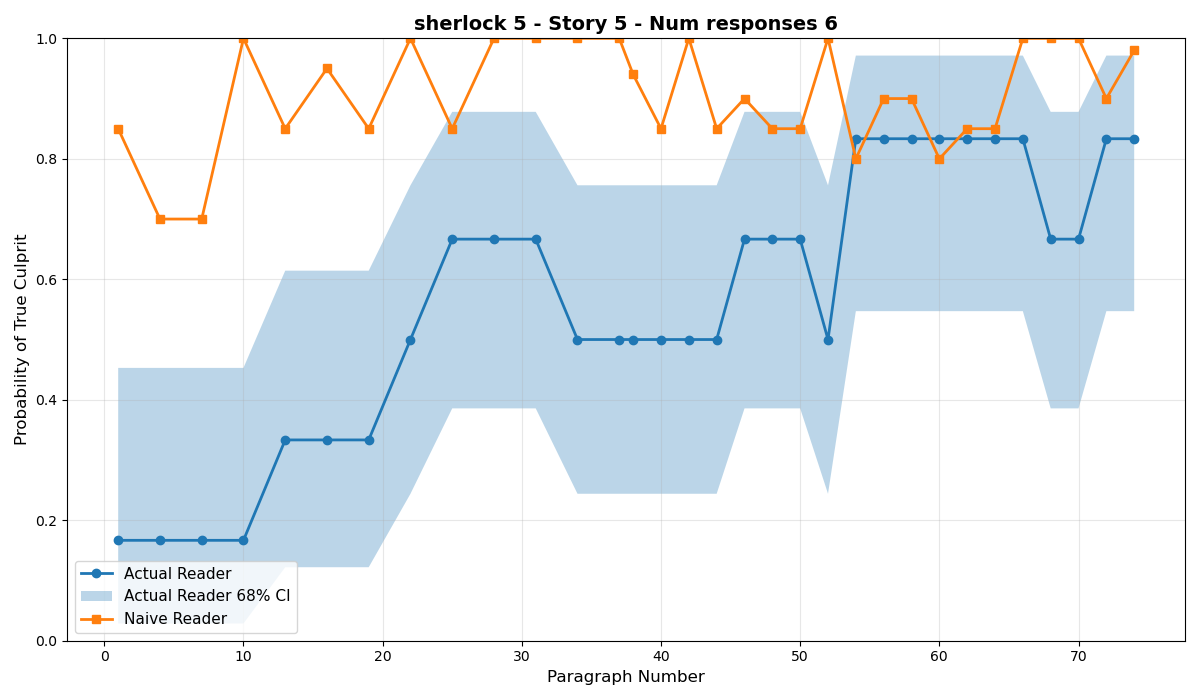}
\caption{Reading curves for Sherlock Holmes story 5 (``The Man with the Twisted Lip'').
The gullible reader (red) outperforms human participants (green) on this story---a notable
exception to the general trend. Human readers appear to discount the obvious suspect due
to genre expectations (``there must be a twist''), while the gullible reader takes the
clues at face value. This illustrates how prior genre experience can work against accuracy
when a story does not conform to conventional patterns.}
\label{fig:app-reading-curves-sherlock5}
\end{figure}

Figure~\ref{fig:app-reading-curves-sherlock5} shows the reading curve for Sherlock story 5 (``The Man with the Twisted Lip''), where human participants perform worse than the gullible reader. This suggests that human readers, familiar with detective fiction conventions, expect a twist and discount the obvious suspect, while the gullible reader takes the clues at face value.

%%----------------------------------------------------------------------
\section{Experienced Reader: Full Results}
\label{app:experienced-reader}

The experienced reader $\expreader$ is provided with a set of previous stories from the
same generating model before reading the current story. It was estimated by prompting Gemini-3-flash---chosen for its strong long-context understanding---with 0, 2, 3, 5, or 9 previous stories from the same generating model before reading the current story. A high experienced reader fair
play score $\expfpscore$ indicates that exposure to previous stories substantially
improves culprit prediction, suggesting the model relies on recurring patterns---a sign
that clue sufficiency may be violated: the story can be solved not from its clues alone
but from knowledge of the model's writing tendencies.

\paragraph*{Mode collapse}
An extreme case of pattern reliance is \emph{mode collapse}, where a model converges on
the same story structure repeatedly. Gemini-1.5-pro is a clear example: approximately
90\% of its stories use the same culprit name, and its human fair play score is close to
its know-it-all upper bound (see Table~\ref{tab:exp-reader-per-model}), indicating that
the stories are easily predicted once a reader is familiar with the model's output.
More generally, models where $\expfpscore$ is close to $\fairplayubscore$ (e.g.,
Llama-3.1-70B) show low diversity---a few previous stories almost suffice to predict
outcomes. Models with larger gaps between $\expfpscore$ and $\fairplayubscore$ (e.g.,
Gemini-2.5-flash TB 0) produce more varied stories where familiarity provides less
predictive advantage.

\paragraph*{Source of experienced reader advantage}
Comparing the 0-shot experienced reader (no previous stories) with the multi-story settings reveals that much of the predictive gain already appears with zero prior-story context: the 0-shot experienced reader achieves performance close to the know-it-all reader, outperforming the actual readers, even without exposure to previous stories.
A possible explanation is that the main power of the experienced reader comes from knowing the generation process itself---prompt, structure, and stylistic conventions---rather than from accumulating information across individual stories.
Incremental gains from additional context are real but smaller, reflecting that story-specific patterns beyond the generating model's general tendencies provide limited additional signal for most models.

Table~\ref{tab:exp-reader-per-model} reports the experienced reader fair play score $\expfpscore$ per model.
For each model, it shows results on the annotated subset (with the know-it-all upper bound $\fairplayubscore$ and human average $\arfpscore$) alongside results on all valid stories ($n_\text{all}$) at 0, 3, and 9 prior stories.
As the number of prior stories increases, $\expfpscore$ generally increases, reflecting that familiarity with a model's output patterns improves culprit prediction. The rate of increase varies substantially across models.

\begin{figure}[p]
\centering
\begin{subfigure}[t]{0.56\linewidth}
\centering
\includegraphics[width=\linewidth]{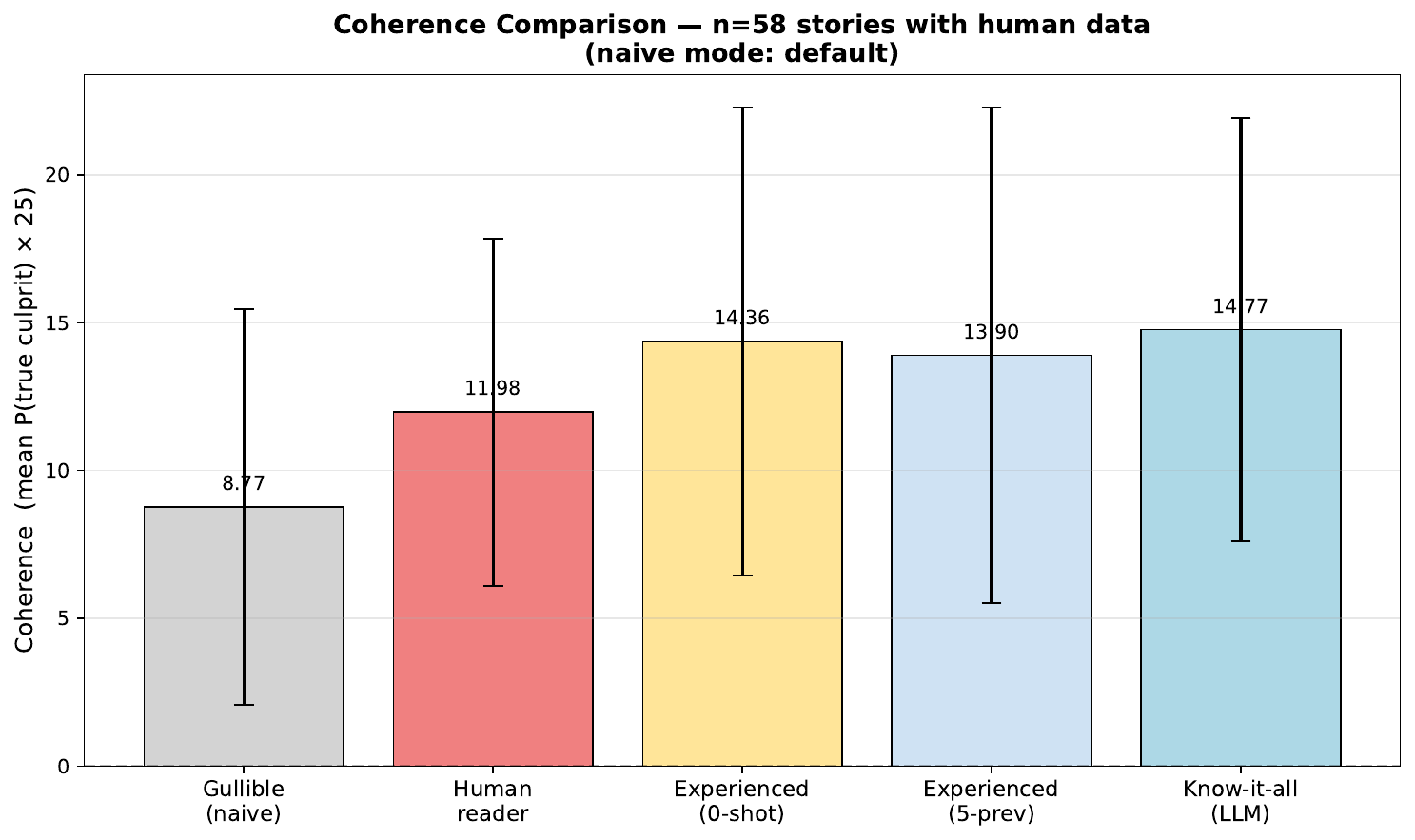}
\caption{Coherence scores per model (all stories), comparing know-it-all, human, and gullible readers. The expected hierarchy (know-it-all $\geq$ human $\geq$ gullible) holds for most models.}
\label{fig:barplot-coherence}
\end{subfigure}
\hfill
\begin{subfigure}[t]{0.39\linewidth}
\centering
\includegraphics[width=\linewidth]{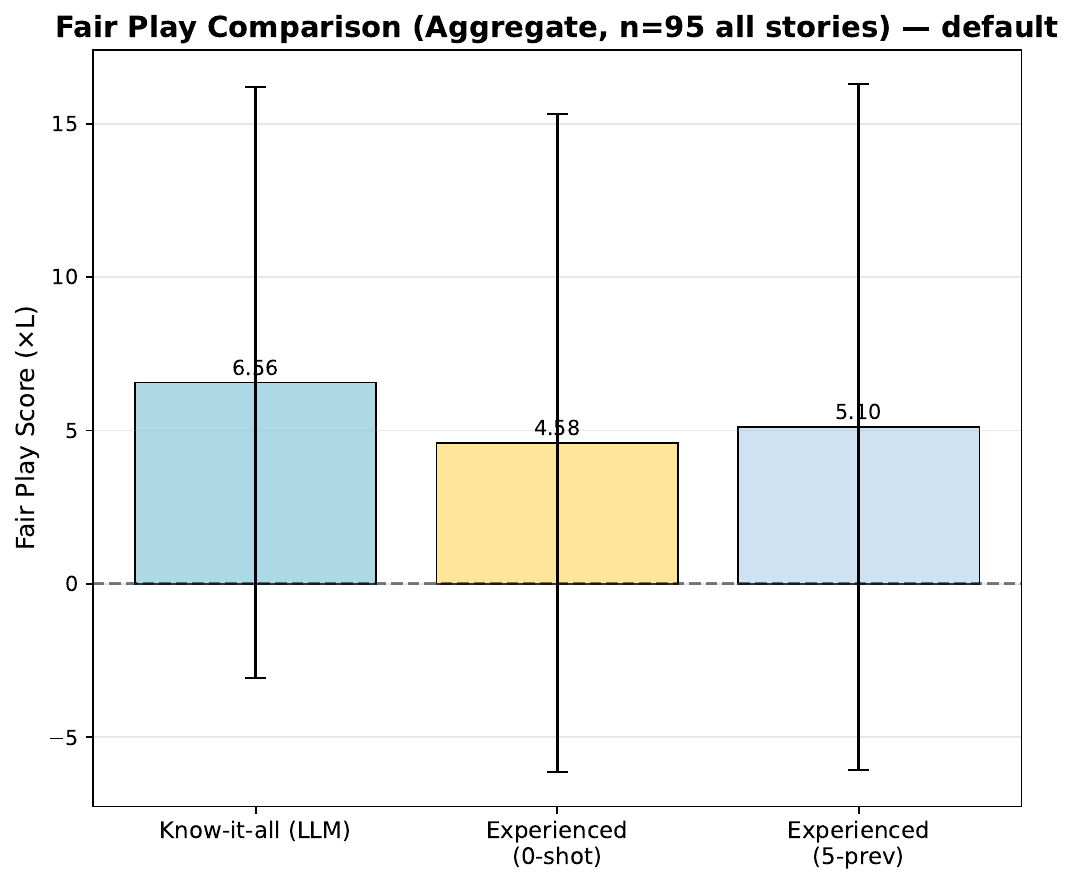}
\caption{Aggregate fair play scores per model for all stories, comparing the know-it-all upper bound ($\fairplayubscore$) and the experienced reader ($\expfpscore$) at 0, 2, 3, 5, and 9 previous stories. The gap between know-it-all and experienced reader is larger when all stories are included (vs.\ human-annotation subset only).}
\label{fig:barplot-fairplay}
\end{subfigure}
\caption{Coherence and fair play scores for all generated stories.}
\label{fig:barplots-combined}
\end{figure}

\paragraph*{Coherence and fair play scores.}
Figure~\ref{fig:barplot-coherence} shows coherence scores ($\coherencescore$) per model across all stories, comparing the know-it-all, human, and gullible reader perspectives. The expected hierarchy (know-it-all $\geq$ human $\geq$ gullible) is visible across most models, though Gemini-1.5-pro deviates due to mode collapse (see below).
Figure~\ref{fig:barplot-fairplay} shows aggregate fair play scores for all stories, including experienced reader fair play at varying numbers of previous stories.
With more prior stories, the experienced reader generally approaches the know-it-all upper bound, though the rate of increase varies across models; notably, using 5 previous stories performed slightly worse than no previous stories for some models.

\begin{table}[p]
    \caption{Per-model experienced reader fair play scores $\expfpscore$.
    Left block: annotated subset ($n_a$ stories with human annotations); $\fairplayubscore$ (know-it-all) and $\arfpscore$ (human average) are shown for comparison; $\expfpscore$ columns show 0-shot, 0-alt, and selected prior-story settings.
    Right block: all valid stories ($n$ stories); $\expfpscore$ at 0, 3, and 9 prior stories.
    ``--'' indicates insufficient data.}
    \label{tab:exp-reader-per-model}
    \centering
    \scriptsize
    \setlength{\tabcolsep}{2.5pt}
    \renewcommand{\arraystretch}{1.2}
    \begin{tabular}{l|c|cc|cccccc|c|ccc}
         \hline
         \multirow{2}{*}{Model}
           & \multirow{2}{*}{$n_a$}
           & \multirow{2}{*}{$\fairplayubscore$}
           & \multirow{2}{*}{$\arfpscore$}
           & \multicolumn{6}{c|}{$\expfpscore$ (annotated)}
           & \multirow{2}{*}{$n$}
           & \multicolumn{3}{c}{$\expfpscore$ (all)} \\
          & & & & 0-shot & 0-alt & 2-prev & 3-prev & 5-prev & 9-prev & & 0-prev & 3-prev & 9-prev \\
         \hline \hline
         Llama-3.2-3B & 4 & 0.118 & 0.169 & $-$0.151 & -- & -- & $-$0.086 & -- & 0.013 & 8 & 0.007 & 0.013 & 0.013 \\
         Llama-3.1-8B & 5 & 0.182 & 0.037 & 0.032 & 0.114 & $-$0.004 & 0.005 & $-$0.004 & $-$0.039 & 9 & $-$0.039 & $-$0.039 & $-$0.039 \\
         Llama-3.1-70B & 6 & 0.538 & 0.345 & 0.480 & 0.463 & 0.425 & 0.439 & 0.461 & 0.522 & 10 & 0.401 & 0.510 & 0.522 \\
         Llama-3.3-70B & 10 & 0.039 & 0.039 & $-$0.084 & $-$0.112 & $-$0.196 & $-$0.126 & $-$0.089 & 0.011 & 10 & 0.006 & 0.009 & 0.011 \\
         \hline
         Gemini-1.5-pro & 5 & 0.234 & 0.246 & 0.235 & 0.263 & 0.237 & 0.287 & 0.414 & 0.308 & 10 & 0.241 & 0.286 & 0.308 \\
         Gemini-2.5-flash TB 0 & 5 & 0.351 & 0.066 & 0.284 & 0.211 & 0.164 & 0.227 & 0.135 & 0.143 & 10 & 0.041 & 0.103 & 0.143 \\
         Gemini-2.5-flash TB -1 & 5 & 0.484 & 0.090 & 0.346 & 0.356 & 0.355 & 0.350 & 0.324 & 0.408 & 10 & 0.118 & 0.316 & 0.408 \\
         Gemini-2.5-pro & 5 & 0.455 & 0.213 & 0.401 & 0.469 & 0.668 & 0.539 & 0.539 & 0.426 & 10 & 0.134 & 0.346 & 0.426 \\
         \hline
         GPT-4o-mini & 4 & $-$0.020 & 0.096 & $-$0.016 & $-$0.197 & $-$0.096 & 0.118 & $-$0.329 & $-$0.076 & 8 & $-$0.086 & $-$0.077 & $-$0.076 \\
         GPT-4o & 5 & 0.105 & 0.011 & 0.111 & 0.092 & $-$0.045 & $-$0.016 & 0.008 & 0.079 & 10 & 0.040 & 0.066 & 0.079 \\
         \hline
    \end{tabular}
\end{table}

Models with low diversity (e.g., Llama-3.1-70B) reach near-saturation with 2--3 previous
stories, while high-diversity models (e.g., Gemini-2.5-pro) continue to improve with more
context. Notably, Llama-3.1-8B shows virtually no change across all settings ($-0.039$ throughout), suggesting that prior context had almost no effect on culprit prediction for this model.
For Gemini-1.5-pro, the experienced reader score slightly \emph{exceeds} the
upper bound ($0.308 > 0.234$), indicating mode collapse: the model's culprit choices are
predictable from prior stories regardless of the clues. While the know-it-all reader is Bayes-optimal and cannot be outperformed in expectation, sporadic exceedances can occur due to sampling noise in the know-it-all estimate or an unrepresentative sample of stories.

%%----------------------------------------------------------------------
\section{Full Results Tables}
\label{app:results_tables}

\begin{figure}[p]
\centering
\includegraphics[width=\linewidth]{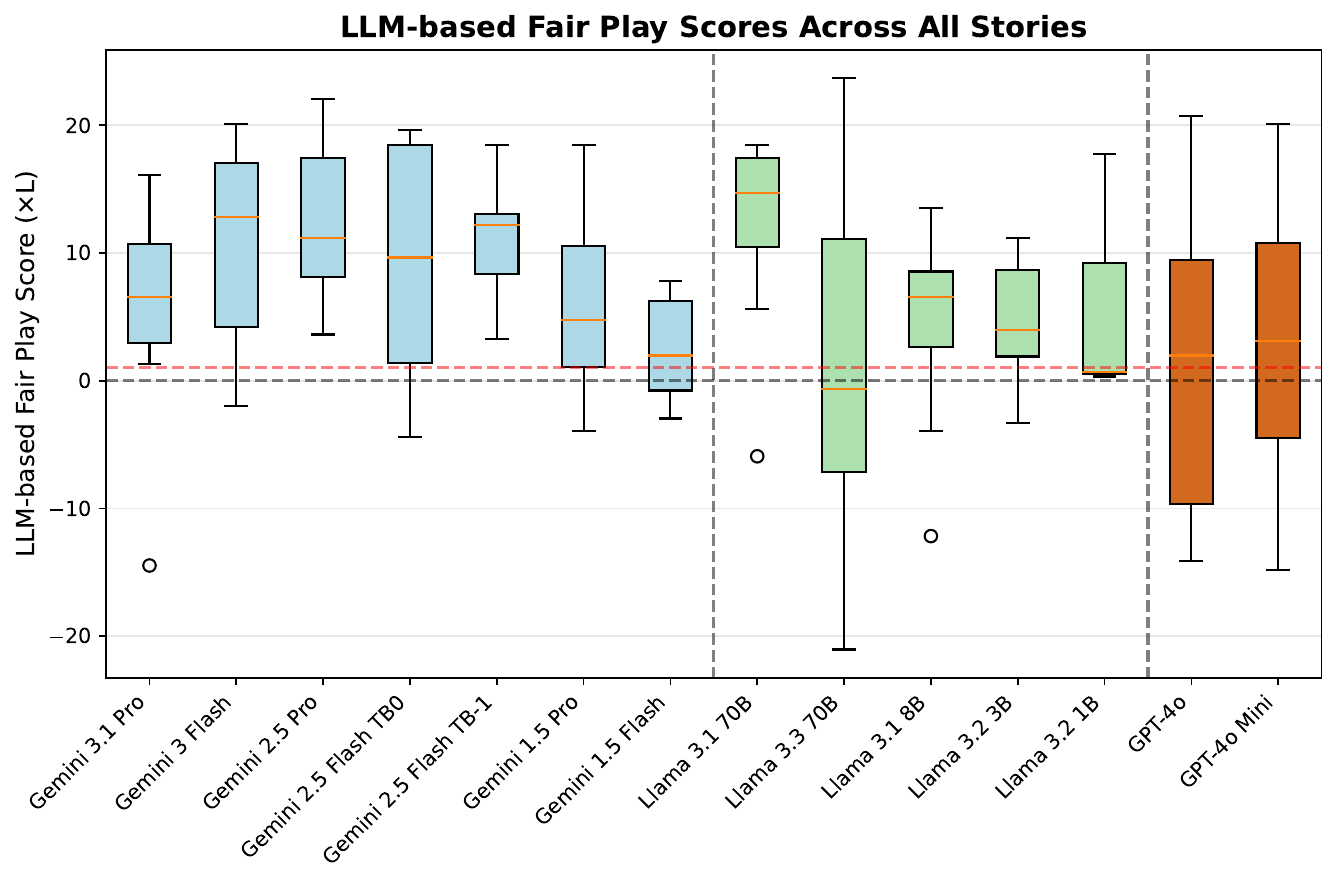}
\caption{Distribution of per-story fair play scores ($\fairplayubscore$) for all models. Each box shows the interquartile range; whiskers extend to $1.5\times$ IQR.}
\label{fig:fp-whisker}
\end{figure}

Table~\ref{tab:results} reports LLM-based scores ($\surprisalscore$, $\coherencescore$, $\fairplayubscore$, $\fairplayskepscore$) for all models on all valid generated stories.
Figure~\ref{fig:fp-whisker} shows the distribution of per-story fair play scores across models.
Table~\ref{tab:results_human} reports human fair play scores for the subset of stories with human annotations, alongside the experienced reader fair play score.

%%----------------------------------------------------------------------
\section{Surprise vs.\ Coherence}
\label{app:scatter}

\begin{figure}[p]
\centering
\includegraphics[width=0.85\linewidth]{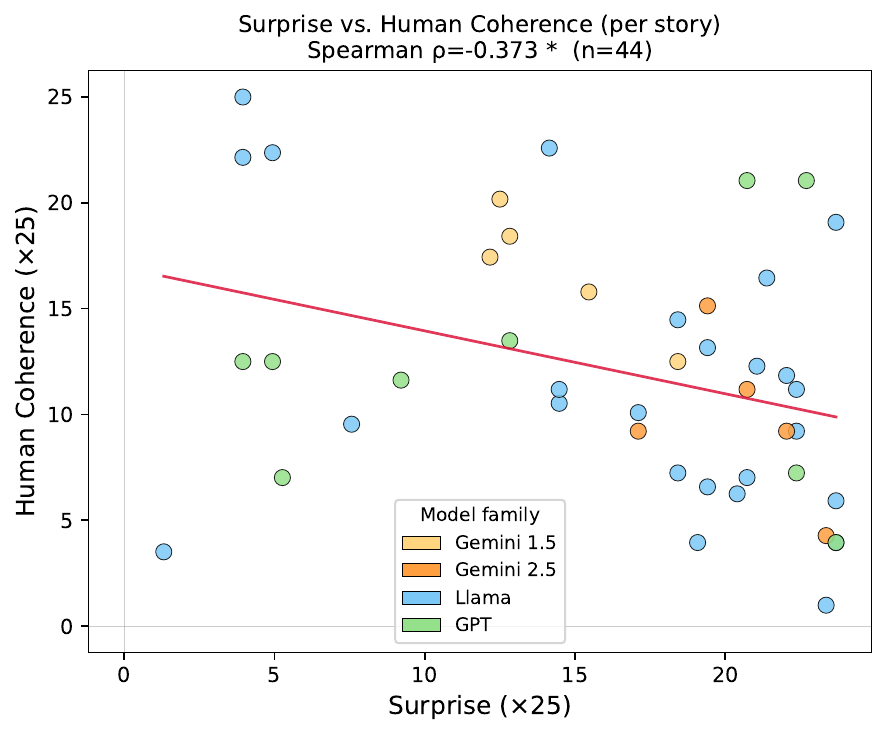}
\caption{Per-story surprise vs.\ fair play (human-based coherence). Each point is one story; colors indicate the generating model.}
\label{fig:scatter-human}
\end{figure}

\begin{figure}[p]
\centering
\includegraphics[width=0.85\linewidth]{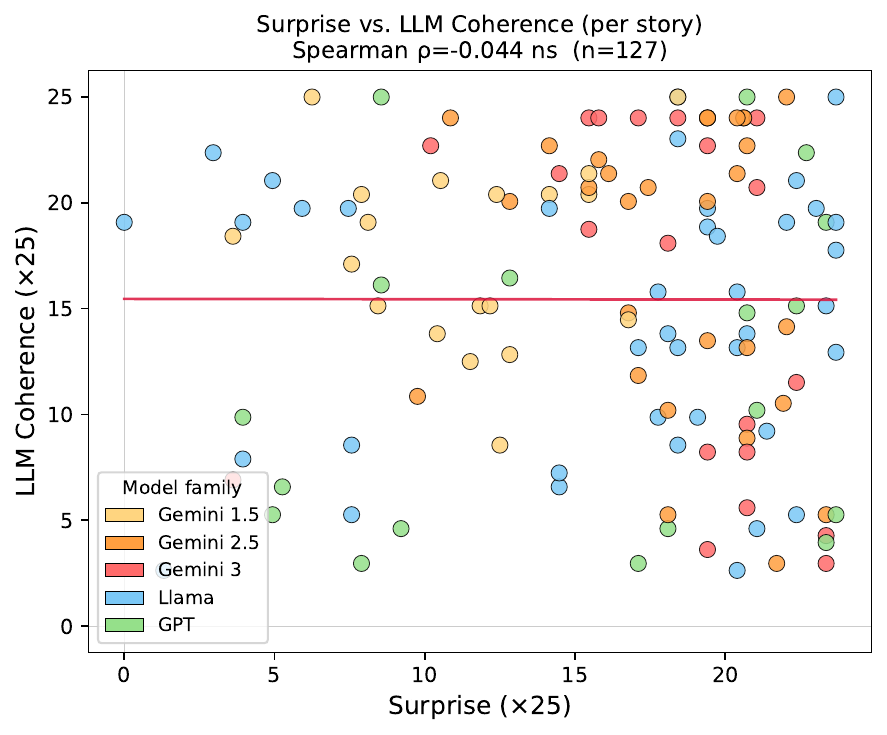}
\caption{Per-story surprise vs.\ fair play (LLM-based coherence). Each point is one story; colors indicate the generating model.}
\label{fig:scatter-llm}
\end{figure}

Figures~\ref{fig:scatter-human} and~\ref{fig:scatter-llm} show scatter plots of per-story surprise scores against fair play scores, using human and LLM-based coherence estimates respectively. Each point represents one story; models are distinguished by color.

\renewcommand{\arraystretch}{1.4}
\begin{table}[p]
    \caption{Results for various models. \generationsuccess\ indicates the ratio of
    generated stories that have a clear ($p>0.5$) culprit and distractor. $n$ is the
    number of valid stories used. $K$ is the number of continuations sampled per checkpoint
    for the know-it-all reader (20 for most models; 5 for GPT-4o, Gemini-1.5-flash, and Gemini-1.5-pro).
    $\surprisalscore$, $\coherencescore$, and $\fairplayubscore$ use the default (gullible) reader.
    $\fairplayskepscore$ is the solvability score using the uniform predictor:
    $\fairplayskepscore = \coherencescore - A(\hat{P}_\text{unif})$, where
    $A(\hat{P}_\text{unif}) = 1/|\culprits|$ is the accuracy of the uniform predictor.
    $\fairplayubscore = \coherencescore - (1-\surprisalscore)$.
    Models with $n < 5$ valid stories have metric values suppressed (``--'') as
    unreliable.
    All Llama models use the Instruct version. Bold entries indicate the two highest $\fairplayubscore$ values.}
    \label{tab:results}
    \centering
    \begin{tabular}{l|c|c|c|cccc}
         \hline
        Model & \generationsuccess & $n$ & $K$ & $\surprisalscore$ & $\coherencescore$ & $\fairplayubscore$ & $\fairplayskepscore$ \\
         \hline \hline
         Llama-3.2-1B & 0.27 & 3 & 20 & -- & -- & -- & -- \\
         Llama-3.2-3B & 0.73 & 8 & 20 & 0.570 & 0.612 & 0.182 & 0.125 \\
         Llama-3.1-8B & 0.90 & 9 & 20 & 0.744 & 0.423 & 0.167 & 0.068 \\
         Llama-3.1-70B & 0.83 & 10 & 20 & 0.745 & 0.746 & \textbf{0.491} & 0.335 \\
         Llama-3.3-70B & 1.00 & 10 & 20 & 0.574 & 0.466 & 0.040 & $-$0.045 \\
         \hline
         Gemini-1.5-flash & 0.71 & 9 & 5 & 0.352 & 0.734 & 0.086 & 0.115 \\
         Gemini-1.5-pro & 1.00 & 10 & 5 & 0.549 & 0.684 & 0.233 & 0.162 \\
         Gemini-2.5-flash TB 0 & 1.00 & 10 & 20 & 0.734 & 0.647 & 0.381 & 0.247 \\
         Gemini-2.5-flash TB -1 & 1.00 & 10 & 20 & 0.674 & 0.771 & 0.445 & 0.347 \\
         Gemini-2.5-pro & 1.00 & 10 & 20 & 0.798 & 0.687 & \textbf{0.485} & 0.307 \\
         Gemini-3-flash & 1.00 & 10 & 20 & 0.795 & 0.637 & 0.432 & 0.264 \\
         Gemini-3.1-pro & 1.00 & 10 & 20 & 0.646 & 0.584 & 0.230 & 0.109 \\
         \hline
         GPT-4o-mini & 0.57 & 8 & 20 & 0.669 & 0.451 & 0.120 & 0.137 \\
         GPT-4o & 0.67 & 10 & 5 & 0.562 & 0.480 & 0.042 & 0.084 \\
         \hline
    \end{tabular}
\end{table}

\begin{table}[p]
    \caption{Human fair play results. $n$ is the number of stories with human annotations.
    ``FP upper'' ($\fairplayubscore = \coherencescore - (1-\surprisalscore)$) uses the know-it-all reader for coherence; ``FP actual'' ($\arfpscore = C_{\text{human}} - (1-\surprisalscore)$) replaces coherence with the human average reader's accuracy, while keeping the same gullible-reader surprise $\surprisalscore$. $\expfpscore$ is the experienced reader fair play (with 9 previous
    stories). ``Diff'' is the upper minus actual fair play score. The experienced reader
    is not applicable to real stories (--).}
    \label{tab:results_human}
    \centering
    \begin{tabular}{l|c|ccccccc}
         \hline
        \multirow{2}{*}{Model} & \multirow{2}{*}{$n$} & \multirow{2}{*}{$\surprisalscore$}& \multicolumn{2}{c}{FP upper} & \multicolumn{2}{c}{FP actual} & \multirow{2}{*}{$\expfpscore$} & \multirow{2}{*}{Diff}\\
         &  &   & $\fairplayubscore$ & $\geq 0.04$ & $\arfpscore$ & $\geq 0.04$ & &\\
         \hline \hline
         Llama-3.2-3B  & 4 & 0.507 & 0.118 & 0.50 & 0.169 & 0.75 & 0.013 & $-$0.051\\
         Llama-3.1-8B & 5 & 0.782 & 0.182 & 0.80 & 0.037 & 0.20 & $-$0.039 & 0.145\\
         Llama-3.1-70B  & 6 & 0.864 & 0.538 & 1.00 & 0.345 & 0.83 & 0.522 & 0.193\\
         Llama-3.3-70B  & 10 & 0.574 & 0.039 & 0.40 & 0.039 & 0.80 & 0.011 & 0.000\\
         \hline
         Gemini-1.5-pro  & 5 & 0.571 & 0.234 & 0.60 & 0.246& 1.00 & 0.308 & $-$0.012\\
         Gemini-2.5-flash TB 0 & 5 & 0.732 & 0.351 & 0.80 & 0.066 & 0.40 & 0.143 & 0.285\\
         Gemini-2.5-flash TB -1 & 5 & 0.713 & 0.484 & 1.00 & 0.090 & 0.80 & 0.408 & 0.394\\
         Gemini-2.5-pro & 5 & 0.821 & 0.455 & 1.00 & 0.213 & 1.00 & 0.426 & 0.242\\
         \hline
         GPT-4o-mini  & 4 & 0.605 & $-$0.020 & 0.50 & 0.096 & 0.50 & $-$0.076 & $-$0.116\\
         GPT-4o  & 5 & 0.521 & 0.105 & 0.60 & 0.011 & 0.60 & 0.079 & 0.094\\
         \hline
         Average generated & 54 & 0.668 & 0.242 & 0.69 & 0.127 & 0.69 & 0.178 & 0.116\\
         \hline \hline
         Sherlock & 6 & 0.315 & -- & -- & 0.067 & 0.50 & -- & --\\
         Poirot & 5 & 0.664 & -- & -- & 0.325 & 1.00 & -- & --\\
         \hline
         Average real & 11 & 0.474 & -- & -- & 0.184 & 0.73 & -- & --\\
         \hline
    \end{tabular}
\end{table}
%%----------------------------------------------------------------------
\section{Variability Statistics}
\label{app:statistics}

Culprit-name diversity, inferred from the character statistics in Table~\ref{tab:results3}, varies considerably across models.
Gemini-1.5-pro exhibits strong culprit-name repetition (90\% of stories use the same
name). This mode collapse may explain the high performance of the experienced reader for this model: (1) the experienced reader score
substantially exceeds the know-it-all upper bound (Table~\ref{tab:exp-reader-per-model}),
showing that prior stories alone predict the culprit better than the story's own clues;
and (2) the gap between $\fairplayubscore$ and $\arfpscore$ is near zero, meaning that human
readers---who may also notice the pattern---perform close to the theoretical upper bound.

Gemini-3-flash shows mode collapse for supporting characters: the victim last name ``Sterling'' appears in 9/10 stories and the detective last name ``Thorne'' in 8/10 stories (Table~\ref{tab:results3}).
Culprit names are more varied (the most common culprit last name, ``Sterling'', appears in only 3/10 stories), so this character-level repetition is less likely to inflate the fair play score than in Gemini-1.5-pro.
Gemini-3.1-pro shows low name repetition across all roles---no single culprit, detective, or victim name appears in more than 2/10 stories---indicating substantially higher diversity than the Gemini-1.5 family.

\subsection{Story lengths}

Table~\ref{tab:story-lengths} reports mean story lengths (in words) per model.

\begin{table}[p]
    \caption{Mean story length (in words) $\pm$ standard deviation across 10 stories per model. Gemini-2.5 models include no thinking (TB 0) or dynamic thinking (TB -1).}
    \label{tab:story-lengths}
    \centering
    \footnotesize
    \setlength{\tabcolsep}{4pt}
    \begin{tabular}{l|ccccc}
         \hline
         & Llama-3.2-1B & Llama-3.2-3B & Llama-3.1-8B & Llama-3.1-70B & Llama-3.3-70B \\
         \hline \hline
         Length $\pm$ STD & $2875.6 \pm 1319.8$ & $3338.6 \pm 232.8$ & $3161.6 \pm 191.2$ & $3349.2 \pm 334.8$ & $5670.6 \pm 399.4$ \\
         \hline
    \end{tabular}

    \vspace{0.8em}

    \begin{tabular}{l|ccccc}
         \hline
         & \makecell{Gemini-1.5\\flash} & \makecell{Gemini-1.5\\pro} & \makecell{Gemini-2.5\\flash TB0} & \makecell{Gemini-2.5\\flash TB -1} & \makecell{Gemini-2.5\\pro TB -1} \\
         \hline \hline
         Length $\pm$ STD & $1183.5 \pm 107.7$ & $1136.7 \pm 83.5$ & $2134.9 \pm 238.2$ & $2026.3 \pm 164.3$ & $2437.1 \pm 241.2$ \\
         \hline
    \end{tabular}

    \vspace{0.8em}

    \begin{tabular}{l|cc}
         \hline
         & \makecell{Gemini-3\\flash} & \makecell{Gemini-3.1\\pro} \\
         \hline \hline
         Length $\pm$ STD & $4432.7 \pm 685.5$ & $4099.9 \pm 219.2$ \\
         \hline
    \end{tabular}

    \vspace{0.8em}

    \begin{tabular}{l|cc}
         \hline
         & GPT-4o-mini & GPT-4o \\
         \hline \hline
         Length $\pm$ STD & $4996.2 \pm 279.9$ & $6731.2 \pm 975.6$ \\
         \hline
    \end{tabular}
\end{table}

\subsection{Character statistics}

Table~\ref{tab:results3} reports character name statistics for generated stories.

\begin{table}[p]
    \caption{Character statistics for generated stories. $p(\text{t-cul.})$, $p(\text{t-det.})$, $p(\text{t-vic.})$ are the fraction of stories in which the same name was used for the true culprit, detective, and victim respectively (with the most common name in parentheses). $p(\text{M})$ is the fraction of stories in which that character was assigned a male gender. Gemini-2.5 models include no thinking (TB 0) or dynamic thinking (TB -1).}
    \label{tab:results3}
    \centering
    \begin{tabular}{l|m{12mm}m{7mm}|m{12mm}m{7mm}|m{12mm}m{7mm}}
         \hline
         Model & $p(\text{t-cul.})$ & $p(\text{M})$ & $p(\text{t-det.})$ & $p(\text{M})$ & $p(\text{t-vic.})$ & $p(\text{M})$  \\
         \hline \hline
         Llama-3.2-1B & 0.1 {\small(many)} & 0.7 & 0.6 \newline{\small(Jameson)} & 0.7& -- & --\\
         Llama-3.2-3B & 0.3 {\small(Rachel)} & 0.3 & 1. \newline{\small(Jameson)} & 1. & 0.5 {\small(Richard)} & 0.8\\
         Llama-3.1-8B& 0.2 {\small(Emily)} & 0.7 & 1. \newline{\small(Jameson)} & 1. &  0.5 {\small(Richard)} & 0.6\\
         Llama-3.1-70B& 0.4 {\small(Emma)} & 0.9 & 0.8 \newline{\small(Jameson)} & 0.9&  0.6 {\small(Richard)} & 0.8\\
         Llama-3.3-70B & 0.5 {\small(Emma)} & 0.5 & 1. \newline{\small(Jameson)} & 1. &  0.6 {\small(Richard)} & 1.\\
         \hline
         Gemini-1.5-flash  & 0.5 \newline {\small(Beatrice)} & 0.8 & 0.6 {\small(Davies)} & 1. &  1. \newline{\small(Ashworth)} & 1.\\
         Gemini-1.5-pro & 0.9 {\small(Blackwood)} & 0.9 & 0.6 {\small(Davies)} & 1. &  1. {\small(Blackwood)} & 1.\\
         Gemini-2.5-flash TB 0  & 0.2 {\small(2 names)} & 0.6 & 0.8 {\small(Corbin)} & 1. &  0.4 \newline{\small(Alistair)} & 1.\\
         Gemini-2.5-flash TB -1  & 0.5 \newline {\small(Finch)} & 0.9 & 0.5 {\small(Corbin)} & 1. &  0.6 \newline{\small(Ashworth)} & 1.\\
         Gemini-2.5-pro TB -1& 0.5 \newline {\small(Beatrice)} & 0.3 & 0.6 {\small(Corbin)} & 1. & 0.9 \newline{\small(Alistair)} & 1.\\
         Gemini-3-flash & 0.3 {\small(Sterling)} & 0.5 & 0.8 \newline{\small(Thorne)} & 1. & 0.9 \newline{\small(Sterling)} & 1.\\
         Gemini-3.1-pro & 0.2 {\small(3 names)} & 0.4 & 0.6 {\small(Vance)} & 1. & 0.5 {\small(Arthur)} & 1.\\
         \hline
         GPT-4o-mini & 0.1 {\small(many)} & 0.8 & 0.4 {\small(Clara)} & 0.3& 0.2 {\small(Harold)} & 0.5\\
         GPT-4o & 0.2 {\small(3 names)} & 0.5 & 0.4 {\small(Elanor)} & 0.0& 0.3 \newline{\small(Victor)} & 0.9\\
         \hline
    \end{tabular}
\end{table}

%%----------------------------------------------------------------------
\section{Additional Discussion}
\label{app:discussion}

When writing stories, humans usually decide many details early on in the writing process.
This is clearly the case when revising the story.
Similarly, Li et~al.\ \citep{li2024predictingvsactingtradeoff} show that models trained with RLHF
tend to use ``anchors'', which are phrases that tend to appear in many generated stories,
to guide the generation process. The anchors are beneficial for long generations,
constraining the output space, but reduce diversity. This phenomenon is sometimes referred
to as ``mode collapse'' \citep{hamilton-2024-detecting,jiang2025artificial}.
In our experiment, we see a similar trend, where stories use extremely similar names.
This could limit the impact of our experiment with sampling. Early convergence of a story
model can sometimes be attributed to stylistic preferences and not to the clue-related
content. A model that always generates the same story would show artificially high
coherence scores, since the know-it-all reader can predict any paragraph from a
point-mass distribution---but this reflects low diversity, not genuine fair play.

\paragraph*{Early convergence: definition and detection}
Early convergence is a form of structural mode collapse: the generating model assigns most probability mass to stories where the true culprit can be identified from the very beginning. This is a broader phenomenon that does not depend on a specific mechanism---it includes, but is not limited to, always naming the same character as culprit, placing the culprit in a stereotyped narrative position, or using distinctive linguistic patterns that the know-it-all reader can exploit without explicit clue-following.
Early convergence inflates $\coherencescore$ without reflecting genuine fair play: the know-it-all reader achieves high accuracy because distributional patterns betray the answer immediately, while the gullible-reader surprise $\surprisalscore$ is unaffected (it depends on a separate reader), so $\fairplayubscore = \coherencescore-(1-\surprisalscore)$ is inflated together with $\coherencescore$.

We detect early convergence by reporting the fraction of stories for which the know-it-all reader assigns probability $> 0.5$ to the true culprit at paragraphs~1, 3, and~5---paragraphs assumed to fall within the introduction for most models (we do not rely on the reported introduction-end markers, which can be unreliable).
A model with high rates at these early paragraphs is likely exhibiting structural mode collapse.
Table~\ref{tab:early-convergence} reports these rates per model.

\begin{table}[p]
\caption{Early convergence analysis: fraction of stories for which the know-it-all reader assigns probability $>0.5$ to the true culprit at paragraphs~1, 3, and~5 (assumed introduction paragraphs). Significance relative to a uniform-chance baseline (one-sided Poisson test): ${}^*p<0.05$, ${}^{**}p<0.01$, ${}^{***}p<0.001$. ``n/a'' indicates data unavailable.}
\label{tab:early-convergence}
\centering
\small
\begin{tabular}{l|ccc}
\hline
Model & Para~1 & Para~3 & Para~5 \\
\hline\hline
\multicolumn{4}{c}{\textit{Real stories}} \\
\hline
Sherlock Holmes     & 2/10 (20\%)          & n/a                       & n/a                       \\
Hercule Poirot      & 1/10 (10\%)          & 4/8 (50\%)$^*$            & 4/8 (50\%)$^*$            \\
\hline
\multicolumn{4}{c}{\textit{Artificial stories}} \\
\hline
Llama-3.1-8B        & 0/9 \ (0\%)           & 0/9 \ (0\%)               & 0/9 \ (0\%)               \\
Llama-3.2-3B        & 0/8 \ (0\%)           & 1/8 (12\%)$^*$            & 1/8 (12\%)$^*$            \\
Llama-3.1-70B       & 3/10 (30\%)$^{***}$  & 1/10 (10\%)$^*$           & 3/10 (30\%)$^{***}$       \\
Llama-3.3-70B       & 2/10 (20\%)$^{***}$  & 2/10 (20\%)$^{***}$       & 1/10 (10\%)$^*$           \\
Gemini-1.5-flash    & 2/9 \ (22\%)          & 2/9 \ (22\%)              & 1/9 \ (11\%)              \\
Gemini-1.5-pro      & 2/10 (20\%)          & 5/10 (50\%)$^{**}$        & 3/10 (30\%)               \\
Gemini-2.5-flash tb0  & 0/10 \ (0\%)       & 2/10 (20\%)$^{***}$       & 4/10 (40\%)$^{***}$       \\
Gemini-2.5-flash tb-1 & 0/10 \ (0\%)       & 2/10 (20\%)$^{***}$       & 1/10 (10\%)$^*$           \\
Gemini-2.5-pro      & 1/10 (10\%)$^*$     & 5/10 (50\%)$^{***}$       & 3/10 (30\%)$^{***}$       \\
Gemini-3-flash      & 0/10 \ (0\%)         & 7/10 (70\%)$^{***}$       & 6/10 (60\%)$^{***}$       \\
Gemini-3.1-pro      & 0/10 \ (0\%)         & 2/10 (20\%)$^{***}$       & 3/10 (30\%)$^{***}$       \\
GPT-4o-mini         & 0/8 \ (0\%)           & 0/8 \ (0\%)               & 1/8 (12\%)$^*$            \\
GPT-4o              & 2/10 (20\%)          & 2/10 (20\%)               & 1/10 (10\%)               \\
\hline
\end{tabular}
\end{table}

Somewhat surprisingly, in most cases the generating models do not exhibit strong early convergence: sampling multiple continuations yields diverse results, so structural mode collapse is usually not a dominant concern.
That said, several models show elevated rates at paragraphs~3--5: notably Gemini-3-flash ($70\%$ at paragraph~3), Gemini-2.5-pro and Gemini-1.5-pro ($50\%$ each), and Llama-3.1-70B ($30\%$ at paragraphs~1 and~5). GPT-4o and Llama-3.3-70B both show $20\%$ at paragraphs~1 and~3, though these rates do not reach statistical significance. Gemini-2.5-flash (tb0) shows a notable ramp-up: $0\%$ at paragraph~1 but $40\%$ by paragraph~5.
For models with elevated rates, the inflated $\coherencescore$ values should be interpreted with caution.
Ideally, we would like the model to sample from the space of relevant stories, yet
as argued by Wagner and Abend \citep{wagner2025language}, post-training with instructions disincentivizes
output diversity, which is precisely what is required for effective sampling from the
space of valid stories.

Our experiments show that some aspects of the story are relatively fixed and others show
variability. This aligns with the findings of Doshi and Hauser \citep{doi:10.1126/sciadv.adn5290}, that
human-AI collaboration seems to increase creativity at the individual level but decrease
the collective novelty.

%%----------------------------------------------------------------------
\section{Learning Effect Analysis}
\label{app:learning_effect}

We examine whether participants' ability to identify the culprit improves as they read more stories, by correlating story reading order with relative performance (accuracy adjusted for story difficulty).
For each participant, we computed the Spearman rank correlation between reading order and final prediction accuracy (whether the culprit was correctly identified by the end of the story).

For real stories, we find no significant correlation between reading order and performance (Spearman $r = 0.03$, $p = 0.82$), suggesting that participants do not noticeably improve at solving real mystery stories with practice.
For artificial stories, there is a statistically significant but small negative correlation ($r = -0.14$, $p = 0.021$), possibly due to fatigue.
A mixed-effects model with reading order as a fixed effect and participant as a random effect yields a significant but very small effect for the order variable ($\beta = -0.001$, $p = 0.034$) for artificial stories, confirming this pattern. The magnitude of the effect is negligible in practical terms, validating our treatment of the reader's predictions as independent in the statistical analysis.

The mean correlation across participants was $\rho = 0.04$ ($p = 0.61$, one-sample $t$-test
against zero), indicating no significant trend overall. The 95\% confidence interval
$[-0.06, 0.14]$ excludes a meaningful learning effect. We therefore conclude that reading
order does not confound the results.

\end{document}